\documentclass{article}
\usepackage[final]{neurips_data_2023}

\usepackage[utf8]{inputenc} 
\usepackage[T1]{fontenc}    
\usepackage{hyperref}       
\usepackage{url}            
\usepackage{booktabs}       
\usepackage{amsfonts}       
\usepackage{nicefrac}       
\usepackage{microtype}      
\usepackage{xcolor}         

\usepackage{natbib}
\setcitestyle{numbers,square}
\usepackage{textcomp, gensymb}
\usepackage{graphicx} 
\usepackage{tabularx}
\usepackage{hyperref}
\usepackage{comment}
\usepackage{bbding}
\usepackage{bm}
\usepackage{makecell}
\usepackage{amsthm}
\usepackage{amsfonts,amssymb}
\usepackage{amsmath}
\usepackage{physics}
\usepackage{pifont}
\usepackage{graphicx}
\usepackage{listings}
\usepackage{subcaption}
\usepackage{adjustbox}
\usepackage{tcolorbox}
\usepackage{algorithm}
\usepackage{algorithmic}
\usepackage{wrapfig}

\usepackage[misc]{ifsym}
\usepackage{enumitem}       
\usepackage{xurl}

\newcommand\nnfootnote[1]{%
  \begin{NoHyper}
  \renewcommand\thefootnote{}\footnote{#1}%
  \addtocounter{footnote}{-1}%
  \end{NoHyper}
}

\definecolor{blue}{RGB}{0,139,139}

\title{Safety-Gymnasium: A Unified Safe Reinforcement Learning Benchmark}

\author{
    \textbf{Jiaming Ji$^{1,*}$},
    \textbf{Borong Zhang$^{1,*}$},
    \textbf{Jiayi Zhou$^{1,*}$},
    \textbf{Xuehai Pan$^{1}$},
    \textbf{Weidong Huang$^{1}$} 
    \vspace{0.5em} \\
    \textbf{Ruiyang Sun$^{1}$},
    \textbf{Yiran Geng$^{1}$},
    \textbf{Yifan Zhong$^{1,2}$},
    \textbf{Juntao Dai$^{1}$},
    \textbf{Yaodong Yang~$^{1,\dag}$}
    \vspace{0.6em} \\
    $^1$\, Institute for AI, Peking University  \\
    $^2$\, Beijing Institute for General Artificial Intelligence (BIGAI) 
    \vspace{0.5em}\\
    \texttt{\{jiamg.ji, borongzh\}@gmail.com},~~ \texttt{gaiejj@outlook.com}\\
    \texttt{yaodong.yang@pku.edu.cn}
}

\begin{document}

\nnfootnote{$^*$Equal Contribution. $^{\dag}$Corresponding author.}
\nnfootnote{\,\,Work done when Jiayi Zhou visited Peking University.}

\maketitle

\begin{abstract}
  Artificial intelligence (AI) systems possess significant potential to drive societal progress. However, their deployment often faces obstacles due to substantial safety concerns. Safe reinforcement learning (SafeRL) emerges as a solution to optimize policies while simultaneously adhering to multiple constraints, thereby addressing the challenge of integrating reinforcement learning in safety-critical scenarios. In this paper, we present an environment suite called \texttt{Safety-Gymnasium}, which encompasses safety-critical tasks in both single and multi-agent scenarios, accepting vector and vision-only input. Additionally, we offer a library of algorithms named Safe Policy Optimization (\texttt{SafePO}), comprising 16 state-of-the-art SafeRL algorithms. This comprehensive library can serve as a validation tool for the research community. By introducing this benchmark, we aim to facilitate the evaluation and comparison of safety performance, thus fostering the development of reinforcement learning for safer, more reliable, and responsible real-world applications. The website of this project can be accessed at \url{https://sites.google.com/view/safety-gymnasium}.
\end{abstract}

\section{Introduction}
\label{introduction}
AI systems possess enormous potential to spur societal progress. However, their deployment is frequently hindered by substantial safety considerations \cite{carlson2010increasing, bi2021safety, yang2020projection, Bai_Zhang_Tao_Wu_Wang_Xu_2023}. Distinct from pure reinforcement learning (RL), Safe reinforcement learning (SafeRL) seeks to optimize policies while concurrently adhering to multiple constraints, addressing the challenge of employing RL in scenarios with critical safety implications \cite{tabular-cmdp-1989, altman2021constrained, zhang2023evaluating, ji2023omnisafe, ji2024ai}. This strategy proves particularly pertinent in real-world applications such as autonomous vehicles \cite{feng2023dense} and healthcare \cite{ou2023towards}, where system failures or unsafe actions can result in grave consequences, such as accidents or harm to individuals. In large language models (LLMs), some studies have also shown that the toxicity of the models can be reduced through SafeRL \cite{ji2023beavertails, dai2023safe}. Incorporating safety constraints ensures adherence to predefined boundaries and regulatory standards, fostering trust and enabling exploration in environments with high-risk potential. Overall, SafeRL is instrumental in guaranteeing the dependable operation of intelligent systems in intricate and high-stake domains.

Simulation environments have become instrumental in fostering the advancement of RL. Eminent examples such as Gym \cite{brockman2016openai}, Atari \cite{mnih2013playing}, and dm-control \cite{tunyasuvunakool2020dm_control} underline their importance. These versatile platforms permit researchers to swiftly design and execute varied tasks, thus enabling efficient evaluation of algorithmic effectiveness and intrinsic limitations. However, within the sphere of SafeRL, there is a notable dearth of dedicated simulation environments, which impedes comprehensive exploration of SafeRL. In recent years, there have been strides to address this gap. DeepMind presented AI-Safety-Gridworlds, a suite of RL environments showcasing various safety properties of intelligent agents \cite{leike2017ai}. Afterward, OpenAI introduced the Safety Gym benchmark suite, a collection of high-dimensional continuous control environments incorporating safety-robot tasks \cite{ray2019benchmarking}. Over the past two years, several additional environments have been developed by researchers, including safe-control-gym \cite{yuan2022safe}, MetaDrive \cite{li2022metadrive}, etc. 

\textbf{Compared to Safety Gym\footnote{Again, we have no intention of attacking Safety Gym; the contribution of Safety Gym to the SafeRL community cannot be ignored, and Safety Gym also inspired this work. We hope that through our efforts, \texttt{Safety-Gymnasium} can further promote the development of SafeRL and give back to the entire RL community.}} \texttt{Safety-Gymnasium} inherits and expands the settings of some tasks of Safety Gym, aiming to bolster the community's growth further. Compared with Safety Gym, we have made the following major improvements:

\begin{itemize}[left=0.3cm]
    \item \textbf{Refactoring of the physics engine.} Safety Gym utilizes \textit{mujoco-py} to enable Python-based customization of MuJoCo components. However, \textit{mujoco-py} stopped updates and support after 2021. In contrast, \texttt{Safety-Gymnasium} supports MuJoCo directly, eliminating the reliance on \textit{mujoco-py}. This facilitates access to the latest MuJoCo features (\textit{e.g.}, rendering speed and accuracy improved, etc.) and lowers the entry barrier, particularly due to \textit{mujoco-py's} dependency on specific GCC versions and more.
    \item \textbf{Extension of Agent and Task Components.} Safety Gym initially supports only three agents and tasks. On this basis, \texttt{Safety-Gymnasium} has been further expanded, introducing more diverse agents and task components and expanding safety tasks to cover multi-agent domains. Finally, \texttt{Safety-Gymnasium} launched a high-dimensional test component based on Issac-Gym \cite{makoviychuk2021isaac}, further enriching the benchmark.
    \item \textbf{Enhanced Visual Task Support.} The visual components of Safety Gym are simplistic (consisting of basic geometric shapes), and \textit{mujoco-py} relies on OpenGL for visual rendering, which results in significant virtualization performance loss on headless servers. In contrast, Safety-Gymnasium, built on MuJoCo, achieves rendering speeds on CPU that are twice as fast as the former. Additionally, it offers more comprehensive visual component support.
    \item \textbf{Easy Installation and High Customization.} Safety Gym is cumbersome to install and relies heavily on the underlying software. One of the design motivations of \texttt{Safety-Gymnasium} is the ease of use so that everyone can focus on algorithm design. \texttt{Safety-Gymnasium} can be easily installed with one simple command pip install safety-gymnasium. While benefiting from the highly integrated framework, \texttt{Safety-Gymnasium} only needs 100 lines of code to customize the required environment.
\end{itemize}

In this work, we introduce \texttt{Safety-Gymnasium}, a collection of environments specifically for SafeRL, built upon the Gymnasium \cite{brockman2016openai, Gymnasium} and MuJoCo \cite{todorov2012mujoco}. Enhancing the extant Safety Gym framework \cite{ray2019benchmarking}, we address various concerns and expand the task scope to include vision-only and multi-agent scenarios. Additionally, we released \texttt{SafePO}, a single-file style algorithm library containing over 16 state-of-the-art algorithms.  Collectively, our contributions are enumerated as follows:
\begin{itemize}[left=0.3cm]
    \item \textbf{Environmental Components.} We provide various safety-oriented tasks under the umbrella of \texttt{Safety-Gymnasium}. These tasks encompass single-agent, multi-agent, and vision-based challenges, each with varying constraints. Our environments are categorized into two primary types: Gymnasium-based, featuring agents of escalating complexity for algorithm verification and comparison, and Issac-Gym-based, incorporating sophisticated agents that harness the parallel processing power of Issac-gym's GPU. This empowers researchers to explore SafeRL algorithms in complex scenarios. Further details can be found in Section \ref{sec:env}.
    \item \textbf{Algorithm Components.} We offer the \texttt{SafePO} algorithm library, which comprises a single-file style housing 16 diverse algorithms. These algorithms encompass both single-agent and multi-agent approaches, along with first-order and second-order variants, as well as Lagrangian-based and Projection-based methods. Through meticulous decoupling, each algorithm's code resides in an individual file. A more in-depth exploration of \texttt{SafePO} is presented in Section \ref{sec:algo}.
    \item \textbf{Insights and Analysis.} Combining \texttt{Safety-Gymnasium} and \texttt{SafePO}, we conduct a detailed analysis of existing algorithms. Our analysis encompasses 16 algorithms across 54 distinct environments, covering various scenarios such as single-agent and multi-agent setups with varying constraint complexities. This analysis delves into each algorithm's strengths, constraints, and avenues for enhancement. We provide access to all metadata, fostering community verification and encouraging further research. Further details can be found in Section \ref{sec:exp}.
\end{itemize}

\section{Related Work}
\paragraph{Safety Environments}
In RL, agents need to explore environments to learn optimal policies by trial and error. It is currently typical to train RL agents mostly or entirely in simulation, where safety concerns are minimal. However, we anticipate that challenges in simulating the complexities of the real world (\textit{e.g.}, human-AI collaborative control~\cite{carlson2010increasing, bi2021safety}) will cause a shift towards training RL agents directly in the real world, where safety concerns are paramount \cite{li2022metadrive, xu2022trustworthy, gu2022review}. OpenAI includes safety requirements in the Safety Gym \cite{ray2019benchmarking}, which is a suite of high-dimensional continuous control environments for measuring research progress on SafeRL. Safe-control-gym~\cite{yuan2022safe} allows for constraint specification and disturbance injection onto a robot’s inputs, states, and inertial properties through a portable configuration system. DeepMind also presents a suite of RL environments, AI-Safety-Gridworlds \cite{leike2017ai}, illustrating various safety properties of intelligent agents.

\paragraph{SafeRL Algorithms}
CMDPs have been extensively studied for different constraint criteria~\cite{tabular-cmdp-1983, garcia2015comprehensive, Dai_Ji_Yang_Zheng_Pan_2023, huang2023safedreamer}. 
With the rise of deep learning, CMDPs are also moving to more high-dimensional continuous control problems. 
CPO~\cite{achiam2017constrained} proposes the first general-purpose policy search algorithm for SafeRL with guarantees for near-constraint satisfaction at each iteration.
However, CPO's policy updates hinge on Taylor approximations and the inversion of high-dimensional Fisher information matrices. These approximations can occasionally lead to inappropriate policy updates. 
FOCOPS~\cite{zhang2020first} applies a primal-dual approach to solve the constrained trust region problem directly and subsequently projects the solution back into the parametric policy space.
Similarly, CUP~\cite{yang2022cup} offers non-convex implementations through a first-order optimizer, thereby not requiring a strong approximation of the convexity of the objective.

\section{Preliminaries}
\subsection{Constrained Markov decision process}
SafeRL \cite{altman2021constrained, sutton2018reinforcement} is often formulated as a Constrained Markov decision process (CMDP) \cite{altman2021constrained}, which is a tuple $\mathcal{M} = (\mathcal{S}, \mathcal{A}, \mathbb{P}, R, \mathcal{C}, \mu, \gamma)$. Here $\mathcal{S}$ and $\mathcal{A}$ are the state space and action space correspondingly. $\mathbb{P}(s'| s,a)$ is the probability of state transition from $s$ to $s'$ after taking action $a$. $R(s' | s,a)$ denotes the reward obtained by the agent performing action $a$ in state $s$ and transitioning to state $s'$. The set $\mathcal{C} = \big\{ (c_i, b_i) \big\}_{i=1}^m$, where $c_i$ are cost functions: $c_i: \mathcal{S} \times \mathcal{A} \rightarrow \mathbb{R}$ and the cost thresholds are $b_i, i = 1, \cdots, m$. $\mu(\cdot): \mathcal{S} \rightarrow [0, 1]$ is the initial state distribution and the discount factor $\gamma \in [0, 1)$. 

A stationary parameterized policy $\pi_{\theta}$ is a probability distribution defined on $\mathcal{S} \times \mathcal{A}$, $\pi_{\theta}(a|s)$ denotes the probability of taking action $a$ in state $s$. We use $\Pi_{\theta} = \{\pi_{\theta}: \theta \in \mathbb{R}^p\}$ to denote the set of all stationary policies and $\theta$ is the network parameter needed to be learned. Let $\bm{P}_{\pi_\theta} \in \mathbb{R} ^{|S| \times |S|}$ denotes a state transition probability matrix and the components are: $\bm{P}_{\pi_\theta} [s, s'] = \mathbb{P}_{\pi_\theta}(s'|s) = \sum_{a \in \mathcal{A}} \pi_\theta(a|s)\mathbb{P}(s'|s, a)$, which denotes one-step state transition probability from $s$ to $s'$ by executing $\pi_\theta$. Finally, we let $d_{\pi_\theta}^{s_0}(s) = (1 - \gamma) \sum_{t=0}^\infty \gamma^t \mathbb{P}_{\pi_\theta}(s_t=s | s_0)$ to be the stationary state distribution of the Markov chain starting at $s_0$ induced by policy $\pi_\theta$ and $d_{\pi_\theta}^{\mu}(s) = \mathbb{E}_{s_0 \sim \mu(\cdot)} [d_{\pi_\theta}^{\mu}(s)]$ to be the discounted state visitation distribution on initial distribution $\mu$.

The objective function is defined via the infinite horizon discounted reward function where for a given $\pi_\theta$, we have $J^R(\pi_\theta) = \mathbb{E} \qty[ \sum_{t=0}^{\infty} \gamma^t R(s_{t+1} | s_t, a_t) \middle| s_0 \sim \mu, a_t \sim \pi_\theta ]$. The cost function is similarly specified via the following infinite horizon discount cost function: $ J^C_i(\pi_\theta) = \mathbb{E} \qty[ \sum_{t=0}^{\infty} \gamma^t C_i(s_{t+1} | s_t, a_t) \middle| s_0 \sim \mu, a_t \sim \pi_\theta ]$.

Then, we define the feasible policy set $\Pi_{\mathcal{C}}$ as : $\Pi_{\mathcal{C}} = \cap_{i=1}^m \{ \pi_\theta \in \Pi_\theta ~~\mathrm{and}~~ J^C_i(\pi_\theta) \le  b_i\}$. The goal of CMDP is to search the optimal policy $\pi_\star$:
    $\pi_\star = \arg \max_{\pi_\theta \in \Pi_{\mathcal{C}}} J^{R}(\pi_\theta)$.

\subsection{Constrained Markov Game}
Safe multi-agent reinforcement learning is often formulated as a Constrained Markov Game $(\mathcal{N}, \mathcal{S}, \mathcal{A}, \mathbb{P}, \mu, \gamma, R, \bm{C}, \bm{b})$. Here, $\mathcal{N}=$ $\{1, \ldots, n\}$ is the set of agents, $\mathcal{S}$ and $\mathcal{A}=$ $\prod_{i=1}^n \mathcal{A}^i$ are the state space and the joint action space (\textit{i.e.}, the product of the agents' action spaces), $\mathbb{P}: \mathcal{S} \times \mathcal{A} \times \mathcal{S} \rightarrow \mathbb{R}$ is the probabilistic transition function, $\mu$ is the initial state distribution, $\gamma \in[0,1)$ is the discount factor, $R: \mathcal{S} \times \mathcal{A} \rightarrow \mathbb{R}$ is the joint reward function, $\bm{C}=\left\{C_j^i\right\}_{1 \leq j \leq m^i}^{i \in \mathcal{N}}$ is the set of sets of cost functions (every agent $i$ has $m^i$ cost functions) of the form $C_j^i: \mathcal{S} \times \mathcal{A}^i \rightarrow \mathbb{R}$, and finally the set of corresponding cost threshold is given by $\bm{b}=\left\{b_j^i\right\}_{1 \leq j \leq m^i}^{i \in \mathcal{N}}$. At time step $t$, the agents are in a state $\mathrm{s}_t$, and every agent $i$ takes an action $\mathrm{a}_t^i$ according to its policy $\pi^i\left(\mathbf{a}^i \mid \mathbf{s}_t\right)$. Together with other agents' actions, it gives a joint action $\mathbf{a}_t=\left(\mathrm{a}_t^1, \ldots, \mathrm{a}_t^n\right)$ and the joint policy $\boldsymbol{\pi}(\mathbf{a} \mid \mathbf{s})=\prod_{i=1}^n \pi^i\left(\mathbf{a}^i \mid \mathbf{s}\right)$. The agents receive the reward $R\left(\mathrm{~s}_t, \mathbf{a}_t\right)$, meanwhile each agent $i$ pays the costs $C_j^i\left(\mathrm{~s}_t, \mathrm{a}_t^i\right)$, $\forall j=1, \ldots, m^i$. The environment then transits to a new state $\mathrm{s}_{t+1} \sim \mathbb{P}\left(\cdot \mid \mathrm{s}_t, \mathbf{a}_t\right)$.

The objective of reward function are $J(\bm{\pi}) \triangleq \mathbb{E}_{\mathrm{s}_0 \sim \rho^0, \mathbf{a}_{0: \infty} \sim \boldsymbol{\pi}, \mathrm{s}_{1: \infty} \sim \mathrm{p}} \qty[ \sum_{t=0}^{\infty} \gamma^t R \qty( \mathbf{s}_t, \mathbf{a}_t ) ]
$, and costs function are $ J_j^i(\vb*{\pi}) \triangleq \mathbb{E}_{\mathrm{s}_0 \sim \rho^0, \mathbf{a}_{0: \infty} \sim \boldsymbol{\pi}, \mathrm{s}_{1: \infty} \sim \mathrm{p}}\left[\sum_{t=0}^{\infty} \gamma^t C_j^i\left(\mathrm{~s}_t, \mathrm{a}_t^i\right)\right] \leq c_j^i, \qquad \forall j=1, \ldots, m^i$.

We are examining a fully cooperative setting where all agents share a common reward function. Consequently, the goal of safe multi-agent RL is to identify the optimal policy that maximizes the expected total reward while simultaneously ensuring that the safety constraints of each agent are satisfied. Then we define the feasible joint policy set $\bm{\pi}_{\mathcal{C}} = \cap_{i=1}^n \{ \pi_\theta \in \Pi_\theta ~~\mathrm{and}~~ J_j^i(\boldsymbol{\pi}) \leq c_j^i, \forall j=1, \ldots, m^i \}$. The goal of CMG is to search the optimal policy $\bm{\pi}_\star = \arg \max_{\pi_\theta \in \Pi_{\mathcal{C}}} J(\pi_\theta)$.


\section{Safety Environments: \texttt{Safety-Gymnasium}}
\label{sec:env}
\begin{figure}[ht]
  \centering
  \includegraphics[width=0.8\linewidth]{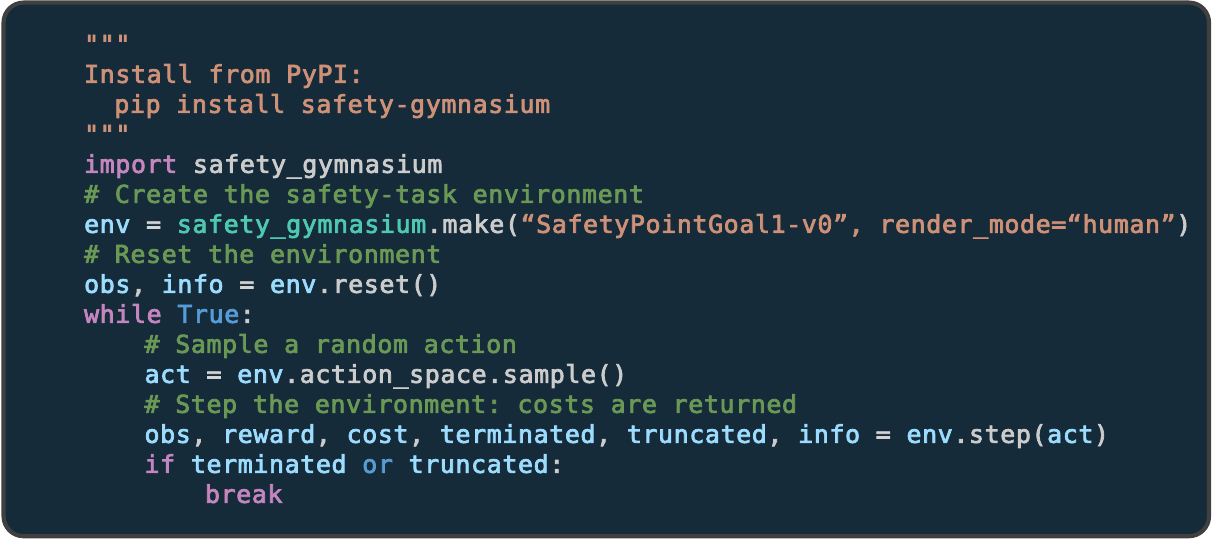}
  \caption{Using \texttt{Safety-Gymnasium} to create, step, render a specific safety-task environment.}
  \label{code}
\end{figure}

\texttt{Safety-Gymnasium} provides a seamless installation process and minimalistic code snippets to basic examples, as shown in \autoref{code}. Due to the limited space of the paper, we provide a more detailed description (\textit{e.g.}, detailed instructions, the composition of the robot's observation space and action space, dynamic structure, physical parameters, etc.) in \autoref{app:environment} and Online Documentation\footnote{Online Documentation: \url{www.safety-gymnasium.com}}.

\subsection{Gynasium-based Learning Environments}
In this section, we introduce Gymnasium-based environment components from three aspects: (1) the robots (both single-agent and multi-agent); (2) the tasks that are supported within the environment; (3) the safety constraints that are upheld.

\begin{figure}[ht]
  \centering
  \includegraphics[width=\linewidth]{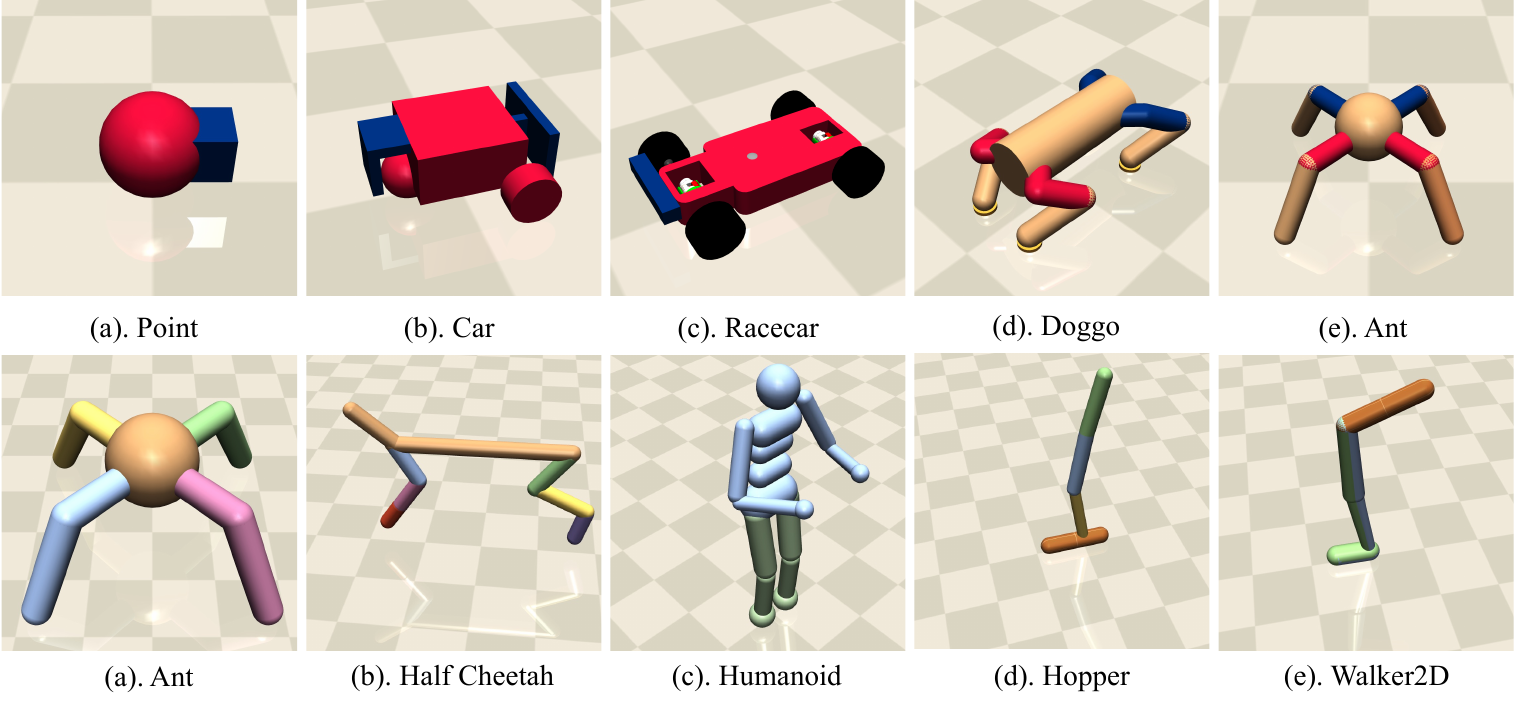}
  \caption{\textbf{Upper: }The Single-Agent Robots of Gymnasium-based Environments. \textbf{Lower:} The Multi-Agent Robots of Gymnasium-based Environments.}
  \label{pic:agent}
\end{figure}

\paragraph{Supported Robots}
As shown in \autoref{pic:agent}, \texttt{Safety-Gymnasium} inherits three pre-existing agents from Safety Gym \cite{ray2019benchmarking}, namely Point, Car, and Doggo. By meticulously adjusting the model parameters, we have successfully mitigated the issue of excessive oscillations during the runtime of \texttt{Point} and \texttt{Car} agents. Building upon this foundation, we have introduced two additional robots: \texttt{racecar} \cite{jacobs2002physiological, betz2019software}, and \texttt{ant} \cite{todorov2012mujoco}, to enrich the single-agent scenarios. As for multi-agent robots, we have leveraged certain configurations from multi-agent MuJoCo \cite{de2020deep}, deconstructing the original single-agent structure and enabling multiple agents to control distinct body segments. This design choice has been widely adopted in various research works \cite{kuba2021trust, yu2022surprising, gu2021multi}.

\begin{figure}[ht]
  \centering
  \includegraphics[width=\linewidth]{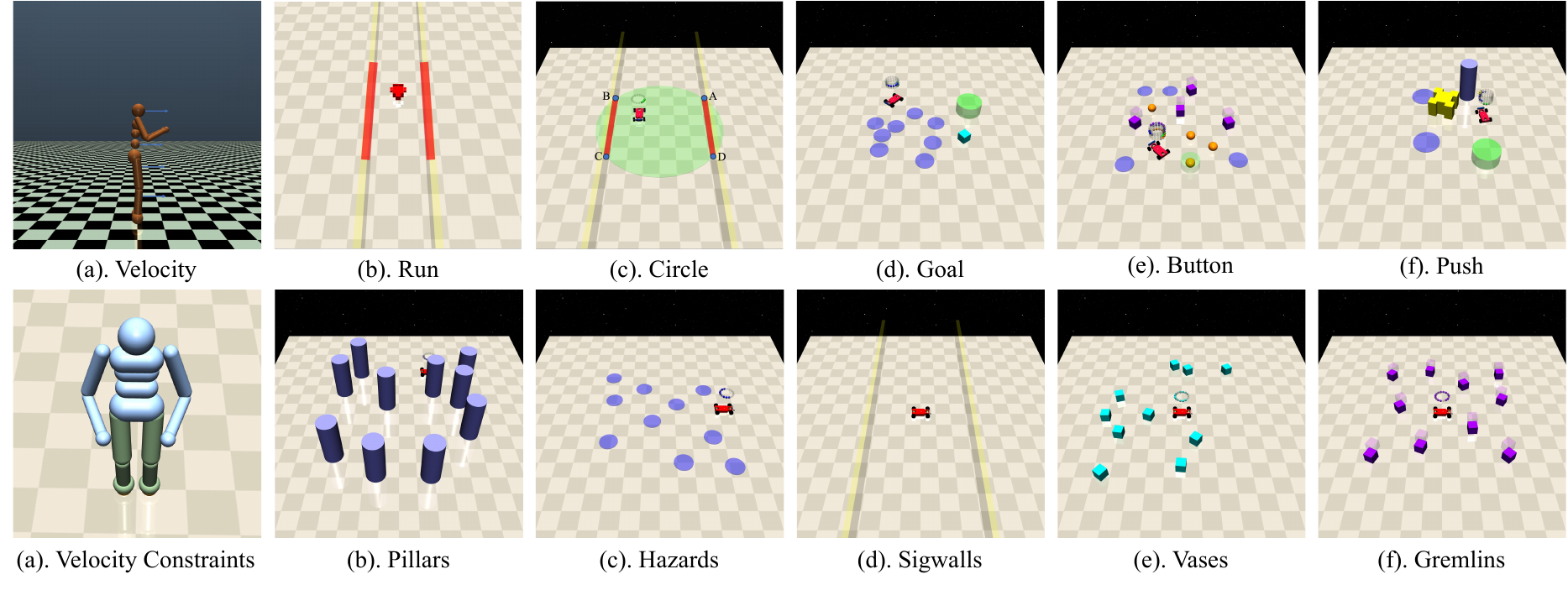}
  \caption{\textbf{Upper: }Tasks of Gymnasium-based Environments; \textbf{Lower: }Constraints of Gymnasium-based Environments.}
  \label{pic:constraints-gym}
\end{figure}

\paragraph{Supported Tasks}
As shown in \autoref{pic:constraints-gym}, the Gymnasium-based learning environments support the following tasks. For a more detailed task specification, please refer to our online documentation\footnote{Task Specification Documentation: \url{https://www.safety-gymnasium.com/en/latest/components_of_environments/tasks.html}}. 
\begin{itemize}[left=0.3cm]
    \item \textit{Velocity.} The robot aims to facilitate coordinated leg movement of the robot in the forward (right) direction by exerting torques on the hinges.
    \item \textit{Run.} The robot starts with a random initial direction and a specific initial speed as it embarks on a journey to reach the opposite side of the map.
    \item \textit{Circle.} The reward is maximized by moving along the green circle and not allowed to enter the outside of the red region, so its optimal path follows the line segments $AD$ and $BC$. 
    \item \textit{Goal.} The robot navigates to multiple goal positions. After successfully reaching a goal, its location is randomly reset while maintaining the overall layout. 
    \item \textit{Push.} The objective is to move a box to a series of goal positions. Like the goal task, a new random goal location is generated after each achievement. 
    \item \textit{Button.} The objective is to activate a series of goal buttons distributed throughout the environment. The agent's goal is to navigate towards and contact the currently highlighted button, known as the goal button.
\end{itemize}

\paragraph{Supported Constraints}
As shown in \autoref{pic:constraints-gym}, the Gymnasium-based environments support the following constraints. For a more detailed task specification, please refer to our online documentation.
\begin{itemize}[left=0.3cm]
    \item \textit{Velocity-Constraint} involves safety tasks using MuJoCo agents~\cite{todorov2012mujoco}. In these tasks, agents aim for higher reward by moving faster, but they must also adhere to velocity constraints for safety. Specifically, in a two-dimensional plane, the cost is computed as the Euclidean norm of the agent's velocities ($v_x$ and $v_y$).
    \item \textit{Pillars} are employed to represent large cylindrical obstacles within the environment. In the general setting, contact with a pillar incurs costs.
    \item \textit{Hazards} are utilized to model areas within the environment that pose a risk, resulting in costs when an agent enters such areas.
    \item \textit{Sigwalls} are designed specifically for Circle tasks. Crossing the wall from inside the safe area to the outside incurs costs.
    \item \textit{Vases} represent static and fragile objects within the environment. Touching or displacing these objects incurs costs for the agent.
    \item \textit{Gremlins} represent moving objects within the environment that can interact with the agent.
\end{itemize}

\subsubsection{Vision-only tasks}
\begin{figure}[ht]
  \centering
  \includegraphics[width=\linewidth]{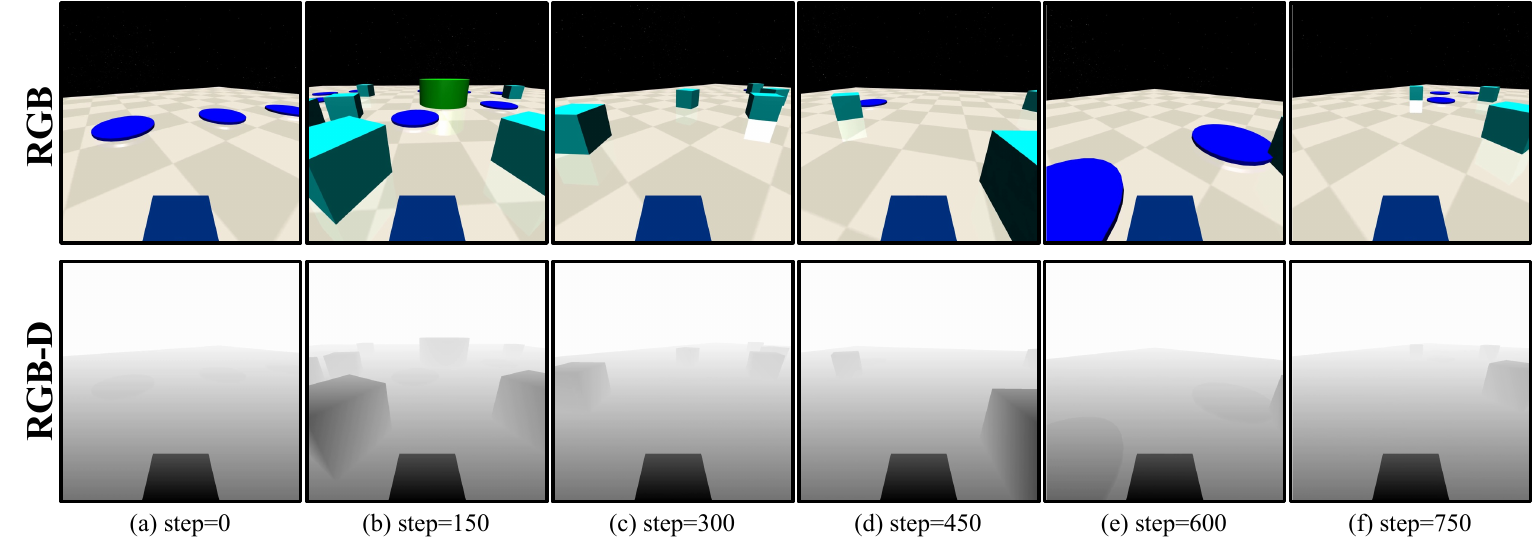}
  \caption{The RGB and RGB-D input of Gymnasium-based Environments.}
  \label{pic:rgdb}
\end{figure}

Vision-only SafeRL has gained significant attention as a focal point of research, primarily due to its applicability in real-world contexts \cite{ma2022conservative,as2022constrained}. While the initial iteration of Safety Gym offered rudimentary visual input support, there is room for enhancing the realism of its environment. To effectively evaluate vision-based SafeRL algorithms, we have devised a more realistic visual environment utilizing MuJoCo. This enhanced environment facilitates the incorporation of both RGB and RGB-D inputs (as shown in \autoref{pic:rgdb}). An exemplar of this environment is depicted in \autoref{pic:vision-only}, while comprehensive descriptions are available in Appendix \ref{app:vision-only-tasks}.

\begin{figure}[ht]
  \centering
  \includegraphics[width=\linewidth]{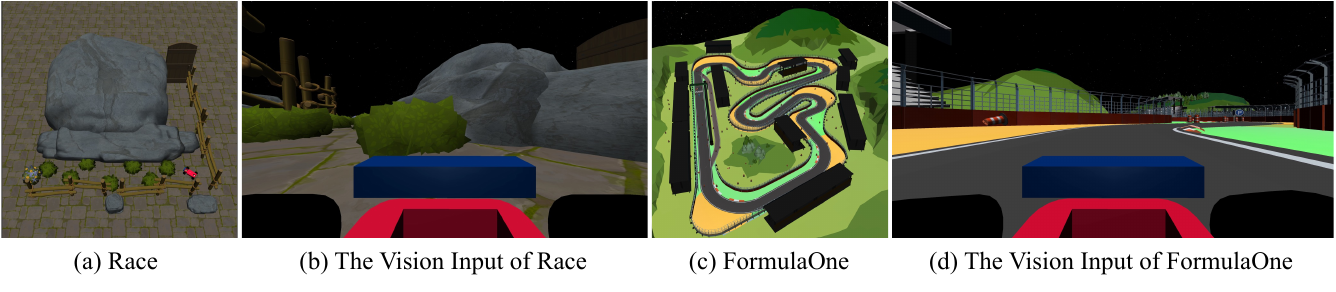}
  \caption{Vision-only Tasks of Gymnasium-based Environments.}
  \label{pic:vision-only}
\end{figure}

\subsection{Issac-Gym-based Learning Environments}
\begin{figure}[ht]
  \centering
  \includegraphics[width=\linewidth]{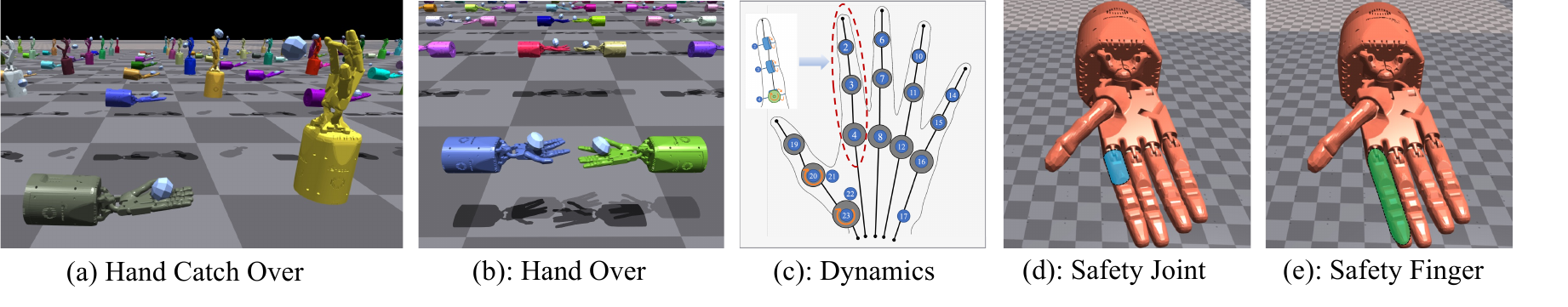}
  \caption{Tasks of Safety-DexterousHands.}
  \label{pic: dexterous-hand}
\end{figure}

In this section, we introduce \texttt{Safety-DexterousHands}, a collection of environments built upon DexterousHands \cite{chen2022towards} and the Isaac Gym engine \cite{makoviychuk2021isaac}. Leveraging GPU capabilities, Safety-DexterousHands enables large-scale parallel sample collection, significantly accelerating the training process. The environments support both single-agent and multi-agent settings. These environments involve two robotic hands (refer to \autoref{pic: dexterous-hand} (a) and (b)). In each episode, a ball randomly descends near the right hand. The right hand needs to grasp and launch the ball toward the left hand, which subsequently catches and deposits it at the target location.

For timestep $t$, let $x_{b,t}$, $x_{g,t}$ to be the position of the ball and the goal, $d_{p,t}$ to denote the positional distance between the ball and the goal $d_{p,t}=\|x_{b,t}-x_{g,t}\|_2$. Let $d_{a,t}$ denote the angular distance between the object and the goal, and the rotational difference is $d_{r,t}=2\arcsin \min\{|d_{a,t}|,1.0\}$. The reward is defined as follows, $r_{t}=\exp\{-0.2(\alpha d_{p,t}+d_{r,t})\}$, where $\alpha$ is a constant balance of positional and rotational reward.

\textbf{Safety Joint} constrains the freedom of joint \ding{175} of the forefinger (refer to \autoref{pic: dexterous-hand} (c) and (d)). Without the constraint, joint \ding{175} has freedom of $[-20\degree,20\degree]$. The safety tasks restrict joint \ding{175} within $[-10\degree, 10\degree]$. Let $\mathtt{ang\_4}$ be the angle of joint \ding{175},
and the cost is defined as:
$c_t=\mathbb{I}(\mathtt{ang\_4}\not\in [-10\degree, 10\degree])$.

\textbf{Safety Finger} constrains the freedom of joints  \ding{173},  \ding{174} and \ding{175} of forefinger (refer to \autoref{pic: dexterous-hand} (c) and (e)). Without the constraint, joints \ding{173} and \ding{174} have freedom of $[0\degree,90\degree]$ and joint \ding{175} of $[-20\degree,20\degree]$. The safety tasks restrict joints \ding{173}, \ding{174}, and \ding{175} within $[22.5\degree, 67.5\degree]$, $[22.5\degree, 67.5\degree]$, and $[-10\degree, 10\degree]$ respectively.
Let $\mathtt{ang\_2},\mathtt{ang\_3}, \mathtt{ang\_4}$ be the angles of joints \ding{173},  \ding{174}, \ding{175}, and the cost is defined as:
\begin{align}
c_t=\mathbb{I}(
\mathtt{ang\_2}\not\in [22.5\degree,67.5\degree],
~\text{or}~
\mathtt{ang\_3}\not\in [22.5\degree,67.5\degree],
~\text{or}~
\mathtt{ang\_4}\not\in [-10\degree,10\degree]
).
\end{align}

\section{Safe Policy Optimization Algorithms: \texttt{SafePO}}
\label{sec:algo}
This section provides a detailed discussion of the design of \texttt{SafePO}. Features such as strong performance, extensibility, customization, visualization, and documentation are all presented to demonstrate the advantages and contributions of \texttt{SafePO}.

\begin{figure}[ht]
  \centering
  \includegraphics[width=\linewidth]{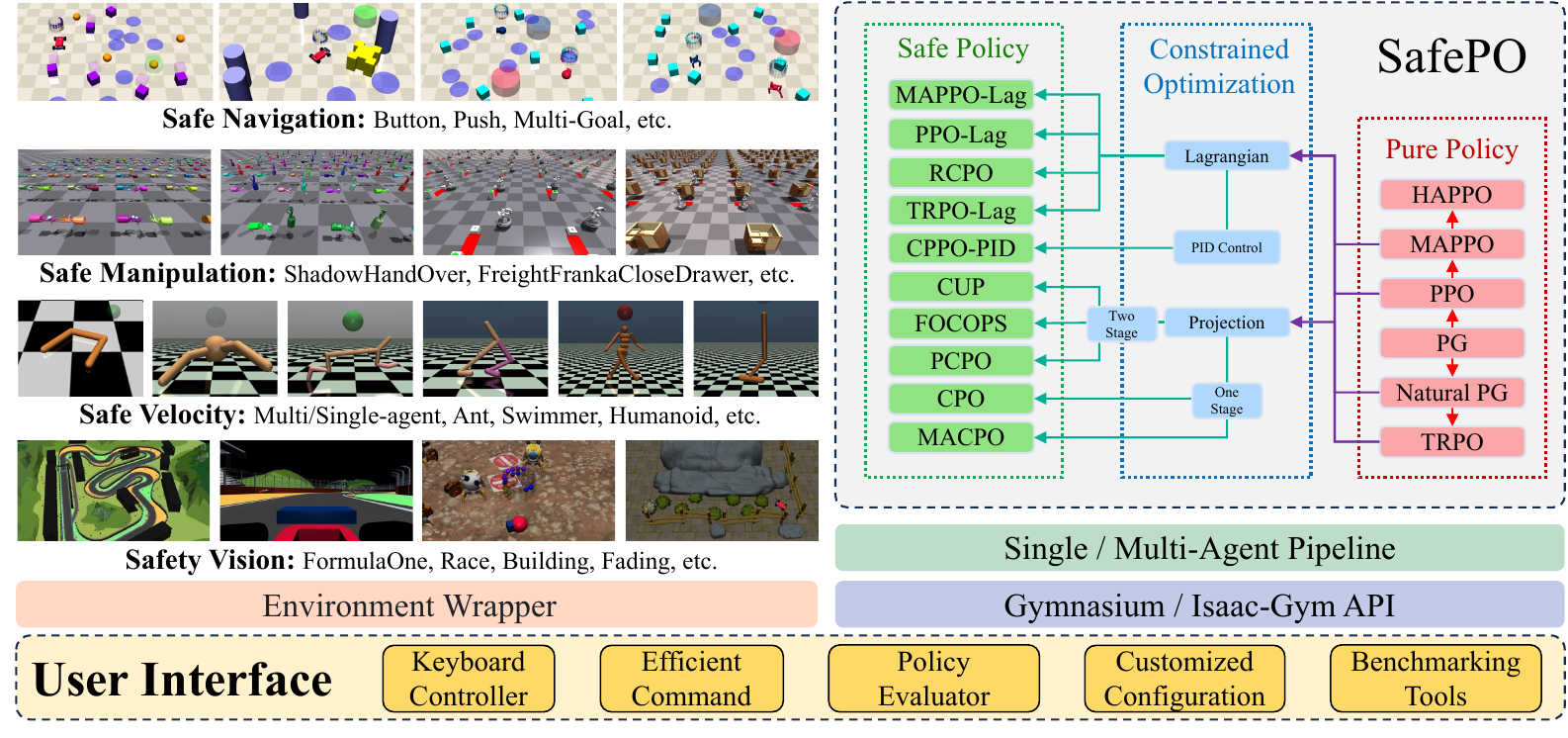}
  \caption{The Architecture of \texttt{SafePO}}
  \label{pic:architecture}
\end{figure}

\paragraph{Correctness} 
For a benchmark, it is critical to ensure its correctness and reliability. 
Firstly, each algorithm is implemented strictly according to the original paper (\textit{e.g.}, ensuring consistency with the gradient flow of the original paper, etc.). Secondly, we compare our implementation with those line by line for algorithms with a commonly acknowledged open-source code base to double-check the correctness. Finally, we compare \texttt{SafePO} with existing benchmarks (\textit{e.g.}, Safety-Starter-Agents\footnote{Safety-Starter-Agents: \url{https://github.com/openai/safety-starter-agents}} and RL-Safety-Algorithms\footnote{RL-Safety-Algorithms: \url{https://github.com/SvenGronauer/RL-Safety-Algorithms}}) and \texttt{SafePO} outperforms or achieves comparable performance with other existing implementations, as shown in Table \ref{tab:compare_with_other}.

\paragraph{Extensibility} 
\texttt{SafePO} enjoys high extensibility thanks to its architecture (as shown in \autoref{pic:architecture}). New algorithms can be integrated into \texttt{SafePO} by inheriting from base algorithms and only implementing their unique features. For example, we integrate PPO by inheriting from policy gradient and only adding the clip ratio variable and rewriting the function that computes the loss of policy $\pi$. Similarly, algorithms can be easily added to \texttt{SafePO}.

\paragraph{Logging and Visualization} 
Another necessary functionality of \texttt{SafePO} is logging and visualization. Supporting both TensorBoard and WandB, we offer code for visualizing more than 40 parameters and intermediate computation results to inspect the training process. Standard parameters and metrics such as KL-divergence, SPS (step per second), and cost variance are visualized universally. Special features of algorithms are also reported, such as the Lagrangian multiplier of Lagrangian-based methods, $g^T H^{-1}g, g^T H^{-1}b, \nu^*, \textrm{and } \lambda^*$ of CPO, proportional, integral, and derivative of PID-Lagrangian algorithms, etc. During training, users can inspect the changes of every parameter, collect the log file, and obtain saved checkpoint models. The complete and comprehensive visualization allows easier observation, model selection, and comparison.

\paragraph{Documentation} 
In addition to its code implementation, \texttt{SafePO} comes with an extensive documentation\footnote{\texttt{SafePO}'s Documentation: \url{https://safe-policy-optimization.readthedocs.io}}. We include detailed guidance on installation and propose solutions to common issues. Moreover, we provide instructions on simple usage and advanced customization of \texttt{SafePO}. Official information concerning maintenance, ethical, and responsible use are stated clearly for reference.

\begin{table}[h]
\centering
\captionsetup{skip=5pt}
\caption{A comparison between \texttt{SafePO} and other implementations. Results are based on 10 evaluation iterations using over 3 seeds under \texttt{cost\_limit=25.00}. 
$\bar{J}^R$ stands for normalized reward from PPO's performance, $\bar{J}^C$ signifies normalized cost relative to \texttt{cost\_limit}, and AvgR/AvgC represents the ratio of the means of both across 10 environments. The $\uparrow$ indicates higher rewards are better, while the $\downarrow$ indicates lower costs (when beyond the threshold of 1.00) are better. \textit{\textcolor{gray}{Gray}} and \textit{Black} depicts violation and compliance with the \texttt{cost\_limit}.
}
\label{tab:compare_with_other}
\resizebox{\columnwidth}{!}{
\begin{tabular}{@{}l|cc|cc|cc|cc|cc|cc|cc|cc|cc|cc@{}}
\toprule
 & \multicolumn{6}{c|}{\textbf{CPO}} & \multicolumn{6}{c|}{\textbf{TRPO-Lag}} & \multicolumn{4}{c|}{\textbf{PPO-Lag}} & \multicolumn{4}{c}{\textbf{FOCOPS}} \\ \midrule
 & \multicolumn{2}{c|}{\textbf{SafePO (Ours)}} & \multicolumn{2}{c|}{\textbf{Safety Starter Agents}} & \multicolumn{2}{c|}{\textbf{RL-Safety-Algorithms}} \hfill & \multicolumn{2}{c|}{\textbf{SafePO (Ours)}} & \multicolumn{2}{c|}{\textbf{Safety Starter Agents}} & \multicolumn{2}{c|}{\textbf{RL-Safety-Algorithms}}\hfill & \multicolumn{2}{c|}{\textbf{SafePO (Ours)}} & \multicolumn{2}{c|}{\textbf{Safety Starter Agents}} \hfill & \multicolumn{2}{c|}{\textbf{SafePO (Ours)}} & \multicolumn{2}{c}{\textbf{Original Implementation}} \hfill \\ \midrule
\textbf{Safety Navigation} \hfill  & $\bar{J}^R \uparrow$ & $\bar{J}^C \downarrow$ & $\bar{J}^R \uparrow$  & $\bar{J}^C \downarrow$  & $\bar{J}^R \uparrow$ & $\bar{J}^C \downarrow$ & $\bar{J}^R \uparrow$ & $\bar{J}^C \downarrow$  & $\bar{J}^R \uparrow$ & $\bar{J}^C \downarrow$ & $\bar{J}^R \uparrow$  & $\bar{J}^C \downarrow$  & $\bar{J}^R \uparrow$ & $\bar{J}^C \downarrow$ & $\bar{J}^R \uparrow$ & $\bar{J}^C \downarrow$  & $\bar{J}^R \uparrow$ & $\bar{J}^C \downarrow$  & $\bar{J}^R \uparrow$ & $\bar{J}^C \downarrow$ \\ \midrule
\textsc{CarButton1} & \textcolor{gray}{0.08} & \textcolor{gray}{1.75} & \textcolor{gray}{0.34} & \textcolor{gray}{3.65} & \textcolor{gray}{-0.06} & \textcolor{gray}{3.30} & \textcolor{gray}{-0.04} & \textcolor{gray}{1.08} & \textcolor{black}{0.02} & \textcolor{black}{0.78} & -0.05 & 0.63 & 0.01 & 0.47 & \textcolor{black}{0.02} & \textcolor{black}{0.67} & \textcolor{gray}{0.04} & \textcolor{gray}{1.21} & \textcolor{gray}{0.53} & \textcolor{gray}{6.02} \\
\textsc{CarGoal1} & \textcolor{gray}{0.78} & \textcolor{gray}{1.63} & \textcolor{gray}{0.94} & \textcolor{gray}{2.49} & \textcolor{gray}{0.46} & \textcolor{gray}{1.25} & \textcolor{gray}{0.82} & \textcolor{gray}{1.09} & \textcolor{gray}{0.72} & \textcolor{gray}{1.04} & \textcolor{black}{0.72} & \textcolor{black}{0.91} & 0.43 & 0.39 & \textcolor{black}{0.52} & \textcolor{black}{0.52} & \textcolor{black}{0.52} & \textcolor{black}{0.93} & \textcolor{gray}{0.79} & \textcolor{gray}{2.45} \\
\textsc{PointButton1} & \textcolor{gray}{0.12} & \textcolor{gray}{1.61} & \textcolor{gray}{0.70} & \textcolor{gray}{3.01} & \textcolor{gray}{0.03} & \textcolor{gray}{3.25} & \textcolor{gray}{0.27} & \textcolor{gray}{1.29} & \textcolor{black}{0.21} & \textcolor{black}{0.92} & 0.04 & 0.87 & \textcolor{gray}{0.22} & \textcolor{gray}{1.32} & \textcolor{black}{0.17} & \textcolor{black}{0.96} & \textcolor{gray}{0.25} & \textcolor{gray}{1.53} & \textcolor{gray}{0.70} & \textcolor{gray}{3.74} \\
\textsc{PointGoal1} & \textcolor{gray}{0.78} & \textcolor{gray}{1.10} & \textcolor{gray}{0.81} & \textcolor{gray}{1.99} & \textcolor{gray}{0.28} & \textcolor{gray}{2.05} & \textcolor{black}{0.72} & \textcolor{black}{0.91} & 0.65 & 0.94 & 0.33 & 0.72 & \textcolor{gray}{0.47} & \textcolor{gray}{1.50} & \textcolor{black}{0.66} & \textcolor{black}{0.77} & \textcolor{gray}{0.56} & \textcolor{gray}{1.32} & \textcolor{gray}{0.81} & \textcolor{gray}{1.53} \\
\midrule
\textbf{Safety Velocity} \hfill  & $\bar{J}^R \uparrow$ & $\bar{J}^C \downarrow$ & $\bar{J}^R \uparrow$  & $\bar{J}^C \downarrow$  & $\bar{J}^R \uparrow$ & $\bar{J}^C \downarrow$ & $\bar{J}^R \uparrow$ & $\bar{J}^C \downarrow$  & $\bar{J}^R \uparrow$ & $\bar{J}^C \downarrow$ & $\bar{J}^R \uparrow$  & $\bar{J}^C \downarrow$  & $\bar{J}^R \uparrow$ & $\bar{J}^C \downarrow$ & $\bar{J}^R \uparrow$ & $\bar{J}^C \downarrow$  & $\bar{J}^R \uparrow$ & $\bar{J}^C \downarrow$  & $\bar{J}^R \uparrow$ & $\bar{J}^C \downarrow$ \\ \midrule
\textsc{AntVel} & \textcolor{black}{0.52} & \textcolor{black}{0.56} & 0.31 & 0.93 & \textcolor{gray}{0.40} & \textcolor{gray}{1.09} & \textcolor{black}{0.53} & \textcolor{black}{0.15} & 0.32 & 0.76 & 0.44 & 0.70 & \textcolor{black}{0.54} & \textcolor{black}{0.22} & 0.31 & 0.61 & \textcolor{black}{0.55} & \textcolor{black}{0.60} & 0.52 & 0.39 \\
\textsc{HalfCheetahVel} & \textcolor{black}{0.40} & \textcolor{black}{0.23} & \textcolor{gray}{0.30} & \textcolor{gray}{1.13} & 0.31 & 0.97 & \textcolor{gray}{0.43} & \textcolor{gray}{1.01} & 0.25 & 0.79 & \textcolor{black}{0.43} & \textcolor{black}{0.67} & \textcolor{black}{0.44} & \textcolor{black}{0.04} & 0.30 & 0.93 & 0.42 & 0.12 & \textcolor{black}{0.44} & \textcolor{black}{0.04} \\
\textsc{HopperVel} & \textcolor{black}{0.73} & \textcolor{black}{0.48} & 0.35 & 0.93 & 0.26 & 0.68 & \textcolor{black}{0.59} & \textcolor{black}{0.71} & \textcolor{gray}{0.41} & \textcolor{gray}{1.11} & 0.24 & 0.57 & \textcolor{black}{0.58} & \textcolor{black}{0.89} & \textcolor{gray}{0.29} & \textcolor{gray}{1.20} & 0.66 & 0.30 & \textcolor{black}{0.74} & \textcolor{black}{0.53} \\
\textsc{HumanoidVel} & \textcolor{black}{0.71} & \textcolor{black}{0.01} & 0.05 & 0.19 & 0.36 & 0.83 & \textcolor{gray}{0.72} & \textcolor{gray}{2.38} & 0.05 & 0.01 & \textcolor{black}{0.71} & \textcolor{black}{0.79} & \textcolor{black}{0.72} & \textcolor{black}{0.76} & 0.07 & 0.09 & 0.71 & 0.93 & \textcolor{black}{0.73} & \textcolor{black}{0.43} \\
\textsc{SwimmerVel} & \textcolor{black}{0.51} & \textcolor{black}{0.82} & \textcolor{gray}{0.38} & \textcolor{gray}{1.11} & 0.41 & 0.82 & \textcolor{black}{0.66} & \textcolor{black}{0.84} & \textcolor{gray}{0.43} & \textcolor{gray}{1.67} & \textcolor{gray}{0.41} & \textcolor{gray}{1.02} & \textcolor{gray}{0.57} & \textcolor{gray}{1.11} & \textcolor{gray}{0.38} & \textcolor{gray}{1.18} & \textcolor{gray}{0.47} & \textcolor{gray}{1.30} & \textcolor{black}{0.68} & \textcolor{black}{0.71} \\
\textsc{Walker2dVel} & \textcolor{black}{0.39} & \textcolor{black}{0.81} & \textcolor{gray}{0.44} & \textcolor{gray}{1.85} & 0.05 & 0.67 & \textcolor{black}{0.51} & \textcolor{black}{0.77} & 0.46 & 0.67 & \textcolor{gray}{0.51} & \textcolor{gray}{1.34} & 0.44 & 0.20 & \textcolor{black}{0.47} & \textcolor{black}{0.81} & \textcolor{black}{0.50} & \textcolor{black}{0.68} & 0.48 & 0.74 \\
\midrule
\textbf{AvgR/AvgC} & \multicolumn{2}{c|}{\textbf{0.56}} & \multicolumn{2}{c|}{0.27} & \multicolumn{2}{c|}{0.17} & \multicolumn{2}{c|}{\textbf{0.51}} & \multicolumn{2}{c|}{0.40} & \multicolumn{2}{c|}{0.46} & \multicolumn{2}{c|}{\textbf{0.64}} & \multicolumn{2}{c|}{0.41} & \multicolumn{2}{c|}{\textbf{0.52}} & \multicolumn{2}{c}{0.39} \\
\bottomrule
\end{tabular}
}
\end{table}

\section{Experiments and Analysis}
\label{sec:exp}

\begin{figure}[ht]
\centering
\includegraphics[width=\linewidth]{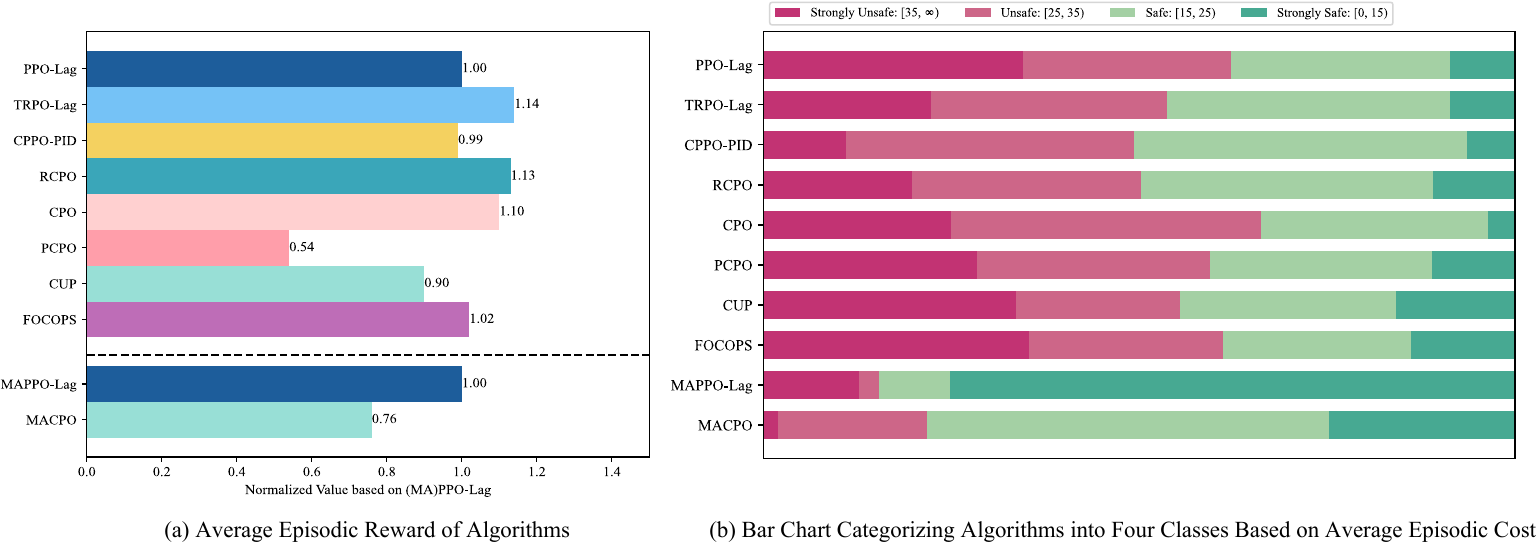}
\caption{A bar chart analyzing the performance of different algorithms. The left graph compares episodic reward with PPO-Lag \cite{ray2019benchmarking} (or MAPPO-Lag \cite{gu2021multi} for multi-agent). The right graph shows episodic costs proportionally under varying constraints. Single-agent data is from 40 navigation and 6 velocity tasks, and multi-agent data is from all 8 velocity tasks in \texttt{Safety-Gymnasium}.}
\label{rew-cost-oscil-analysis}
\end{figure}

\begin{table}[ht]
\centering
\caption{The performance of single-agent algorithms. 
$\bar{J}^R$ stands for normalized reward from PPO's performance, and $\bar{J}^C$ signifies normalized cost relative to \texttt{cost\_limit}. The $\uparrow$ indicates higher rewards are better, while the $\downarrow$ indicates lower costs (when beyond the threshold of 1.00) are better. \textit{\textcolor{gray}{Gray}} and \textit{Black} depicts breach and compliance with the \texttt{cost\_limit}, while \textit{\textcolor{blue}{Green}} represents the optimal policy, maximizing reward within safety constraints.}
\resizebox{\columnwidth}{!}{
\begin{tabular}{@{}l|cc|cc|cc|cc|cc|cc|cc|cc|cc@{}}\toprule
    & \multicolumn{2}{c|}{\textbf{PPO}} & \multicolumn{2}{c|}{\textbf{PPO-Lag}} & \multicolumn{2}{c|}{\textbf{TRPO-Lag}} & \multicolumn{2}{c|}{\textbf{CPPO-PID}} & \multicolumn{2}{c|}{\textbf{RCPO}} & \multicolumn{2}{c|}{\textbf{CPO}} & \multicolumn{2}{c|}{\textbf{PCPO}} & \multicolumn{2}{c|}{\textbf{CUP}} & \multicolumn{2}{c}{\textbf{FOCOPS}} \\
\midrule
\textbf{Safety Navigation} \hfill  & $\bar{J}^R \uparrow$ & $\bar{J}^C \downarrow$ & $\bar{J}^R \uparrow$ & $\bar{J}^C \downarrow$ & $\bar{J}^R \uparrow$  & $\bar{J}^C \downarrow$  & $\bar{J}^R \uparrow$ & $\bar{J}^C \downarrow$ & $\bar{J}^R \uparrow$ & $\bar{J}^C \downarrow$  & $\bar{J}^R \uparrow$ & $\bar{J}^C \downarrow$ & $\bar{J}^R \uparrow$  & $\bar{J}^C \downarrow$  & $\bar{J}^R \uparrow$ & $\bar{J}^C \downarrow$ & $\bar{J}^R \uparrow$ & $\bar{J}^C \downarrow$  \\ \midrule
\textsc{AntButton1} & \textcolor{gray}{1.00} & \textcolor{gray}{4.42} & 0.09 & 0.86 & \textcolor{gray}{0.23} & \textcolor{gray}{1.95} & \textcolor{blue}{0.10} & \textcolor{blue}{0.70} & \textcolor{gray}{0.16} & \textcolor{gray}{2.07} & \textcolor{gray}{0.12} & \textcolor{gray}{4.01} & \textcolor{gray}{0.03} & \textcolor{gray}{1.01} & 0.03 & 0.17 & 0.01 & 0.46 \\
\textsc{AntCircle1} & \textcolor{gray}{1.00} & \textcolor{gray}{16.81} & \textcolor{gray}{0.79} & \textcolor{gray}{2.56} & \textcolor{gray}{0.65} & \textcolor{gray}{1.05} & \textcolor{gray}{0.69} & \textcolor{gray}{1.90} & \textcolor{gray}{0.63} & \textcolor{gray}{1.04} & \textcolor{gray}{0.47} & \textcolor{gray}{1.07} & \textcolor{gray}{0.28} & \textcolor{gray}{1.87} & \textcolor{blue}{0.60} & \textcolor{blue}{0.82} & \textcolor{gray}{0.02} & \textcolor{gray}{1.22} \\
\textsc{AntGoal1} & \textcolor{gray}{1.00} & \textcolor{gray}{1.81} & 0.26 & 0.94 & 0.25 & 0.74 & \textcolor{gray}{0.47} & \textcolor{gray}{1.94} & \textcolor{blue}{0.29} & \textcolor{blue}{0.78} & 0.19 & 0.55 & 0.09 & 0.42 & \textcolor{gray}{0.34} & \textcolor{gray}{1.33} & 0.09 & 0.67 \\
\textsc{AntPush1} & \textcolor{gray}{1.00} & \textcolor{gray}{1.90} & 0.13 & 0.00 & 0.30 & 0.00 & 0.13 & 0.00 & \textcolor{blue}{0.33} & \textcolor{blue}{0.00} & 0.17 & 0.00 & 0.07 & 0.00 & 0.20 & 0.00 & -0.30 & 0.03 \\
\textsc{CarButton1} & \textcolor{gray}{1.00} & \textcolor{gray}{16.09} & \textcolor{blue}{0.01} & \textcolor{blue}{0.47} & \textcolor{gray}{-0.04} & \textcolor{gray}{1.08} & -0.10 & 0.40 & \textcolor{gray}{-0.19} & \textcolor{gray}{1.73} & \textcolor{gray}{0.08} & \textcolor{gray}{1.75} & \textcolor{gray}{0.02} & \textcolor{gray}{1.90} & \textcolor{gray}{0.04} & \textcolor{gray}{5.50} & \textcolor{gray}{0.04} & \textcolor{gray}{1.21} \\
\textsc{CarCircle1} & \textcolor{gray}{1.00} & \textcolor{gray}{8.42} & \textcolor{blue}{0.81} & \textcolor{blue}{0.82} & \textcolor{gray}{1.69} & \textcolor{gray}{2.77} & \textcolor{gray}{1.61} & \textcolor{gray}{1.79} & \textcolor{gray}{1.70} & \textcolor{gray}{3.11} & \textcolor{gray}{1.67} & \textcolor{gray}{3.13} & \textcolor{gray}{1.41} & \textcolor{gray}{1.99} & \textcolor{gray}{0.76} & \textcolor{gray}{1.04} & \textcolor{gray}{0.84} & \textcolor{gray}{1.12} \\
\textsc{CarGoal1} & \textcolor{gray}{1.00} & \textcolor{gray}{2.38} & 0.43 & 0.39 & \textcolor{gray}{0.82} & \textcolor{gray}{1.09} & \textcolor{gray}{0.03} & \textcolor{gray}{2.47} & \textcolor{blue}{0.55} & \textcolor{blue}{0.86} & \textcolor{gray}{0.78} & \textcolor{gray}{1.63} & \textcolor{gray}{0.61} & \textcolor{gray}{1.42} & 0.19 & 0.63 & 0.52 & 0.93 \\
\textsc{CarPush1} & \textcolor{gray}{1.00} & \textcolor{gray}{7.16} & 0.46 & 0.78 & \textcolor{blue}{1.38} & \textcolor{blue}{0.70} & 0.03 & 0.47 & \textcolor{gray}{1.11} & \textcolor{gray}{1.42} & \textcolor{gray}{0.83} & \textcolor{gray}{1.14} & \textcolor{gray}{0.64} & \textcolor{gray}{2.36} & 0.32 & 0.95 & 0.29 & 0.36 \\
\textsc{DoggoButton1} & \textcolor{gray}{1.00} & \textcolor{gray}{7.57} & 0.01 & 0.03 & \textcolor{gray}{0.00} & \textcolor{gray}{1.27} & 0.01 & 0.07 & 0.01 & 0.09 & 0.00 & 0.15 & 0.00 & 0.25 & \textcolor{blue}{0.02} & \textcolor{blue}{0.45} & \textcolor{gray}{0.06} & \textcolor{gray}{3.68} \\
\textsc{DoggoCircle1} & \textcolor{gray}{1.00} & \textcolor{gray}{33.14} & \textcolor{blue}{0.77} & \textcolor{blue}{0.46} & \textcolor{gray}{0.67} & \textcolor{gray}{1.37} & \textcolor{gray}{0.82} & \textcolor{gray}{2.16} & \textcolor{gray}{0.55} & \textcolor{gray}{1.32} & \textcolor{gray}{0.66} & \textcolor{gray}{1.22} & 0.31 & 0.55 & \textcolor{gray}{0.80} & \textcolor{gray}{2.04} & \textcolor{gray}{0.73} & \textcolor{gray}{4.49} \\
\textsc{DoggoGoal1} & \textcolor{gray}{1.00} & \textcolor{gray}{2.28} & 0.05 & 0.00 & 0.18 & 0.69 & 0.00 & 0.00 & \textcolor{gray}{0.16} & \textcolor{gray}{2.08} & \textcolor{blue}{0.30} & \textcolor{blue}{0.50} & 0.00 & 0.00 & 0.00 & 0.90 & \textcolor{gray}{0.04} & \textcolor{gray}{1.27} \\
\textsc{DoggoPush1} & \textcolor{gray}{1.00} & \textcolor{gray}{1.31} & 0.09 & 0.00 & \textcolor{blue}{0.53} & \textcolor{blue}{0.78} & 0.32 & 0.44 & \textcolor{gray}{0.54} & \textcolor{gray}{1.55} & 0.46 & 0.00 & 0.36 & 0.00 & 0.30 & 0.68 & \textcolor{gray}{0.64} & \textcolor{gray}{3.40} \\
\textsc{PointButton1} & \textcolor{gray}{1.00} & \textcolor{gray}{6.06} & \textcolor{gray}{0.22} & \textcolor{gray}{1.32} & \textcolor{gray}{0.27} & \textcolor{gray}{1.29} & \textcolor{blue}{0.00} & \textcolor{blue}{0.84} & \textcolor{gray}{0.12} & \textcolor{gray}{1.13} & \textcolor{gray}{0.12} & \textcolor{gray}{1.61} & \textcolor{gray}{0.08} & \textcolor{gray}{2.19} & \textcolor{gray}{0.18} & \textcolor{gray}{1.26} & \textcolor{gray}{0.25} & \textcolor{gray}{1.53} \\
\textsc{PointCircle1} & \textcolor{gray}{1.00} & \textcolor{gray}{8.10} & \textcolor{blue}{0.86} & \textcolor{blue}{0.93} & \textcolor{gray}{1.67} & \textcolor{gray}{1.35} & \textcolor{gray}{1.72} & \textcolor{gray}{2.09} & \textcolor{gray}{1.66} & \textcolor{gray}{1.42} & \textcolor{gray}{1.69} & \textcolor{gray}{1.74} & \textcolor{gray}{1.33} & \textcolor{gray}{2.26} & 0.82 & 0.62 & 0.84 & 0.89 \\
\textsc{PointGoal1} & \textcolor{gray}{1.00} & \textcolor{gray}{1.93} & \textcolor{gray}{0.47} & \textcolor{gray}{1.50} & \textcolor{blue}{0.72} & \textcolor{blue}{0.91} & \textcolor{gray}{0.31} & \textcolor{gray}{1.05} & 0.53 & 0.99 & \textcolor{gray}{0.78} & \textcolor{gray}{1.10} & 0.71 & 0.82 & 0.46 & 0.73 & \textcolor{gray}{0.56} & \textcolor{gray}{1.32} \\
\textsc{PointPush1} & \textcolor{gray}{1.00} & \textcolor{gray}{2.31} & \textcolor{gray}{0.98} & \textcolor{gray}{1.33} & 0.85 & 1.00 & 0.35 & 0.35 & \textcolor{blue}{5.30} & \textcolor{blue}{0.94} & 2.22 & 0.80 & \textcolor{gray}{1.72} & \textcolor{gray}{1.25} & 2.32 & 0.80 & \textcolor{gray}{1.13} & \textcolor{gray}{2.51} \\
\textsc{RacecarButton1} & \textcolor{gray}{1.00} & \textcolor{gray}{13.73} & \textcolor{gray}{-0.01} & \textcolor{gray}{1.94} & \textcolor{gray}{-0.02} & \textcolor{gray}{1.77} & \textcolor{gray}{-0.16} & \textcolor{gray}{2.06} & \textcolor{gray}{-0.07} & \textcolor{gray}{1.19} & \textcolor{gray}{0.00} & \textcolor{gray}{2.44} & \textcolor{gray}{0.02} & \textcolor{gray}{1.82} & \textcolor{gray}{0.00} & \textcolor{gray}{5.23} & \textcolor{gray}{-0.10} & \textcolor{gray}{3.37} \\
\textsc{RacecarCircle1} & \textcolor{gray}{1.00} & \textcolor{gray}{15.87} & \textcolor{gray}{0.83} & \textcolor{gray}{1.90} & \textcolor{gray}{0.80} & \textcolor{gray}{2.18} & \textcolor{gray}{0.58} & \textcolor{gray}{1.33} & \textcolor{gray}{0.83} & \textcolor{gray}{2.07} & \textcolor{blue}{0.79} & \textcolor{blue}{0.81} & \textcolor{gray}{0.22} & \textcolor{gray}{2.87} & \textcolor{gray}{0.74} & \textcolor{gray}{3.53} & \textcolor{gray}{0.77} & \textcolor{gray}{2.11} \\
\textsc{RacecarGoal1} & \textcolor{gray}{1.00} & \textcolor{gray}{4.26} & 0.26 & 0.51 & \textcolor{blue}{1.19} & \textcolor{blue}{0.77} & \textcolor{gray}{-0.04} & \textcolor{gray}{1.07} & 0.88 & 0.83 & \textcolor{gray}{1.18} & \textcolor{gray}{2.58} & 0.33 & 0.24 & \textcolor{gray}{0.13} & \textcolor{gray}{1.22} & 0.31 & 0.62 \\
\textsc{RacecarPush1} & \textcolor{gray}{1.00} & \textcolor{gray}{2.34} & -0.40 & 0.00 & \textcolor{gray}{0.74} & \textcolor{gray}{1.79} & \textcolor{gray}{-0.84} & \textcolor{gray}{2.87} & \textcolor{gray}{0.58} & \textcolor{gray}{1.92} & \textcolor{blue}{0.94} & \textcolor{blue}{0.13} & -0.16 & 0.18 & \textcolor{gray}{-0.06} & \textcolor{gray}{3.79} & \textcolor{gray}{0.30} & \textcolor{gray}{2.04} \\
\midrule
\textbf{Safety Velocity} \hfill  & $\bar{J}^R \uparrow$ & $\bar{J}^C \downarrow$ & $\bar{J}^R \uparrow$ & $\bar{J}^C \downarrow$ & $\bar{J}^R \uparrow$  & $\bar{J}^C \downarrow$  & $\bar{J}^R \uparrow$ & $\bar{J}^C \downarrow$ & $\bar{J}^R \uparrow$ & $\bar{J}^C \downarrow$  & $\bar{J}^R \uparrow$ & $\bar{J}^C \downarrow$ & $\bar{J}^R \uparrow$  & $\bar{J}^C \downarrow$  & $\bar{J}^R \uparrow$ & $\bar{J}^C \downarrow$ & $\bar{J}^R \uparrow$ & $\bar{J}^C \downarrow$  \\ \midrule
\textsc{AntVel} & \textcolor{gray}{1.00} & \textcolor{gray}{38.33} & 0.54 & 0.22 & 0.53 & 0.15 & 0.51 & 0.41 & 0.52 & 0.56 & 0.52 & 0.56 & 0.38 & 0.41 & \textcolor{blue}{0.55} & \textcolor{blue}{0.94} & 0.55 & 0.60 \\
\textsc{HalfCheetahVel} & \textcolor{gray}{1.00} & \textcolor{gray}{36.77} & 0.44 & 0.00 & \textcolor{gray}{0.43} & \textcolor{gray}{1.01} & \textcolor{blue}{0.48} & \textcolor{blue}{0.04} & 0.36 & 0.56 & 0.40 & 0.23 & 0.25 & 0.63 & 0.40 & 0.17 & 0.42 & 0.12 \\
\textsc{HopperVel} & \textcolor{gray}{1.00} & \textcolor{gray}{22.00} & 0.58 & 0.89 & 0.59 & 0.71 & 0.73 & 0.44 & 0.58 & 0.59 & 0.73 & 0.48 & 0.65 & 0.51 & \textcolor{blue}{0.73} & \textcolor{blue}{0.21} & 0.66 & 0.30 \\
\textsc{HumanoidVel} & \textcolor{gray}{1.00} & \textcolor{gray}{38.42} & 0.72 & 0.76 & \textcolor{gray}{0.72} & \textcolor{gray}{2.38} & \textcolor{blue}{0.73} & \textcolor{blue}{0.00} & 0.68 & 0.82 & 0.71 & 0.01 & 0.64 & 0.01 & 0.68 & 0.80 & 0.71 & 0.93 \\
\textsc{SwimmerVel} & \textcolor{gray}{1.00} & \textcolor{gray}{6.61} & \textcolor{gray}{0.57} & \textcolor{gray}{1.11} & 0.66 & 0.84 & \textcolor{blue}{0.91} & \textcolor{blue}{0.92} & 0.54 & 0.90 & 0.51 & 0.82 & 0.50 & 0.69 & 0.59 & 0.96 & \textcolor{gray}{0.47} & \textcolor{gray}{1.30} \\
\textsc{Walker2dVel} & \textcolor{gray}{1.00} & \textcolor{gray}{36.11} & 0.44 & 0.20 & \textcolor{blue}{0.51} & \textcolor{blue}{0.77} & 0.27 & 0.36 & 0.49 & 0.15 & 0.39 & 0.81 & 0.27 & 0.71 & 0.44 & 0.18 & 0.50 & 0.16 \\
\bottomrule
\end{tabular}
\label{sa-table}
}
\end{table}
\textbf{Reward and Cost.}
Episodic reward and cost exhibit a trade-off relationship. Unconstrained algorithms aim to maximize reward through risky behaviors. HAPPO \cite{kuba2021trust} achieves higher rewards compared to MAPPO \cite{yu2022surprising} across 8 velocity-based tasks, accompanied by a simultaneous increase in average costs. SafeRL algorithms tend to maximize reward while adhering to constraints. As depicted in \autoref{sa-table}, in the velocity task, compared to PPO \cite{schulman2017proximal}, PPO-Lag \cite{ray2019benchmarking} achieves a reduction of $98\%$ in cost while only experiencing a decrease of $45\%$ in reward.

\textbf{Randomness and Oscillation.}
The randomness of tasks is correlated with the oscillation of algorithms' performance. All SafeRL algorithms achieve average episodic costs within the \texttt{cost\_limit} for velocity tasks. The divergence in episodic rewards between algorithms is negligible, and the distribution of optimal policies is tightly clustered. However, pronounced oscillations are present in navigation tasks characterized by high stochasticity. 
Out of the 20 navigation tasks examined, optimal policies are spread out more, leading to observable differences in algorithm performance across various tasks.

\begin{wraptable}{tr}{0.5\textwidth}

\centering
\caption{The normalized performance of \texttt{SafePO}'s multi-agent algorithms on \texttt{Safety-Gymnasium}.}
\resizebox{0.5\columnwidth}{!}{
\begin{tabular}{@{}l|cc|cc|cc|cc@{}}\toprule
    & \multicolumn{2}{c|}{\textbf{MAPPO}} & \multicolumn{2}{c|}{\textbf{HAPPO}} & \multicolumn{2}{c|}{\textbf{MAPPO-Lag}} & \multicolumn{2}{c}{\textbf{MACPO}} \\
\midrule
\textbf{Safety Velocity} \hfill  & $\bar{J}^R \uparrow$ & $\bar{J}^C \downarrow$ & $\bar{J}^R \uparrow$  & $\bar{J}^C \downarrow$  & $\bar{J}^R \uparrow$ & $\bar{J}^C \downarrow$ & $\bar{J}^R \uparrow$ & $\bar{J}^C \downarrow$  \\ \midrule
\textsc{2x4AntVel} & \textcolor{gray}{1.00} & \textcolor{gray}{35.76} & \textcolor{gray}{1.26} & \textcolor{gray}{39.12} & \textcolor{blue}{0.57} & \textcolor{blue}{0.00} & 0.51 & 0.14 \\
\textsc{4x2AntVel} & \textcolor{gray}{1.00} & \textcolor{gray}{38.01} & \textcolor{gray}{1.07} & \textcolor{gray}{34.34} & 0.50 & 0.00 & \textcolor{blue}{0.50} & \textcolor{blue}{0.01} \\
\textsc{2x3HalfCheetahVel} & \textcolor{gray}{1.00} & \textcolor{gray}{39.02} & \textcolor{gray}{1.11} & \textcolor{gray}{37.70} & \textcolor{blue}{0.35} & \textcolor{blue}{0.01} & \textcolor{gray}{0.49} & \textcolor{gray}{1.28} \\
\textsc{6x1HalfCheetahVel} & \textcolor{gray}{1.00} & \textcolor{gray}{39.23} & \textcolor{gray}{1.09} & \textcolor{gray}{37.74} & 0.28 & 0.02 & \textcolor{blue}{0.36} & \textcolor{blue}{0.37} \\
\textsc{3x1HopperVel} & \textcolor{gray}{1.00} & \textcolor{gray}{22.58} & \textcolor{gray}{1.04} & \textcolor{gray}{22.05} & \textcolor{blue}{0.47} & \textcolor{blue}{0.00} & \textcolor{gray}{0.22} & \textcolor{gray}{1.03} \\
\textsc{9|8HumanoidVel} & \textcolor{gray}{1.00} & \textcolor{gray}{6.34} & \textcolor{gray}{2.79} & \textcolor{gray}{17.18} & \textcolor{blue}{0.54} & \textcolor{blue}{0.84} & \textcolor{gray}{0.53} & \textcolor{gray}{1.30} \\
\textsc{2x3Walker2dVel} & \textcolor{gray}{1.00} & \textcolor{gray}{22.99} & \textcolor{gray}{1.55} & \textcolor{gray}{33.67} & \textcolor{blue}{0.60} & \textcolor{blue}{0.01} & \textcolor{gray}{0.27} & \textcolor{gray}{1.21} \\
\bottomrule
\end{tabular}
}
\label{ma-table}
\end{wraptable}

\textbf{Lagrangian vs. Projection.} In contrast to projection-based methods, the Lagrangian-based methods tend to display more oscillation. A notable disparity becomes apparent upon examining the oscillatory patterns in the episodic cost around the designated safety constraints during training, as presented in \autoref{rew-cost-oscil-analysis}(b). Both CPO \cite{achiam2017constrained} and PPO-Lag \cite{ray2019benchmarking} demonstrate oscillations; however, those exhibited by PPO-Lag are more conspicuous. This discrepancy is manifested in a higher proportion of instances classified as \textit{Strongly Unsafe} and \textit{Strongly Safe} for PPO-Lag, while CPO maintains a more centered distribution. Nevertheless, an excessively cautious policy has the potential to undermine performance. In contrast, the projection-based method PCPO \cite{yang2020projection} exhibits lower average costs and rewards in navigation and velocity tasks than CPO. This distinction is further accentuated when examining the contrast between MACPO and MAPPO-Lag.

\textbf{Lagrangian vs. PID-Lagrangian.}
Incorporating a PID controller within the Lagrangian-based framework proves to be effective in mitigating inherent oscillations. As shown in \autoref{rew-cost-oscil-analysis}, CPPO-PID \cite{stooke2020responsive} displays episodic rewards during training that closely resemble those of PPO-Lag. However, CPPO-PID demonstrates a reduced frequency of instances entering the \textit{Strongly Unsafe} region, resulting in a more significant proportion of \textit{Safe} states and improved safety performance.

\section{Limitations and Future Works}

Ensuring safety remains a paramount concern. Across various tasks, safety concerns can be transformed into corresponding constraints. However, a limitation of this study is its inability to encompass all forms of constraints. For instance, safety constraints related to human-centric considerations are paramount in human-AI collaboration, yet these considerations have not been fully integrated within the scope of this study. This work focuses on safety tasks within a simulated environment and introduces an extensive testing component. However, the transferability of the results to complex real-world safety-critical applications may be limited. A promising work for the future involves transferring policy refined within the \texttt{Safety-Gymnasium} to physical robotic platforms, which holds profound implications.

\section*{Acknowledgement}
This work was supported by National Key R\&D Program of China (2022ZD0114900) and by Beijing Municipal Science \& Technology Commission (Project ID: Z231100007423015).

\newpage
\bibliographystyle{unsrt}
\bibliography{ref}

\begin{thebibliography}{10}

\bibitem{carlson2010increasing}
Tom Carlson and Yiannis Demiris.
\newblock Increasing robotic wheelchair safety with collaborative control:
  Evidence from secondary task experiments.
\newblock In {\em 2010 IEEE International Conference on Robotics and
  Automation}, pages 5582--5587. IEEE, 2010.

\bibitem{bi2021safety}
Zhu~Ming Bi, Chaomin Luo, Zhonghua Miao, Bing Zhang, WJ~Zhang, and Lihui Wang.
\newblock Safety assurance mechanisms of collaborative robotic systems in
  manufacturing.
\newblock {\em Robotics and Computer-Integrated Manufacturing}, 67:102022,
  2021.

\bibitem{yang2020projection}
Tsung-Yen Yang, Justinian Rosca, Karthik Narasimhan, and Peter~J Ramadge.
\newblock Projection-based constrained policy optimization.
\newblock {\em arXiv preprint arXiv:2010.03152}, 2020.

\bibitem{Bai_Zhang_Tao_Wu_Wang_Xu_2023}
Fengshuo Bai, Hongming Zhang, Tianyang Tao, Zhiheng Wu, Yanna Wang, and Bo~Xu.
\newblock Picor: Multi-task deep reinforcement learning with policy correction.
\newblock {\em Proceedings of the AAAI Conference on Artificial Intelligence},
  37(6):6728--6736, Jun. 2023.

\bibitem{tabular-cmdp-1989}
Keith~W Ross and Ravi Varadarajan.
\newblock Markov decision processes with sample path constraints: the
  communicating case.
\newblock {\em Operations Research}, 37(5):780--790, 1989.

\bibitem{altman2021constrained}
Eitan Altman.
\newblock {\em Constrained Markov decision processes}.
\newblock Routledge, 2021.

\bibitem{zhang2023evaluating}
Linrui Zhang, Qin Zhang, Li~Shen, Bo~Yuan, Xueqian Wang, and Dacheng Tao.
\newblock Evaluating model-free reinforcement learning toward safety-critical
  tasks.
\newblock In {\em Proceedings of the AAAI Conference on Artificial
  Intelligence}, volume 37(12), pages 15313--15321, 2023.

\bibitem{ji2023omnisafe}
Jiaming Ji, Jiayi Zhou, Borong Zhang, Juntao Dai, Xuehai Pan, Ruiyang Sun,
  Weidong Huang, Yiran Geng, Mickel Liu, and Yaodong Yang.
\newblock Omnisafe: An infrastructure for accelerating safe reinforcement
  learning research, 2023.

\bibitem{ji2024ai}
Jiaming Ji, Tianyi Qiu, Boyuan Chen, Borong Zhang, Hantao Lou, Kaile Wang,
  Yawen Duan, Zhonghao He, Jiayi Zhou, Zhaowei Zhang, Fanzhi Zeng, Kwan~Yee Ng,
  Juntao Dai, Xuehai Pan, Aidan O'Gara, Yingshan Lei, Hua Xu, Brian Tse, Jie
  Fu, Stephen McAleer, Yaodong Yang, Yizhou Wang, Song-Chun Zhu, Yike Guo, and
  Wen Gao.
\newblock Ai alignment: A comprehensive survey, 2024.

\bibitem{feng2023dense}
Shuo Feng, Haowei Sun, Xintao Yan, Haojie Zhu, Zhengxia Zou, Shengyin Shen, and
  Henry~X Liu.
\newblock Dense reinforcement learning for safety validation of autonomous
  vehicles.
\newblock {\em Nature}, 615(7953):620--627, 2023.

\bibitem{ou2023towards}
Yafei Ou and Mahdi Tavakoli.
\newblock Towards safe and efficient reinforcement learning for surgical robots
  using real-time human supervision and demonstration.
\newblock In {\em 2023 International Symposium on Medical Robotics (ISMR)},
  pages 1--7. IEEE, 2023.

\bibitem{ji2023beavertails}
Jiaming Ji, Mickel Liu, Juntao Dai, Xuehai Pan, Chi Zhang, Ce~Bian, Chi Zhang,
  Ruiyang Sun, Yizhou Wang, and Yaodong Yang.
\newblock Beavertails: Towards improved safety alignment of llm via a
  human-preference dataset, 2023.

\bibitem{dai2023safe}
Josef Dai, Xuehai Pan, Ruiyang Sun, Jiaming Ji, Xinbo Xu, Mickel Liu, Yizhou
  Wang, and Yaodong Yang.
\newblock Safe rlhf: Safe reinforcement learning from human feedback, 2023.

\bibitem{brockman2016openai}
Greg Brockman, Vicki Cheung, Ludwig Pettersson, Jonas Schneider, John Schulman,
  Jie Tang, and Wojciech Zaremba.
\newblock Openai gym.
\newblock {\em arXiv preprint arXiv:1606.01540}, 2016.

\bibitem{mnih2013playing}
Volodymyr Mnih, Koray Kavukcuoglu, David Silver, Alex Graves, Ioannis
  Antonoglou, Daan Wierstra, and Martin Riedmiller.
\newblock Playing atari with deep reinforcement learning.
\newblock {\em arXiv preprint arXiv:1312.5602}, 2013.

\bibitem{tunyasuvunakool2020dm_control}
Saran Tunyasuvunakool, Alistair Muldal, Yotam Doron, Siqi Liu, Steven Bohez,
  Josh Merel, Tom Erez, Timothy Lillicrap, Nicolas Heess, and Yuval Tassa.
\newblock dm\_control: Software and tasks for continuous control.
\newblock {\em Software Impacts}, 6:100022, 2020.

\bibitem{leike2017ai}
Jan Leike, Miljan Martic, Victoria Krakovna, Pedro~A Ortega, Tom Everitt,
  Andrew Lefrancq, Laurent Orseau, and Shane Legg.
\newblock Ai safety gridworlds.
\newblock {\em arXiv preprint arXiv:1711.09883}, 2017.

\bibitem{ray2019benchmarking}
Alex Ray, Joshua Achiam, and Dario Amodei.
\newblock Benchmarking safe exploration in deep reinforcement learning.
\newblock {\em arXiv preprint arXiv:1910.01708}, 7(1):2, 2019.

\bibitem{yuan2022safe}
Zhaocong Yuan, Adam~W Hall, Siqi Zhou, Lukas Brunke, Melissa Greeff, Jacopo
  Panerati, and Angela~P Schoellig.
\newblock Safe-control-gym: A unified benchmark suite for safe learning-based
  control and reinforcement learning in robotics.
\newblock {\em IEEE Robotics and Automation Letters}, 7(4):11142--11149, 2022.

\bibitem{li2022metadrive}
Quanyi Li, Zhenghao Peng, Lan Feng, Qihang Zhang, Zhenghai Xue, and Bolei Zhou.
\newblock Metadrive: Composing diverse driving scenarios for generalizable
  reinforcement learning.
\newblock {\em IEEE transactions on pattern analysis and machine intelligence},
  45(3):3461--3475, 2022.

\bibitem{makoviychuk2021isaac}
Viktor Makoviychuk, Lukasz Wawrzyniak, Yunrong Guo, Michelle Lu, Kier Storey,
  Miles Macklin, David Hoeller, Nikita Rudin, Arthur Allshire, Ankur Handa,
  et~al.
\newblock Isaac gym: High performance gpu-based physics simulation for robot
  learning.
\newblock {\em arXiv preprint arXiv:2108.10470}, 2021.

\bibitem{Gymnasium}
Farama Foundation.
\newblock A standard api for single-agent reinforcement learning environments,
  with popular reference environments and related utilities (formerly gym).
\newblock \url{https://github.com/Farama-Foundation/Gymnasium}, 2022.

\bibitem{todorov2012mujoco}
Emanuel Todorov, Tom Erez, and Yuval Tassa.
\newblock Mujoco: A physics engine for model-based control.
\newblock In {\em 2012 IEEE/RSJ international conference on intelligent robots
  and systems}, pages 5026--5033. IEEE, 2012.

\bibitem{xu2022trustworthy}
Mengdi Xu, Zuxin Liu, Peide Huang, Wenhao Ding, Zhepeng Cen, Bo~Li, and Ding
  Zhao.
\newblock Trustworthy reinforcement learning against intrinsic vulnerabilities:
  Robustness, safety, and generalizability.
\newblock {\em arXiv preprint arXiv:2209.08025}, 2022.

\bibitem{gu2022review}
Shangding Gu, Long Yang, Yali Du, Guang Chen, Florian Walter, Jun Wang, Yaodong
  Yang, and Alois Knoll.
\newblock A review of safe reinforcement learning: Methods, theory and
  applications.
\newblock {\em arXiv preprint arXiv:2205.10330}, 2022.

\bibitem{tabular-cmdp-1983}
Lodewijk~CM Kallenberg.
\newblock Linear programming and finite markovian control problems.
\newblock {\em MC Tracts}, 1983.

\bibitem{garcia2015comprehensive}
Javier Garc{\i}a and Fernando Fern{\'a}ndez.
\newblock A comprehensive survey on safe reinforcement learning.
\newblock {\em Journal of Machine Learning Research}, 16(1):1437--1480, 2015.

\bibitem{Dai_Ji_Yang_Zheng_Pan_2023}
Juntao Dai, Jiaming Ji, Long Yang, Qian Zheng, and Gang Pan.
\newblock Augmented proximal policy optimization for safe reinforcement
  learning.
\newblock {\em Proceedings of the AAAI Conference on Artificial Intelligence},
  37(6):7288--7295, Jun. 2023.

\bibitem{huang2023safedreamer}
Weidong Huang, Jiaming Ji, Borong Zhang, Chunhe Xia, and Yaodong Yang.
\newblock Safedreamer: Safe reinforcement learning with world models, 2023.

\bibitem{achiam2017constrained}
Joshua Achiam, David Held, Aviv Tamar, and Pieter Abbeel.
\newblock Constrained policy optimization.
\newblock In {\em International conference on machine learning}, pages 22--31.
  PMLR, 2017.

\bibitem{zhang2020first}
Yiming Zhang, Quan Vuong, and Keith Ross.
\newblock First order constrained optimization in policy space.
\newblock {\em Advances in Neural Information Processing Systems},
  33:15338--15349, 2020.

\bibitem{yang2022cup}
Long Yang, Jiaming Ji, Juntao Dai, Yu~Zhang, Pengfei Li, and Gang Pan.
\newblock Cup: A conservative update policy algorithm for safe reinforcement
  learning, 2022.

\bibitem{sutton2018reinforcement}
Richard~S Sutton and Andrew~G Barto.
\newblock {\em Reinforcement learning: An introduction}.
\newblock MIT press, 2018.

\bibitem{jacobs2002physiological}
Patrick~L Jacobs, Stephen~E Olvey, Brad~M Johnson, and Kelly Cohn.
\newblock Physiological responses to high-speed, open-wheel racecar driving.
\newblock {\em Medicine and science in sports and exercise}, 34(12):2085--2090,
  2002.

\bibitem{betz2019software}
Johannes Betz, Alexander Wischnewski, Alexander Heilmeier, Felix Nobis, Tim
  Stahl, Leonhard Hermansdorfer, and Markus Lienkamp.
\newblock A software architecture for an autonomous racecar.
\newblock In {\em 2019 IEEE 89th Vehicular Technology Conference
  (VTC2019-Spring)}, pages 1--6. IEEE, 2019.

\bibitem{de2020deep}
Christian~Schroeder de~Witt, Bei Peng, Pierre-Alexandre Kamienny, Philip Torr,
  Wendelin B{\"o}hmer, and Shimon Whiteson.
\newblock Deep multi-agent reinforcement learning for decentralized continuous
  cooperative control.
\newblock {\em arXiv preprint arXiv:2003.06709}, 19, 2020.

\bibitem{kuba2021trust}
Jakub~Grudzien Kuba, Ruiqing Chen, Muning Wen, Ying Wen, Fanglei Sun, Jun Wang,
  and Yaodong Yang.
\newblock Trust region policy optimisation in multi-agent reinforcement
  learning.
\newblock {\em arXiv preprint arXiv:2109.11251}, 2021.

\bibitem{yu2022surprising}
Chao Yu, Akash Velu, Eugene Vinitsky, Jiaxuan Gao, Yu~Wang, Alexandre Bayen,
  and Yi~Wu.
\newblock The surprising effectiveness of ppo in cooperative multi-agent games.
\newblock {\em Advances in Neural Information Processing Systems},
  35:24611--24624, 2022.

\bibitem{gu2021multi}
Shangding Gu, Jakub~Grudzien Kuba, Munning Wen, Ruiqing Chen, Ziyan Wang, Zheng
  Tian, Jun Wang, Alois Knoll, and Yaodong Yang.
\newblock Multi-agent constrained policy optimisation.
\newblock {\em arXiv preprint arXiv:2110.02793}, 2021.

\bibitem{ma2022conservative}
Yecheng~Jason Ma, Andrew Shen, Osbert Bastani, and Jayaraman Dinesh.
\newblock Conservative and adaptive penalty for model-based safe reinforcement
  learning.
\newblock In {\em Proceedings of the AAAI conference on artificial
  intelligence}, volume 36(5), pages 5404--5412, 2022.

\bibitem{as2022constrained}
Yarden As, Ilnura Usmanova, Sebastian Curi, and Andreas Krause.
\newblock Constrained policy optimization via bayesian world models.
\newblock {\em arXiv preprint arXiv:2201.09802}, 2022.

\bibitem{chen2022towards}
Yuanpei Chen, Tianhao Wu, Shengjie Wang, Xidong Feng, Jiechuan Jiang, Zongqing
  Lu, Stephen McAleer, Hao Dong, Song-Chun Zhu, and Yaodong Yang.
\newblock Towards human-level bimanual dexterous manipulation with
  reinforcement learning.
\newblock {\em Advances in Neural Information Processing Systems},
  35:5150--5163, 2022.

\bibitem{schulman2017proximal}
John Schulman, Filip Wolski, Prafulla Dhariwal, Alec Radford, and Oleg Klimov.
\newblock Proximal policy optimization algorithms.
\newblock {\em arXiv preprint arXiv:1707.06347}, 2017.

\bibitem{stooke2020responsive}
Adam Stooke, Joshua Achiam, and Pieter Abbeel.
\newblock Responsive safety in reinforcement learning by pid lagrangian
  methods.
\newblock In {\em International Conference on Machine Learning}, pages
  9133--9143. PMLR, 2020.

\bibitem{schulman2015high}
John Schulman, Philipp Moritz, Sergey Levine, Michael Jordan, and Pieter
  Abbeel.
\newblock High-dimensional continuous control using generalized advantage
  estimation.
\newblock {\em arXiv preprint arXiv:1506.02438}, 2015.

\bibitem{kingma2014adam}
Diederik~P Kingma and Jimmy Ba.
\newblock Adam: A method for stochastic optimization.
\newblock {\em arXiv preprint arXiv:1412.6980}, 2014.

\end{thebibliography}
\appendix
\newpage
\section{Details of Experimental Results}
\label{app:experiment_analysis}
\subsection{Hyperparameters Analysis}

This section presents the disclosure of \texttt{SafePO} hyperparameters settings and their rationales. We employed the Generalized Advantage Estimation (\texttt{GAE})\cite{schulman2015high} method to estimate the values of rewards and cost advantages and used Adam\cite{kingma2014adam} for learning the neural network parameters. 

\textbf{Single-agent Algorithm Settings.}
The models employed in the single-agent algorithms were 3-layer MLPs with \texttt{Tanh} activation functions and hidden layer sizes of [64, 64], for more intricate navigation agents Ant and Doggo, hidden layers of [256, 256] were employed. 

\textbf{Multi-agent Algorithms Settings.}
The models employed in the multi-agent algorithms were 3-layer MLPs with \texttt{ReLU} activation functions and hidden layer sizes of [128, 128].

\begin{table}[ht]
\captionsetup{skip=5pt}
\caption{Hyperparameters of \texttt{SafePO} algorithms in \texttt{Safety-Gymnasium} tasks. Second-order algorithms set the parameters to the actor model directly, instead of iterative gradient descent, so the \textit{Actor Learning Rate} of them are marked \textcolor{gray}{Gray}.}
\centering

\resizebox{\columnwidth}{!}{
\begin{subtable}{.5\textwidth}
  \centering
  \begin{tabular}{ll}\toprule
    \textbf{PG/PPO/PPO-Lag} & \textbf{Value} \\
    \midrule
    Discount Factor $\gamma$ & 0.99 \\
    Target KL & 0.02 \\
    GAE $\lambda$ & 0.95 \\
    Number of SGD Iterations & 40 \\
    Training Batch Size & 20000 \\
    Actor Learning Rate & 0.0003 \\
    Critic Learning Rate & 0.0003 \\
    Cost Limit & 25.00 \\
    Clip Coefficient & 0.20\\
    Lagrangian Initial Value & 0.001 \\
    Lagrangian Learning Rate & 0.035 \\
    Lagrangian Optimizer & \texttt{Adam} \\
    \bottomrule
  \end{tabular}
\end{subtable}\hfill 

\begin{subtable}{.5\textwidth}
  \centering
  \begin{tabular}{ll}
    \toprule
    \textbf{TRPO/TRPO-Lag} & \textbf{Value} \\
    \midrule
    Discount Factor $\gamma$ & 0.99 \\
    Target KL & 0.01 \\
    GAE $\lambda$ & 0.95 \\
    Number of SGD Iterations & 10 \\
    Training Batch Size & 20000 \\
    \textcolor{gray}{Actor Learning Rate} & \textcolor{gray}{None} \\
    Critic Learning Rate & 0.001 \\
    Cost Limit & 25.00 \\
    Conjugate Gradient Iterations & 15 \\
    Lagrangian Initial Value & 0.001 \\
    Lagrangian Learning Rate & 0.035 \\
    Lagrangian Optimizer & \texttt{Adam} \\
    \bottomrule
  \end{tabular}
\end{subtable}

\begin{subtable}{.5\textwidth}
  \centering
  \begin{tabular}{ll}
    \toprule
    \textbf{CPPO-PID} & \textbf{Value} \\
    \midrule
    Discount Factor $\gamma$ & 0.99 \\
    Target KL & 0.02 \\
    GAE $\lambda$ & 0.95 \\
    Number of SGD Iterations & 40 \\
    Training Batch Size & 20000 \\
    Actor Learning Rate & 0.0003 \\
    Critic Learning Rate & 0.0003 \\
    Cost Limit & 25.00 \\
    Clip Coefficient & 0.20\\
    PID Controller Kp & 0.10 \\
    PID Controller Ki & 0.01 \\
    PID Controller Kd & 0.01 \\
    \bottomrule
  \end{tabular}
\end{subtable}\hfill 
\begin{subtable}{.5\textwidth}
  \centering
  \begin{tabular}{ll}
    \toprule
    \textbf{NPG/RCPO} & \textbf{Value} \\
    \midrule
    Discount Factor $\gamma$ & 0.99 \\
    Target KL & 0.01 \\
    GAE $\lambda$ & 0.95 \\
    Number of SGD Iterations & 10 \\
    Training Batch Size & 20000 \\
    \textcolor{gray}{Actor Learning Rate} & \textcolor{gray}{None} \\
    Critic Learning Rate & 0.001 \\
    Cost Limit & 25.00 \\
    Conjugate Gradient Iterations & 15 \\
    Lagrangian Initial Value & 0.001 \\
    Lagrangian Learning Rate & 0.035 \\
    Lagrangian Optimizer & \texttt{Adam} \\
    \bottomrule
  \end{tabular}
\end{subtable}
\begin{subtable}{.5\textwidth}
  \centering
  \begin{tabular}{ll}
    \toprule
    \textbf{HAPPO/MAPPO/MAPPO-Lag} & \textbf{Value} \\
    \midrule
    Discount Factor $\gamma$ & 0.99 \\
    Target KL & 0.016 \\
    GAE $\lambda$ & 0.95 \\
    Number of SGD Iterations & 5 \\
    Training Batch Size & 10000 \\
    Actor Learning Rate & 0.0005 \\
    Critic Learning Rate & 0.0005 \\
    Cost Limit & 25.00 \\
    Clip Coefficient & 0.20\\
    Lagrangian Initial Value & 0.00001 \\
    Lagrangian Learning Rate & 0.78 \\
    Lagrangian Optimizer & \texttt{SGD} \\
    \bottomrule
  \end{tabular}
\end{subtable}\hfill 
} 

\smallskip

\centering
\resizebox{\columnwidth}{!}{
\begin{subtable}{.5\textwidth}
  \centering
  \begin{tabular}{ll}
    \toprule
    \textbf{CPO} & \textbf{Value} \\
    \midrule
    Discount Factor $\gamma$ & 0.99 \\
    Target KL & 0.01 \\
    GAE $\lambda$ & 0.95 \\
    Number of SGD Iterations & 10 \\
    Training Batch Size & 20000 \\
    \textcolor{gray}{Actor Learning Rate} & \textcolor{gray}{None} \\
    Critic Learning Rate & 0.001 \\
    Cost Limit & 25.00 \\
    Conjugate Gradient Iterations & 15 \\
    CPO Searching Steps & 15 \\
    Step Fraction & 0.80 \\
    \bottomrule
  \end{tabular}
\end{subtable}\hfill 
\begin{subtable}{.5\textwidth}
  \centering
  \begin{tabular}{ll}
    \toprule
    \textbf{PCPO} & \textbf{Value} \\
    \midrule
    Discount Factor $\gamma$ & 0.99 \\
    Target KL & 0.01 \\
    GAE $\lambda$ & 0.95 \\
    Number of SGD Iterations & 10 \\
    Training Batch Size & 20000 \\
    \textcolor{gray}{Actor Learning Rate} & \textcolor{gray}{None} \\
    Critic Learning Rate & 0.001 \\
    Cost Limit & 25.00 \\
    Conjugate Gradient Iterations & 15 \\
    PCPO Searching Steps & 200 \\
    Step Fraction & 0.80 \\
    \bottomrule
  \end{tabular}
\end{subtable}

\begin{subtable}{.5\textwidth}
  \centering
  \begin{tabular}{ll}
    \toprule
    \textbf{CUP} & \textbf{Value} \\
    \midrule
    Discount Factor $\gamma$ & 0.99 \\
    Target KL & 0.02 \\
    GAE $\lambda$ & 0.95 \\
    Number of SGD Iteration & 40 \\
    Training Batch Size & 20000 \\
    Actor Learning Rate & 0.0003 \\
    Critic Learning Rate & 0.0003 \\
    Cost Limit & 25.00 \\
    Clip Coefficient & 0.20 \\
    CUP $\lambda$ & 0.95 \\
    CUP $\nu$ & 2.00 \\
    \bottomrule
  \end{tabular}
\end{subtable}\hfill 
\begin{subtable}{.5\textwidth}
  \centering
  \begin{tabular}{ll}
    \toprule
    \textbf{FOCOPS} & \textbf{Value} \\
    \midrule
    Discount Factor $\gamma$ & 0.99 \\
    Target KL & 0.02 \\
    GAE $\lambda$ & 0.95 \\
    Number of SGD Iteration & 40 \\
    Training Batch Size & 20000 \\
    Actor Learning Rate & 0.0003 \\
    Critic Learning Rate & 0.0003 \\
    Cost Limit & 25.00 \\
    Clip Coefficient & 0.20 \\
    FOCOPS $\lambda$ & 1.50 \\
    FOCOPS $\nu$ & 2.00 \\
    \bottomrule
  \end{tabular}
\end{subtable}
\begin{subtable}{.5\textwidth}
  \centering
  \begin{tabular}{ll}
    \toprule
    \textbf{MACPO} & \textbf{Value} \\
    \midrule
    Discount Factor $\gamma$ & 0.99 \\
    Target KL & 0.01 \\
    GAE $\lambda$ & 0.95 \\
    Number of SGD Iteration & 15 \\
    Training Batch Size & 10000 \\
    \textcolor{gray}{Actor Learning Rate} & \textcolor{gray}{None} \\
    Critic Learning Rate & 0.0005 \\
    Cost Limit & 25.00 \\
    Conjugate Gradient Iterations & 10 \\
    MACPO Searching Steps & 10 \\
    Step Fraction & 0.50 \\
    \bottomrule
  \end{tabular}
\end{subtable}\hfill 
} 

\end{table}

\textbf{Lagrangian Multiplier Settings.}\label{lag_analysis}
Lagrangian-based methods are sensitive to hyperparameters. We present the following detailed description of the settings for both the naive and the PID-controlled Lagrangian multiplier.
\begin{itemize}
\item Lagrangian Initial Value: The initial value of the Lagrangian multiplier. It impacts the early-stage performance of the Lagrangian-based methods. A higher initial value promotes safer exploration but may impede task completion. Conversely, a lower initial value delays the agent's exploration of safe policies.
\item Lagrangian Learning Rate: The learning rate of the Lagrangian multiplier. A high learning rate induces excessive oscillations, impedes convergence speed, and hinders the algorithm's ability to attain the desired solution. Conversely, a low learning rate slows down convergence and adversely affects training.
\item PID Controller Kp: The PID controller's proportional gain determines the output's response to changes in the episodic costs. If \texttt{pid\_kp} is too large, the Lagrangian multiplier oscillates, and performance deteriorates. If \texttt{pid\_kp} is too small, the Lagrangian multiplier updates slowly, also impacting performance negatively.
\item PID Controller Kd: The PID controller's derivative gain governs the output's response to changes in the episodic costs. If \texttt{pid\_kd} is too large, the Lagrangian multiplier becomes excessively sensitive to noise or changes in the episodic costs, leading to instability or oscillations. If \texttt{pid\_kd} is too small, the Lagrangian multiplier may not respond quickly or accurately enough to changes in the episodic costs.
\item PID Controller Ki: The PID controller's integral gain determines the controller's ability to eliminate the steady-state error by integrating the episodic costs over time. If \texttt{pid\_ki} is too large, the Lagrangian multiplier may become overly responsive to previous errors, adversely affecting performance.
\end{itemize}

\clearpage

\subsection{Performance Table of \texttt{Safety-Gymnasium}}

\begin{table}[htbp]
\caption{The performance of \texttt{SafePO} algorithms on \texttt{Safety-Gymnasium}. All experimental outcomes were
derived from 10 assessment iterations encompassing multiple random seeds and under the experimental setting of \texttt{cost\_limit=25.00}. The $\uparrow$ indicates higher rewards are better, while the $\downarrow$ indicates lower costs (when beyond the threshold of 25.00) are better. \textit{\textcolor{gray}{Gray}} and \textit{Black} depicts breach and compliance with the \texttt{cost\_limit}, while \textit{\textcolor{blue}{Green}} represents the optimal policy, maximizing reward within safety constraints.}
\label{performance-table-appendix}
\begin{subtable}{\linewidth}\centering
\caption{The performance of \texttt{SafePO} single-agent algorithms on \texttt{Safety-Gymnasium}.}
\label{sa-table-appendix}
\resizebox{\columnwidth}{!}{
\begin{tabular}{@{}l|cc|cc|cc|cc|cc|cc|cc|cc|cc@{}}\toprule
    & \multicolumn{2}{c|}{\textbf{PPO}} & \multicolumn{2}{c|}{\textbf{PPO-Lag}} & \multicolumn{2}{c|}{\textbf{CPPO-PID}} & \multicolumn{2}{c|}{\textbf{TRPO-Lag}} & \multicolumn{2}{c|}{\textbf{RCPO}} & \multicolumn{2}{c|}{\textbf{CPO}} & \multicolumn{2}{c|}{\textbf{PCPO}} & \multicolumn{2}{c|}{\textbf{CUP}} & \multicolumn{2}{c}{\textbf{FOCOPS}} \\
\midrule
\textbf{Safety Navigation} & $J^R$ & $J^C$ & $J^R$  & $J^C$  & $J^R$ & $J^C$ & $J^R$ & $J^C$  & $J^R$ & $J^C$ & $J^R$  & $J^C$  & $J^R$ & $J^C$ & $J^R$ & $J^C$  & $J^R$ & $J^C$\\ \midrule
\textsc{AntButton1} & \textcolor{gray}{38.70} & \textcolor{gray}{110.60} & 3.63 & 21.60 & \textcolor{blue}{4.06} & \textcolor{blue}{17.45} & \textcolor{gray}{8.93} & \textcolor{gray}{48.70} & \textcolor{gray}{6.16} & \textcolor{gray}{51.70} & \textcolor{gray}{4.50} & \textcolor{gray}{100.30} & \textcolor{gray}{1.27} & \textcolor{gray}{25.35} & 1.26 & 4.25 & 0.22 & 11.55 \\
\textsc{AntButton2} & \textcolor{gray}{36.15} & \textcolor{gray}{95.00} & \textcolor{blue}{2.72} & \textcolor{blue}{14.85} & \textcolor{gray}{2.86} & \textcolor{gray}{28.70} & \textcolor{gray}{8.66} & \textcolor{gray}{49.45} & \textcolor{gray}{8.66} & \textcolor{gray}{37.40} & \textcolor{gray}{4.63} & \textcolor{gray}{35.60} & \textcolor{gray}{3.04} & \textcolor{gray}{27.50} & \textcolor{gray}{1.60} & \textcolor{gray}{32.90} & -0.04 & 6.80 \\
\textsc{AntCircle1} & \textcolor{gray}{94.04} & \textcolor{gray}{420.30} & \textcolor{gray}{74.31} & \textcolor{gray}{63.90} & \textcolor{gray}{64.90} & \textcolor{gray}{47.50} & \textcolor{gray}{61.02} & \textcolor{gray}{26.30} & \textcolor{gray}{59.42} & \textcolor{gray}{26.00} & \textcolor{gray}{43.74} & \textcolor{gray}{26.80} & \textcolor{gray}{26.47} & \textcolor{gray}{46.85} & \textcolor{blue}{56.77} & \textcolor{blue}{20.50} & \textcolor{gray}{2.27} & \textcolor{gray}{30.50} \\
\textsc{AntCircle2} & \textcolor{gray}{84.80} & \textcolor{gray}{736.00} & 65.72 & 22.45 & \textcolor{gray}{64.49} & \textcolor{gray}{39.85} & \textcolor{blue}{66.75} & \textcolor{blue}{22.75} & 63.04 & 19.00 & \textcolor{gray}{53.74} & \textcolor{gray}{43.90} & 16.41 & 15.85 & 42.65 & 10.80 & \textcolor{gray}{4.78} & \textcolor{gray}{66.30} \\
\textsc{AntGoal1} & \textcolor{gray}{82.02} & \textcolor{gray}{45.30} & 21.33 & 23.60 & \textcolor{gray}{38.79} & \textcolor{gray}{48.55} & 20.64 & 18.50 & \textcolor{blue}{23.38} & \textcolor{blue}{19.60} & 15.35 & 13.80 & 7.31 & 10.50 & \textcolor{gray}{27.98} & \textcolor{gray}{33.25} & 6.99 & 16.75 \\
\textsc{AntGoal2} & \textcolor{gray}{86.14} & \textcolor{gray}{165.60} & 1.01 & 0.00 & 0.10 & 0.00 & \textcolor{blue}{4.44} & \textcolor{blue}{13.45} & \textcolor{gray}{6.27} & \textcolor{gray}{54.00} & 0.85 & 4.60 & 0.02 & 0.00 & 0.76 & 1.15 & 0.08 & 1.15 \\
\textsc{AntPush1} & \textcolor{gray}{0.46} & \textcolor{gray}{47.55} & 0.06 & 0.00 & 0.06 & 0.00 & 0.14 & 0.00 & \textcolor{blue}{0.15} & \textcolor{blue}{0.00} & 0.08 & 0.00 & 0.03 & 0.00 & 0.09 & 0.00 & -0.14 & 0.70 \\
\textsc{AntPush2} & \textcolor{gray}{0.77} & \textcolor{gray}{139.20} & 0.01 & 0.02 & 0.02 & 0.00 & 0.01 & 0.00 & \textcolor{blue}{0.10} & \textcolor{blue}{0.00} & 0.05 & 0.00 & 0.02 & 0.00 & 0.02 & 0.10 & 0.07 & 0.20 \\
\textsc{CarButton1} & \textcolor{gray}{15.74} & \textcolor{gray}{398.81} & \textcolor{blue}{0.11} & \textcolor{blue}{11.87} & -1.70 & 10.03 & \textcolor{gray}{-0.66} & \textcolor{gray}{26.90} & \textcolor{gray}{-3.16} & \textcolor{gray}{43.20} & \textcolor{gray}{1.30} & \textcolor{gray}{43.73} & \textcolor{gray}{0.27} & \textcolor{gray}{47.60} & \textcolor{gray}{0.68} & \textcolor{gray}{137.47} & \textcolor{gray}{0.60} & \textcolor{gray}{30.23} \\
\textsc{CarButton2} & \textcolor{gray}{19.32} & \textcolor{gray}{333.82} & \textcolor{gray}{1.23} & \textcolor{gray}{46.14} & \textcolor{gray}{-1.83} & \textcolor{gray}{26.55} & \textcolor{blue}{-2.23} & \textcolor{blue}{17.98} & \textcolor{gray}{-0.02} & \textcolor{gray}{27.09} & \textcolor{gray}{-0.10} & \textcolor{gray}{36.97} & \textcolor{gray}{0.49} & \textcolor{gray}{38.54} & \textcolor{gray}{0.80} & \textcolor{gray}{154.50} & \textcolor{gray}{0.07} & \textcolor{gray}{53.49} \\
\textsc{CarCircle1} & \textcolor{gray}{21.92} & \textcolor{gray}{208.73} & \textcolor{blue}{17.91} & \textcolor{blue}{20.62} & \textcolor{gray}{35.71} & \textcolor{gray}{44.87} & \textcolor{gray}{37.42} & \textcolor{gray}{69.30} & \textcolor{gray}{37.78} & \textcolor{gray}{77.77} & \textcolor{gray}{37.10} & \textcolor{gray}{78.23} & \textcolor{gray}{31.37} & \textcolor{gray}{49.80} & \textcolor{gray}{16.89} & \textcolor{gray}{25.88} & \textcolor{gray}{18.63} & \textcolor{gray}{27.98} \\
\textsc{CarCircle2} & \textcolor{gray}{19.75} & \textcolor{gray}{401.83} & \textcolor{gray}{16.27} & \textcolor{gray}{29.88} & \textcolor{gray}{30.80} & \textcolor{gray}{40.37} & \textcolor{gray}{33.23} & \textcolor{gray}{54.20} & \textcolor{gray}{33.74} & \textcolor{gray}{42.17} & \textcolor{gray}{33.42} & \textcolor{gray}{78.97} & \textcolor{gray}{27.93} & \textcolor{gray}{70.40} & \textcolor{blue}{14.74} & \textcolor{blue}{15.46} & \textcolor{gray}{15.60} & \textcolor{gray}{31.20} \\
\textsc{CarGoal1} & \textcolor{gray}{32.57} & \textcolor{gray}{58.91} & 14.57 & 9.84 & \textcolor{gray}{1.00} & \textcolor{gray}{61.71} & \textcolor{gray}{27.49} & \textcolor{gray}{27.28} & \textcolor{blue}{18.49} & \textcolor{blue}{21.45} & \textcolor{gray}{26.23} & \textcolor{gray}{40.71} & \textcolor{gray}{20.64} & \textcolor{gray}{35.41} & 6.38 & 15.67 & 17.58 & 23.22 \\
\textsc{CarGoal2} & \textcolor{gray}{31.59} & \textcolor{gray}{215.74} & 0.59 & 16.81 & 0.12 & 23.09 & \textcolor{gray}{3.27} & \textcolor{gray}{47.18} & \textcolor{gray}{2.61} & \textcolor{gray}{25.45} & \textcolor{gray}{3.55} & \textcolor{gray}{32.63} & \textcolor{gray}{1.83} & \textcolor{gray}{57.82} & \textcolor{gray}{2.45} & \textcolor{gray}{125.80} & \textcolor{blue}{3.28} & \textcolor{blue}{23.01} \\
\textsc{CarPush1} & \textcolor{gray}{1.13} & \textcolor{gray}{181.04} & 0.49 & 19.60 & 0.03 & 11.83 & \textcolor{blue}{1.48} & \textcolor{blue}{17.60} & \textcolor{gray}{1.19} & \textcolor{gray}{35.50} & \textcolor{gray}{0.89} & \textcolor{gray}{28.50} & \textcolor{gray}{0.68} & \textcolor{gray}{59.03} & 0.34 & 23.86 & 0.31 & 8.96 \\
\textsc{CarPush2} & \textcolor{gray}{1.03} & \textcolor{gray}{46.87} & \textcolor{gray}{0.54} & \textcolor{gray}{43.32} & \textcolor{gray}{0.57} & \textcolor{gray}{37.37} & \textcolor{gray}{0.43} & \textcolor{gray}{38.63} & \textcolor{gray}{0.12} & \textcolor{gray}{27.57} & \textcolor{blue}{0.15} & \textcolor{blue}{19.03} & \textcolor{gray}{0.29} & \textcolor{gray}{60.10} & \textcolor{gray}{0.41} & \textcolor{gray}{82.20} & \textcolor{gray}{-0.28} & \textcolor{gray}{40.42} \\
\textsc{DoggoButton1} & \textcolor{gray}{27.23} & \textcolor{gray}{189.30} & 0.33 & 0.80 & 0.22 & 1.67 & \textcolor{gray}{0.01} & \textcolor{gray}{31.75} & 0.30 & 2.25 & 0.03 & 3.70 & -0.06 & 6.20 & \textcolor{blue}{0.67} & \textcolor{blue}{11.17} & \textcolor{gray}{1.52} & \textcolor{gray}{91.90} \\
\textsc{DoggoButton2} & \textcolor{gray}{29.84} & \textcolor{gray}{194.60} & 0.10 & 1.00 & 0.16 & 2.70 & -0.05 & 17.05 & 0.07 & 0.00 & 0.03 & 1.40 & 0.01 & 8.01 & \textcolor{gray}{0.35} & \textcolor{gray}{43.37} & \textcolor{blue}{0.22} & \textcolor{blue}{2.10} \\
\textsc{DoggoCircle2} & \textcolor{gray}{41.90} & \textcolor{gray}{442.70} & \textcolor{blue}{30.13} & \textcolor{blue}{14.20} & \textcolor{gray}{34.82} & \textcolor{gray}{62.03} & \textcolor{gray}{21.97} & \textcolor{gray}{46.75} & \textcolor{gray}{20.68} & \textcolor{gray}{37.35} & \textcolor{gray}{20.41} & \textcolor{gray}{32.55} & 15.41 & 24.05 & \textcolor{gray}{33.08} & \textcolor{gray}{58.33} & \textcolor{gray}{28.91} & \textcolor{gray}{122.80} \\
\textsc{DoggoCircle1} & \textcolor{gray}{41.61} & \textcolor{gray}{828.50} & \textcolor{blue}{32.03} & \textcolor{blue}{11.50} & \textcolor{gray}{34.26} & \textcolor{gray}{53.93} & \textcolor{gray}{27.86} & \textcolor{gray}{34.20} & \textcolor{gray}{22.93} & \textcolor{gray}{32.90} & \textcolor{gray}{27.65} & \textcolor{gray}{30.55} & 12.94 & 13.70 & \textcolor{gray}{33.45} & \textcolor{gray}{50.97} & \textcolor{gray}{30.29} & \textcolor{gray}{112.20} \\
\textsc{DoggoGoal1} & \textcolor{gray}{43.10} & \textcolor{gray}{57.10} & 2.00 & 0.00 & 0.13 & 0.00 & 7.88 & 17.25 & \textcolor{gray}{6.82} & \textcolor{gray}{52.05} & \textcolor{blue}{12.73} & \textcolor{blue}{12.40} & 0.14 & 0.00 & 0.16 & 22.47 & \textcolor{gray}{1.88} & \textcolor{gray}{31.80} \\
\textsc{DoggoGoal2} & \textcolor{gray}{42.04} & \textcolor{gray}{123.30} & 0.06 & 0.00 & 0.09 & 0.00 & 0.02 & 0.00 & 0.06 & 0.00 & 0.03 & 0.00 & 0.06 & 0.00 & \textcolor{blue}{0.28} & \textcolor{blue}{3.33} & 0.08 & 0.00 \\
\textsc{DoggoPush2} & \textcolor{gray}{0.82} & \textcolor{gray}{32.70} & -0.02 & 0.00 & 0.08 & 0.00 & 0.16 & 0.00 & 0.18 & 0.00 & \textcolor{gray}{0.54} & \textcolor{gray}{39.08} & 0.14 & 0.00 & \textcolor{gray}{0.22} & \textcolor{gray}{52.70} & \textcolor{blue}{0.52} & \textcolor{blue}{0.00} \\
\textsc{DoggoPush1} & \textcolor{gray}{0.90} & \textcolor{gray}{32.70} & 0.08 & 0.00 & 0.29 & 11.03 & \textcolor{blue}{0.48} & \textcolor{blue}{19.40} & \textcolor{gray}{0.49} & \textcolor{gray}{38.80} & 0.41 & 0.00 & 0.32 & 0.00 & 0.27 & 17.10 & \textcolor{gray}{0.58} & \textcolor{gray}{85.10} \\
\textsc{PointButton1} & \textcolor{gray}{26.10} & \textcolor{gray}{151.38} & \textcolor{gray}{5.83} & \textcolor{gray}{32.98} & \textcolor{blue}{-0.12} & \textcolor{blue}{20.88} & \textcolor{gray}{7.13} & \textcolor{gray}{32.31} & \textcolor{gray}{3.01} & \textcolor{gray}{28.14} & \textcolor{gray}{3.20} & \textcolor{gray}{40.16} & \textcolor{gray}{2.18} & \textcolor{gray}{54.74} & \textcolor{gray}{4.70} & \textcolor{gray}{31.39} & \textcolor{gray}{6.60} & \textcolor{gray}{38.27} \\
\textsc{PointButton2} & \textcolor{gray}{27.96} & \textcolor{gray}{166.74} & \textcolor{gray}{0.27} & \textcolor{gray}{31.49} & \textcolor{gray}{0.44} & \textcolor{gray}{30.87} & \textcolor{blue}{4.87} & \textcolor{blue}{24.94} & \textcolor{gray}{7.90} & \textcolor{gray}{53.82} & \textcolor{gray}{5.58} & \textcolor{gray}{47.68} & \textcolor{gray}{1.12} & \textcolor{gray}{41.49} & \textcolor{gray}{3.52} & \textcolor{gray}{61.98} & \textcolor{gray}{1.29} & \textcolor{gray}{26.13} \\
\textsc{PointCircle1} & \textcolor{gray}{54.57} & \textcolor{gray}{202.54} & \textcolor{blue}{47.00} & \textcolor{blue}{23.28} & \textcolor{gray}{93.84} & \textcolor{gray}{52.23} & \textcolor{gray}{90.87} & \textcolor{gray}{33.83} & \textcolor{gray}{90.65} & \textcolor{gray}{35.53} & \textcolor{gray}{92.10} & \textcolor{gray}{43.50} & \textcolor{gray}{72.81} & \textcolor{gray}{56.53} & 44.98 & 15.50 & 46.06 & 22.36 \\
\textsc{PointCircle2} & \textcolor{gray}{54.39} & \textcolor{gray}{397.54} & 41.60 & 19.92 & \textcolor{gray}{83.67} & \textcolor{gray}{45.27} & 82.62 & 6.63 & 83.39 & 7.40 & \textcolor{blue}{85.22} & \textcolor{blue}{21.20} & 79.22 & 22.67 & \textcolor{gray}{41.45} & \textcolor{gray}{30.98} & 42.38 & 20.96 \\
\textsc{PointGoal1} & \textcolor{gray}{26.32} & \textcolor{gray}{48.20} & \textcolor{gray}{12.46} & \textcolor{gray}{37.62} & \textcolor{gray}{8.15} & \textcolor{gray}{26.31} & \textcolor{blue}{18.99} & \textcolor{blue}{22.87} & 13.90 & 24.66 & \textcolor{gray}{20.52} & \textcolor{gray}{27.44} & 18.79 & 20.48 & 11.99 & 18.15 & \textcolor{gray}{14.77} & \textcolor{gray}{32.95} \\
\textsc{PointGoal2} & \textcolor{gray}{26.43} & \textcolor{gray}{159.28} & \textcolor{gray}{0.59} & \textcolor{gray}{59.43} & \textcolor{gray}{-0.56} & \textcolor{gray}{60.37} & \textcolor{gray}{4.18} & \textcolor{gray}{26.80} & \textcolor{gray}{1.84} & \textcolor{gray}{29.19} & \textcolor{gray}{2.65} & \textcolor{gray}{42.40} & \textcolor{gray}{1.32} & \textcolor{gray}{37.66} & \textcolor{gray}{1.00} & \textcolor{gray}{162.97} & \textcolor{blue}{2.71} & \textcolor{blue}{18.63} \\
\textsc{PointPush1} & \textcolor{gray}{0.82} & \textcolor{gray}{57.80} & \textcolor{gray}{0.80} & \textcolor{gray}{33.18} & 0.29 & 8.87 & 0.70 & 24.93 & \textcolor{blue}{4.35} & \textcolor{blue}{23.47} & 1.82 & 19.90 & \textcolor{gray}{1.41} & \textcolor{gray}{31.33} & 1.90 & 19.98 & \textcolor{gray}{0.93} & \textcolor{gray}{62.64} \\
\textsc{PointPush2} & \textcolor{gray}{1.39} & \textcolor{gray}{42.82} & \textcolor{gray}{0.52} & \textcolor{gray}{25.90} & \textcolor{blue}{1.01} & \textcolor{blue}{25.87} & \textcolor{gray}{1.05} & \textcolor{gray}{56.07} & \textcolor{gray}{0.54} & \textcolor{gray}{29.83} & \textcolor{gray}{1.50} & \textcolor{gray}{29.17} & \textcolor{gray}{0.59} & \textcolor{gray}{27.57} & \textcolor{gray}{1.26} & \textcolor{gray}{56.08} & \textcolor{gray}{0.44} & \textcolor{gray}{39.24} \\
\textsc{RacecarButton1} & \textcolor{gray}{8.48} & \textcolor{gray}{343.15} & \textcolor{gray}{-0.05} & \textcolor{gray}{48.55} & \textcolor{gray}{-1.37} & \textcolor{gray}{51.57} & \textcolor{gray}{-0.18} & \textcolor{gray}{44.25} & \textcolor{gray}{-0.63} & \textcolor{gray}{29.70} & \textcolor{gray}{0.02} & \textcolor{gray}{60.95} & \textcolor{gray}{0.13} & \textcolor{gray}{45.45} & \textcolor{gray}{0.04} & \textcolor{gray}{130.63} & \textcolor{gray}{-0.88} & \textcolor{gray}{84.20} \\
\textsc{RacecarButton2} & \textcolor{gray}{5.77} & \textcolor{gray}{284.15} & -0.58 & 22.35 & \textcolor{gray}{-0.64} & \textcolor{gray}{31.80} & \textcolor{gray}{0.19} & \textcolor{gray}{65.00} & \textcolor{blue}{0.38} & \textcolor{blue}{18.45} & \textcolor{gray}{0.01} & \textcolor{gray}{32.90} & \textcolor{gray}{0.04} & \textcolor{gray}{51.95} & \textcolor{gray}{-0.40} & \textcolor{gray}{72.57} & \textcolor{gray}{-0.40} & \textcolor{gray}{57.65} \\
\textsc{RacecarCircle1} & \textcolor{gray}{81.62} & \textcolor{gray}{396.80} & \textcolor{gray}{67.49} & \textcolor{gray}{47.55} & \textcolor{gray}{47.66} & \textcolor{gray}{33.13} & \textcolor{gray}{65.54} & \textcolor{gray}{54.55} & \textcolor{gray}{67.39} & \textcolor{gray}{51.75} & \textcolor{blue}{64.77} & \textcolor{blue}{20.20} & \textcolor{gray}{18.05} & \textcolor{gray}{71.65} & \textcolor{gray}{60.68} & \textcolor{gray}{88.33} & \textcolor{gray}{62.77} & \textcolor{gray}{52.85} \\
\textsc{RacecarCircle2} & \textcolor{gray}{82.61} & \textcolor{gray}{831.00} & \textcolor{gray}{46.85} & \textcolor{gray}{26.05} & \textcolor{gray}{28.04} & \textcolor{gray}{47.37} & \textcolor{gray}{60.83} & \textcolor{gray}{45.65} & \textcolor{gray}{61.40} & \textcolor{gray}{33.00} & \textcolor{gray}{59.17} & \textcolor{gray}{48.30} & \textcolor{gray}{8.81} & \textcolor{gray}{35.05} & \textcolor{blue}{41.50} & \textcolor{blue}{16.13} & \textcolor{gray}{52.38} & \textcolor{gray}{35.10} \\
\textsc{RacecarGoal1} & \textcolor{gray}{11.29} & \textcolor{gray}{106.40} & 2.90 & 12.70 & \textcolor{gray}{-0.42} & \textcolor{gray}{26.87} & \textcolor{blue}{13.40} & \textcolor{blue}{19.20} & 9.89 & 20.70 & \textcolor{gray}{13.30} & \textcolor{gray}{64.50} & 3.72 & 5.90 & \textcolor{gray}{1.47} & \textcolor{gray}{30.57} & 3.47 & 15.40 \\
\textsc{RacecarGoal2} & \textcolor{gray}{9.61} & \textcolor{gray}{158.25} & \textcolor{gray}{0.08} & \textcolor{gray}{54.40} & \textcolor{gray}{-0.85} & \textcolor{gray}{30.50} & 0.40 & 14.30 & \textcolor{blue}{0.55} & \textcolor{blue}{16.80} & \textcolor{gray}{1.19} & \textcolor{gray}{109.85} & \textcolor{gray}{0.69} & \textcolor{gray}{41.90} & \textcolor{gray}{-0.09} & \textcolor{gray}{62.33} & \textcolor{gray}{0.17} & \textcolor{gray}{93.05} \\
\textsc{RacecarPush1} & \textcolor{gray}{0.50} & \textcolor{gray}{58.45} & -0.20 & 0.00 & \textcolor{gray}{-0.42} & \textcolor{gray}{71.83} & \textcolor{gray}{0.37} & \textcolor{gray}{44.75} & \textcolor{gray}{0.29} & \textcolor{gray}{48.00} & \textcolor{blue}{0.47} & \textcolor{blue}{3.30} & -0.08 & 4.50 & \textcolor{gray}{-0.03} & \textcolor{gray}{94.70} & \textcolor{gray}{0.15} & \textcolor{gray}{51.00} \\
\textsc{RacecarPush2} & \textcolor{gray}{0.58} & \textcolor{gray}{213.95} & \textcolor{gray}{0.37} & \textcolor{gray}{43.85} & -0.08 & 24.07 & -0.12 & 5.50 & -0.03 & 0.00 & \textcolor{blue}{0.23} & \textcolor{blue}{9.55} & \textcolor{gray}{-0.51} & \textcolor{gray}{49.75} & \textcolor{gray}{-1.54} & \textcolor{gray}{101.50} & \textcolor{gray}{-0.54} & \textcolor{gray}{56.00}
\\\midrule
\textbf{Safety Velocity} \hfill  & $J^R$ & $J^C$ & $J^R$  & $J^C$  & $J^R$ & $J^C$ & $J^R$ & $J^C$  & $J^R$ & $J^C$ & $J^R$  & $J^C$  & $J^R$ & $J^C$ & $J^R$ & $J^C$  & $J^R$ & $J^C$ \\ \midrule
\textsc{AntVel} & \textcolor{gray}{5899.64} & \textcolor{gray}{943.57} & 3221.90 & 5.43 & 3070.67 & 10.23 & 3157.40 & 3.63 & 3087.03 & 14.12 & 3116.77 & 14.10 & 2276.19 & 10.18 & \textcolor{blue}{3297.29} & \textcolor{blue}{23.56} & 3291.30 & 15.07 \\
\textsc{HalfCheetahVel} & \textcolor{gray}{7013.92} & \textcolor{gray}{933.18} & 3025.42 & 0.00 & \textcolor{blue}{3336.80} & \textcolor{blue}{1.09} & \textcolor{gray}{2952.08} & \textcolor{gray}{25.23} & 2520.50 & 13.95 & 2738.36 & 5.68 & 1743.71 & 15.64 & 2765.42 & 4.28 & 2873.14 & 2.88 \\
\textsc{HopperVel} & \textcolor{gray}{2378.23} & \textcolor{gray}{543.14} & 1347.98 & 22.30 & 1709.13 & 11.11 & 1377.89 & 17.67 & 1355.69 & 14.85 & 1713.22 & 12.12 & 1519.59 & 12.79 & \textcolor{blue}{1716.35} & \textcolor{blue}{5.37} & 1538.79 & 7.43 \\
\textsc{HumanoidVel} & \textcolor{gray}{9117.61} & \textcolor{gray}{959.76} & 6586.70 & 18.95 & \textcolor{blue}{6620.69} & \textcolor{blue}{0.00} & \textcolor{gray}{6552.06} & \textcolor{gray}{59.85} & 6236.18 & 20.57 & 6486.40 & 0.22 & 5863.98 & 0.18 & 6181.80 & 19.88 & 6502.90 & 23.23 \\
\textsc{SwimmerVel} & \textcolor{gray}{121.23} & \textcolor{gray}{171.21} & \textcolor{gray}{68.10} & \textcolor{gray}{27.68} & \textcolor{blue}{109.34} & \textcolor{blue}{22.92} & 79.63 & 20.98 & 64.73 & 22.56 & 61.49 & 20.46 & 60.48 & 17.31 & 70.86 & 23.93 & \textcolor{gray}{55.87} & \textcolor{gray}{32.62} \\
\textsc{Walker2dVel} & \textcolor{gray}{6312.27} & \textcolor{gray}{899.82} & 2756.61 & 4.90 & 1704.06 & 8.90 & \textcolor{blue}{3209.78} & \textcolor{blue}{19.18} & 3072.07 & 3.72 & 2440.82 & 20.15 & 1698.31 & 17.73 & 2739.50 & 4.39 & 3116.08 & 3.93\\
\bottomrule
\end{tabular}
}
\end{subtable}

\begin{subtable}{\linewidth}\centering
\caption{The performance of \texttt{SafePO} multi-agent algorithms on \texttt{Safety-Gymnasium}.}
\resizebox{0.5\columnwidth}{!}{
\begin{tabular}{@{}l|cc|cc|cc|cc@{}}\toprule
    & \multicolumn{2}{c|}{\textbf{MAPPO}} & \multicolumn{2}{c|}{\textbf{HAPPO}} & \multicolumn{2}{c|}{\textbf{MAPPO-Lag}} & \multicolumn{2}{c}{\textbf{MACPO}} \\
\midrule
\textbf{Safety Velocity} \hfill  & $J^R$ & $J^C$ & $J^R$  & $J^C$  & $J^R$ & $J^C$ & $J^R$ & $J^C$  \\ \midrule
\textsc{2x4AntVel} & \textcolor{gray}{4259.52} & \textcolor{gray}{894.06} & \textcolor{gray}{5368.61} & \textcolor{gray}{978.06} & \textcolor{blue}{2423.47} & \textcolor{blue}{0.00} & 2169.23 & \textbf{3.39} \\
\textsc{4x2AntVel} & \textcolor{gray}{4309.05} & \textcolor{gray}{950.33} & \textcolor{gray}{4613.69} & \textcolor{gray}{858.50} & 2171.40 & 0.00 & \textcolor{blue}{2172.31} & \textcolor{blue}{0.17} \\
\textsc{2x3HalfCheetahVel} & \textcolor{gray}{5057.63} & \textcolor{gray}{975.50} & \textcolor{gray}{5605.98} & \textcolor{gray}{942.56} & \textcolor{blue}{1750.96} & \textcolor{blue}{0.33} & \textcolor{gray}{2470.29} & \textcolor{gray}{32.06} \\
\textsc{6x1HalfCheetahVel} & \textcolor{gray}{5061.53} & \textcolor{gray}{980.67} & \textcolor{gray}{5540.57} & \textcolor{gray}{943.56} & 1439.38 & 0.61 & \textcolor{blue}{1830.65} & \textcolor{blue}{9.33} \\
\textsc{3x1HopperVel} & \textcolor{gray}{2115.35} & \textcolor{gray}{564.56} & \textcolor{gray}{2207.50} & \textcolor{gray}{551.33} & \textcolor{blue}{1002.01} & \textcolor{blue}{0.00} & \textcolor{gray}{461.25} & \textcolor{gray}{25.78} \\
\textsc{9|8HumanoidVel} & \textcolor{gray}{974.50} & \textcolor{gray}{158.61} & \textcolor{gray}{2718.48} & \textcolor{gray}{429.61} & \textcolor{blue}{526.69} & \textcolor{blue}{21.00} & \textcolor{gray}{512.29} & \textcolor{gray}{32.50} \\
\textsc{2x1SwimmerVel} & \textcolor{gray}{39.88} & \textcolor{gray}{101.89} & \textcolor{gray}{51.95} & \textcolor{gray}{267.00} & \textcolor{gray}{27.89} & \textcolor{gray}{59.73} & \textcolor{blue}{-4.02} & \textcolor{blue}{20.83} \\
\textsc{2x3Walker2dVel} & \textcolor{gray}{2691.41} & \textcolor{gray}{574.72} & \textcolor{gray}{4183.34} & \textcolor{gray}{841.83} & \textcolor{blue}{1618.98} & \textcolor{blue}{0.33} & \textcolor{gray}{714.18} & \textcolor{gray}{30.22} \\
\bottomrule
\end{tabular}
}
\label{ma-table-appendix}
\end{subtable}
\end{table}

\textbf{Experimental Results Analysis.}

During the observation of the experimental results, we have discovered some Insightful findings that are presented below:

\begin{itemize}
    \item The Lagrangian method is a promising yet constrained baseline approach, successfully optimizing rewards while adhering to constraints. However, its effectiveness heavily relies on hyperparameters configuration, as discussed in \autoref{lag_analysis}. Consequently, despite being a dependable baseline, the Lagrangian method is not exempt from limitations.
    \item Second-order algorithms perform worse in achieving higher rewards in the MuJoCo velocity series but better in navigation series tasks that require higher safety standards, \textit{i.e.}, achieving similar or approximate rewards while minimizing the number and smoothness of cost violations.
\end{itemize}


\newpage
\section{Details Documentation of Gymnasium-based Learning Environments}
\label{app:environment}
\subsection{Single-agent Specification}

\begin{figure}[H]
  \centering
  \includegraphics[width=0.6\linewidth]{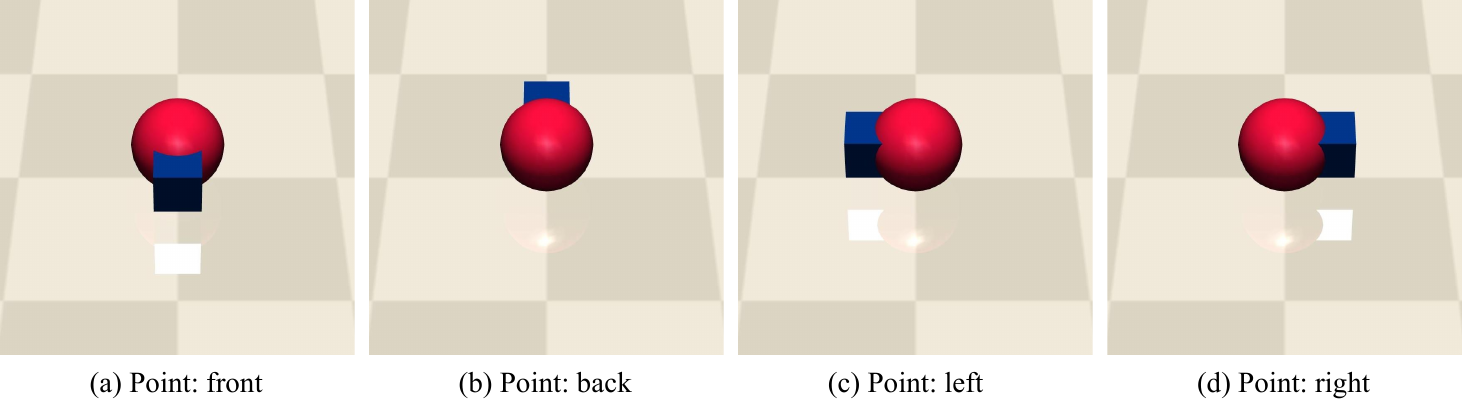}
  \caption{A different view of the robot: Point.}
  \label{pic: point}
\end{figure}
\begin{table}[H]
\caption{The overall information of Point}
\label{pic:overall_point}
\centering
\resizebox{0.5\textwidth}{!}{
\begin{tabular}{lc}
\hline
Specific Action Space      & Box(-1.0, 1.0, (2,), float64) \\ \hline
Specific Observation Space & (12, )                        \\ \hline
Observation High           & inf                           \\ \hline
Observation Low            & -inf                          \\ \hline
\end{tabular}
}
\end{table}

\begin{table}[H]
\centering
\caption{The specific observation space of Point}
\label{pic:obs_spcae_point}
\resizebox{0.85\textwidth}{!}{
\begin{tabular}{ccccccc}
\hline
Size & Observation   & Min  & Max & Name (in XML file) & Joint/Site & Unit                                  \\ \hline
3    & accelerometer & -inf & inf & accelerometer      & site       & acceleration (m/s\textasciicircum{}2) \\ \hline
3    & velocimeter   & -inf & inf & velocimeter        & site       & velocity (m/s)                        \\ \hline
3    & gyro          & -inf & inf & gyro               & site       & anglular velocity (rad/s)             \\ \hline
3    & magnetometer  & -inf & inf & magnetometer       & site       & magnetic flux (Wb)                    \\ \hline
\end{tabular}
}
\end{table}

\begin{table}[H]
\centering
\caption{The specific action space of Point}
\label{pic:action_space_point}
\resizebox{\textwidth}{!}{
\begin{tabular}{ccccccc}
\hline
Num & Action                                                                                           & Control Min & Control Max & Name (in XML file) & Joint/Site & Unit           \\ \hline
0   & \begin{tabular}[c]{@{}c@{}}force applied on the agent\\ to move forward or backward\end{tabular} & -1           & 1           & x                  & site       & force (N)      \\ \hline
1   & \begin{tabular}[c]{@{}c@{}}velocity of the agent, \\ which is around the z-axis\end{tabular}     & -1           & 1           & z                  & hinge      & velocity (m/s) \\ \hline
\end{tabular}
}
\end{table}
\textbf{Point:} As shown in \autoref{pic: point}, \texttt{Point} operating within a 2D plane is equipped with two distinct actuators: one for rotation and another for forward/backward movement. This decomposed control system greatly facilitates the navigation of the robot. Moreover, there is a small square positioned in front of the robot, aiding in the visual identification of its orientation. Additionally, this square plays a crucial role in assisting the robot, named Point, to effectively push any boxes encountered during its tasks. The overall information of \texttt{Point}, the specific action and observation space of \texttt{Point} is shown in \autoref{pic:overall_point}, \autoref{pic:action_space_point}, \autoref{pic:obs_spcae_point}.

\begin{figure}[H]
  \centering
  \includegraphics[width=0.6\linewidth]{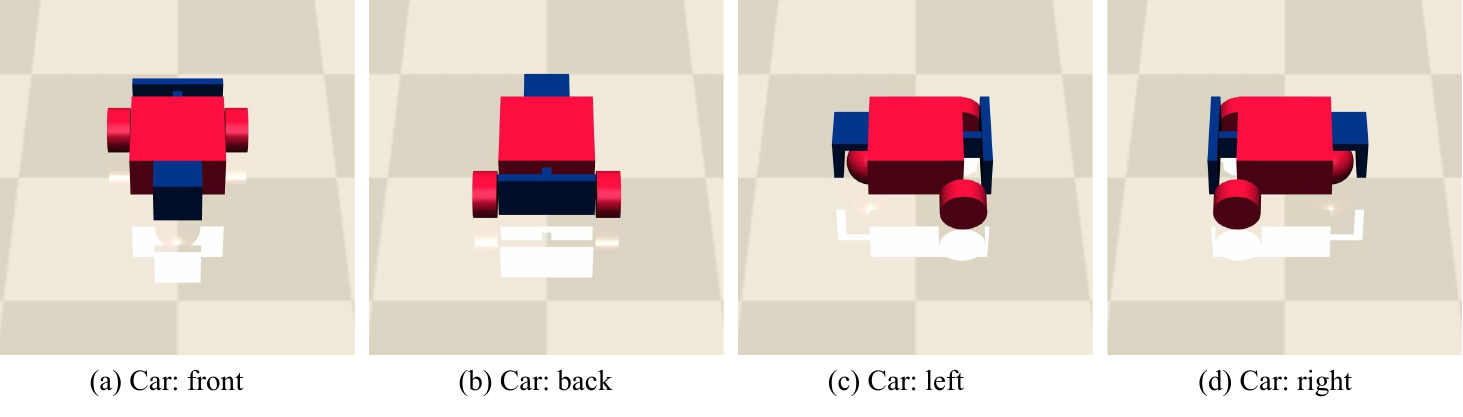}
  \caption{A different view of the robot: Car.}
  \label{pic:car}
\end{figure}

\begin{table}[H]
\caption{The overall information of Car}
\label{pic:car_overall}
\centering
\resizebox{0.5\textwidth}{!}{
\begin{tabular}{cc}
\hline
Specific Action Space      & Box(-1.0, 1.0, (2,), float64) \\ \hline
Specific Observation Space & (24, )                        \\ \hline
Observation High           & inf                           \\ \hline
Observation Low            & -inf                          \\ \hline
\end{tabular}
}
\end{table}

\begin{table}[H]
\caption{The specific action space of Car}
\label{pic:car_action}
\resizebox{\textwidth}{!}{
\begin{tabular}{ccccccc}
\hline
Num & Action                    & Control Min & Control Max & Name (in XML file) & Joint/Site & Unit      \\ \hline
0   & force to applied on left wheel  & -1& 1           & left               & hinge      & force (N) \\ \hline
1   & force to applied on right wheel & -1& 1           & right              & hinge      & force (N) \\ \hline
\end{tabular}
}
\end{table}

\begin{table}[H]
\caption{The specific observation space of Car}
\label{pic:car_obs}
\resizebox{\textwidth}{!}{
\begin{tabular}{ccccccc}
\hline
Size & Observation   & Min  & Max & Name (in XML file) & Joint/Site & Unit                                  \\ \hline
3    & \begin{tabular}[c]{@{}c@{}}Quaternions of the rear wheel which are\\ turned into 3x3 rotation matrices\end{tabular}    & -inf & inf & ballquat\_rear        & ball       & unitless                        \\ \hline
3    & Angle velocity of the rear wheel   & -inf & inf & ballangvel\_rear        & ball        & anglular velocity (rad/s)                        \\ \hline
3    & accelerometer & -inf & inf & accelerometer      & site       & acceleration (m/s\textasciicircum{}2) \\ \hline
3    & velocimeter   & -inf & inf & velocimeter        & site       & velocity (m/s)                        \\ \hline
3    & gyro          & -inf & inf & gyro               & site       & anglular velocity (rad/s)             \\ \hline
3    & magnetometer  & -inf & inf & magnetometer       & site       & magnetic flux (Wb)                    \\ \hline
\end{tabular}
}
\end{table}

\textbf{Car:} As shown in \autoref{pic:car}, the robot in question operates in three dimensions and features two independently driven parallel wheels, along with a freely rolling rear wheel. This design requires coordinated operation of the two drives for both steering and forward/backward movement. While the robot shares similarities with a basic Point robot, it possesses added complexity. The overall information of \texttt{Car}, the specific action and observation space of \texttt{Car} is shown in \autoref{pic:car_overall}, \autoref{pic:car_action}, \autoref{pic:car_obs}.

\begin{figure}[H]
  \centering
  \includegraphics[width=0.6\linewidth]{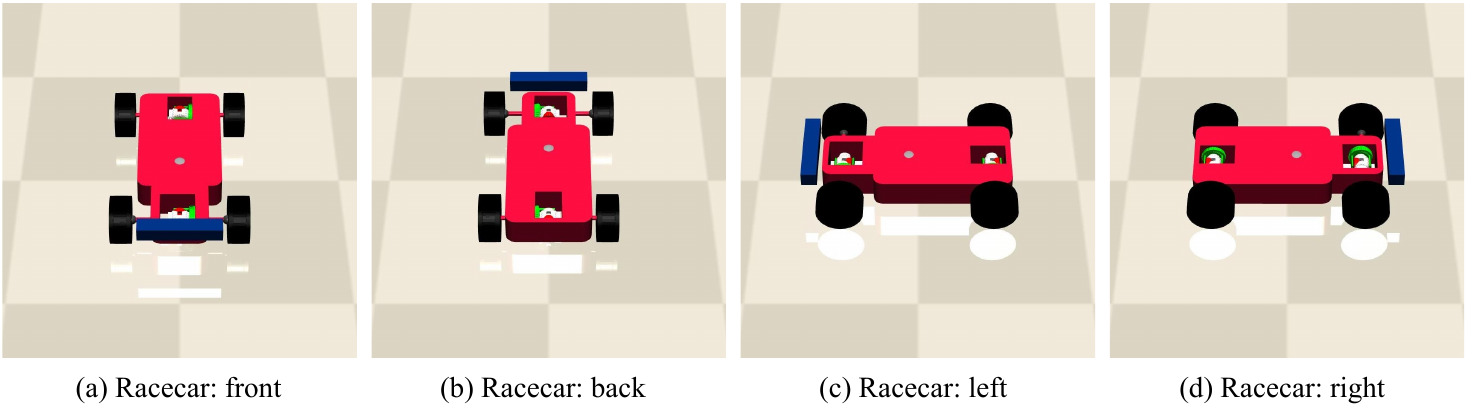}
  \caption{A different view of the robot: Racecar.}
  \label{pic:racecar}
\end{figure}

\begin{table}[H]
\caption{The overall information of Racear}
\label{pic:racecar_overall}
\centering
\resizebox{0.65\textwidth}{!}{
\begin{tabular}{cc}
\hline
Specific Action Space      & Box([-20. -0.785], [20. 0.785], (2,), float64) \\ \hline
Specific Observation Space & (12, )                        \\ \hline
Observation High           & inf                           \\ \hline
Observation Low            & -inf                          \\ \hline
\end{tabular}
}
\end{table}

\begin{table}[H]
\caption{The specific action space of Racecar}
\label{pic:racecar_action}
\resizebox{\textwidth}{!}{
\begin{tabular}{ccccccc}
\hline
Num & Action & Control Min & Control Max & Name (in XML file) & Joint/Site & Unit     \\ \hline
0   & \begin{tabular}[c]{@{}c@{}}Velocity of the\\ rear wheels.\end{tabular} & -20           & 20           & diff\_ring                  & hinge       & velocity (m/s)      \\ \hline
1   & \begin{tabular}[c]{@{}c@{}}Angle of the front \\ wheel.\end{tabular}     & -0.785& 0.785& steering\_hinge              & hinge      & angle (rad) \\ \hline
\end{tabular}
}
\end{table}

\begin{table}[H]
\caption{The specific observation space of Racecar}
\label{pic:racecar_obs}
\resizebox{\textwidth}{!}{
\begin{tabular}{ccccccc}
\hline
Size & Observation   & Min  & Max & Name (in XML file) & Joint/Site & Unit                                  \\ \hline
3    & accelerometer & -inf & inf & accelerometer      & site       & acceleration (m/s\textasciicircum{}2) \\ \hline
3    & velocimeter   & -inf & inf & velocimeter        & site       & velocity (m/s)                        \\ \hline
3    & gyro          & -inf & inf & gyro               & site       & anglular velocity (rad/s)             \\ \hline
3    & magnetometer  & -inf & inf & magnetometer       & site       & magnetic flux (Wb)                    \\ \hline
\end{tabular}
}
\end{table}

\textbf{Racecar.} As shown in \autoref{pic:racecar},  the robot is closer to realistic car dynamics, moving in three dimensions, it has one velocity servo and one position servo, one to adjusts the rear wheel speed to the target speed and the other to adjust the front wheel steering angle to the target angle. Racecar references the widely known MIT Racecar project’s dynamics model. For it to accomplish the specified goal, it must coordinate the relationship between the steering angle of the tires and the speed, just like a human driving a car. The overall information of \texttt{Racecar}, the specific action and observation space of \texttt{Racecar} is shown in \autoref{pic:racecar_overall}, \autoref{pic:racecar_action}, \autoref{pic:racecar_obs}.

\begin{table}[H]
\caption{The overall information of Ant}
\label{pic:ant_overall}
\centering
\resizebox{0.5\textwidth}{!}{
\begin{tabular}{cc}
\hline
Specific Action Space      & Box(-1.0, 1.0, (8,), float64) \\ \hline
Specific Observation Space & (40, )                        \\ \hline
Observation High           & inf                           \\ \hline
Observation Low            & -inf                          \\ \hline
\end{tabular}
}
\end{table}

\begin{table}[ht]
\caption{The specific action space of Ant}
\label{pic:ant_action}
\resizebox{\textwidth}{!}{
\begin{tabular}{ccccccc}
\hline
Num & Action & Control Min & Control Max & Name (in XML file) & Joint/Site & Unit     \\ \hline
0 & \begin{tabular}[c]{@{}c@{}}torque applied on the\\ rotor between the torso \\ and front left hip\end{tabular} & -1 & 1 & hip\_1 (front\_left\_leg) & hinge & torque (N m) \\ \hline
1 & \begin{tabular}[c]{@{}c@{}}torque applied on the \\ rotor between the front \\ left two links\end{tabular} & -1 & 1 & angle\_1 (front\_left\_leg) & hinge & torque (N m) \\ \hline
2 & \begin{tabular}[c]{@{}c@{}}torque applied on the \\ rotor between the torso \\ and front right hip\end{tabular} & -1 & 1 & hip\_2 (front\_right\_leg) & hinge & torque (N m) \\ \hline
3 & \begin{tabular}[c]{@{}c@{}}torque applied on the \\ rotor between the front \\ right two links\end{tabular} & -1 & 1 & angle\_2 (front\_right\_leg) & hinge & torque (N m) \\ \hline
4 & \begin{tabular}[c]{@{}c@{}}torque applied on the \\ rotor between the torso\\ and back left hip\end{tabular} & -1 & 1 & hip\_3 (back\_leg) & hinge & torque (N m) \\ \hline
5 & \begin{tabular}[c]{@{}c@{}}torque applied on the \\ rotor between the back \\ left two links\end{tabular} & -1 & 1 & angle\_3 (back\_leg) & hinge & torque (N m) \\ \hline
6 & \begin{tabular}[c]{@{}c@{}}torque applied on the \\ rotor between the torso \\ and back right hip\end{tabular} & -1 & 1 & hip\_4 (right\_back\_leg) & hinge & torque (N m) \\ \hline
7 & \begin{tabular}[c]{@{}c@{}}torque applied on the \\ rotor between the back \\ right two links\end{tabular} & -1 & 1 & angle\_4 (right\_back\_leg) & hinge & torque (N m) \\ \hline
\end{tabular}
}
\end{table}

\begin{table}[ht]
\caption{The specific observation space of Ant}
\label{pic:ant_obs}
\resizebox{1.0\textwidth}{!}{
\begin{tabular}{ccccccc}
\hline
Size & Observation   & Min  & Max & Name (in XML file) & Joint/Site & Unit                                  \\ \hline
3 & accelerometer & -inf & inf & accelerometer & site & acceleration (m/s\textasciicircum{}2) \\ \hline
3 & velocimeter & -inf & inf & velocimeter & site & velocity (m/s) \\ \hline
3 & gyro & -inf & inf & gyro & site & anglular velocity (rad/s) \\ \hline
3 & magnetometer & -inf & inf & magnetometer & site & magnetic flux (Wb) \\ \hline
1 & \begin{tabular}[c]{@{}c@{}}angular velocity of angle \\ between torso and front left link\end{tabular} & -inf& inf& hip\_1 (front\_left\_leg) & hinge & angle (rad) \\ \hline
1 & \begin{tabular}[c]{@{}c@{}}angular velocity of the angle \\ between front left links\end{tabular} & -inf& inf& ankle\_1 (front\_left\_leg) & hinge & angle (rad) \\ \hline
1 & \begin{tabular}[c]{@{}c@{}}angular velocity of angle \\ between torso and front right link\end{tabular} & -inf & inf & hip\_2 (front\_right\_leg) & hinge & angle (rad) \\ \hline
1 & \begin{tabular}[c]{@{}c@{}}angular velocity of the angle \\ between front right links\end{tabular} & -inf & inf & ankle\_2 (front\_right\_leg) & hinge & angle (rad) \\ \hline
1 & \begin{tabular}[c]{@{}c@{}}angular velocity of angle \\ between torso and back left link\end{tabular} & -inf & inf & hip\_3 (back\_leg) & hinge & angle (rad) \\ \hline
1 & \begin{tabular}[c]{@{}c@{}}angular velocity of the angle \\ between back left links\end{tabular} & -inf & inf & ankle\_3 (back\_leg) & hinge & angle (rad) \\ \hline
1 & \begin{tabular}[c]{@{}c@{}}angular velocity of angle \\ between torso and back right link\end{tabular} & -inf & inf & hip\_4 (right\_back\_leg) & hinge & angle (rad) \\ \hline
1 & \begin{tabular}[c]{@{}c@{}}angular velocity of the angle \\ between back right links\end{tabular} & -inf & inf & ankle\_4 (right\_back\_leg) & hinge & angle (rad) \\ \hline
1 & \begin{tabular}[c]{@{}c@{}}z-coordinate of the torso \\ (centre).\end{tabular} & -inf & inf & torso & site & position (m) \\ \hline
3 & \begin{tabular}[c]{@{}c@{}}xyz-coordinate angular \\ velocity of the tors.\end{tabular} & -inf & inf & torso & site & angular velocity (rad/s) \\ \hline
2 & \begin{tabular}[c]{@{}c@{}}sin() and cos() of angle \\ between torso and first link on front left\end{tabular} & -inf & inf & hip\_1 (front\_left\_leg) & hinge & unitless \\ \hline
2 & \begin{tabular}[c]{@{}c@{}}sin() and cos() of angle \\ between torso and first link on front left\end{tabular} & -inf & inf & ankle\_1 (front\_left\_leg) & hinge & unitless \\ \hline
2 & \begin{tabular}[c]{@{}c@{}}sin() and cos() of angle \\ between torso and first link on front left\end{tabular} & -inf & inf & hip\_2 (front\_right\_leg) & hinge & unitless \\ \hline
2 & \begin{tabular}[c]{@{}c@{}}sin() and cos() of angle \\ between torso and first link on front left\end{tabular} & -inf & inf & ankle\_2 (front\_right\_leg) & hinge & unitless \\ \hline
2 & \begin{tabular}[c]{@{}c@{}}sin() and cos() of angle \\ between torso and first link on front left\end{tabular} & -inf & inf & hip\_3 (back\_leg) & hinge & unitless \\ \hline
2 & \begin{tabular}[c]{@{}c@{}}sin() and cos() of angle \\ between torso and first link on front left\end{tabular} & -inf & inf & ankle\_3 (back\_leg) & hinge & unitless \\ \hline
2 & \begin{tabular}[c]{@{}c@{}}sin() and cos() of angle \\ between torso and first link on front left\end{tabular} & -inf& inf & hip\_4 (right\_back\_leg) & hinge & unitless \\ \hline
2 & \begin{tabular}[c]{@{}c@{}}sin() and cos() of angle \\ between torso and first link on front left\end{tabular} & -inf& inf& ankle\_4 (right\_back\_leg) & hinge & unitless \\ \hline
\end{tabular}
}
\end{table}

\begin{figure}[H]
  \centering
  \includegraphics[width=0.6\linewidth]{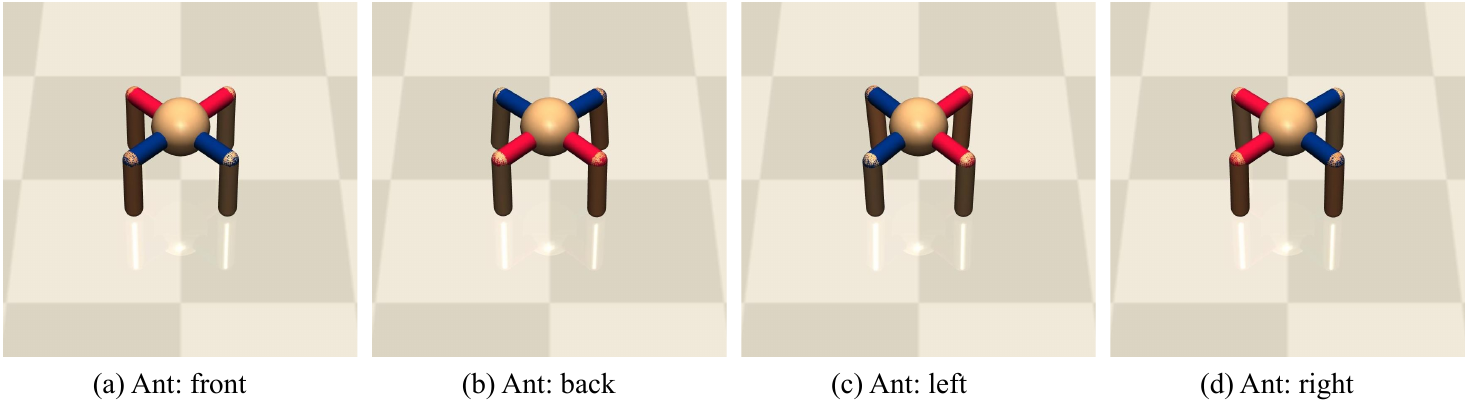}
  \caption{A different view of the robot: Ant.}
  \label{pic:ant}
\end{figure}

\textbf{Ant.} As depicted in \autoref{pic:ant}, the quadrupedal robot, inspired by the model proposed in \cite{schulman2015high}. It consists of a torso and four interconnected legs. Each leg is composed of two hinged connecting limbs, which, in turn, are connected to the torso via hinges. To achieve movement in the desired direction, coordination of the four legs is required by applying moments to the eight hinge drivers. For a comprehensive understanding of the robot, please refer to \autoref{pic:ant_overall}, \autoref{pic:ant_action}, and \autoref{pic:ant_obs}, which provide an overview of the \texttt{Ant} robot, its specific action space, and observation space, respectively.

\subsection{Multi-agents Specification}
\begin{figure}[H]
  \centering
  \includegraphics[width=\linewidth]{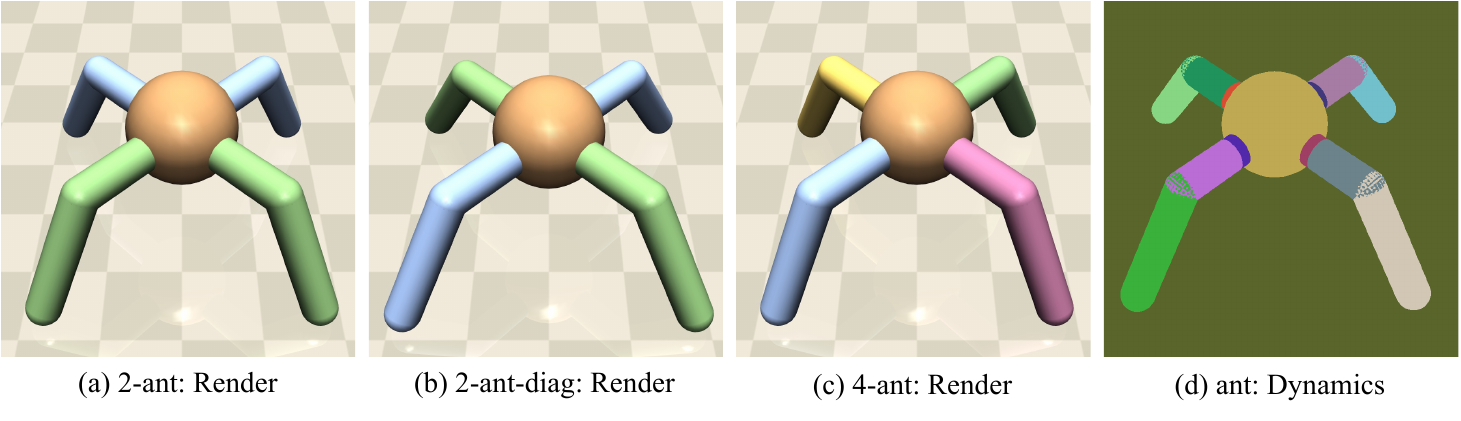}
  \caption{A different view of the MA-Ant.}
  \label{pic:ma-ant}
\end{figure}
\textbf{2-ant.} The \texttt{Ant} is partitioned into 2 parts, the front part (containing the front legs) and the back part (containing the back legs). The action space of agent-0 and agent-1 as shown in \autoref{pic:2-ant:agen-0} and \autoref{pic:2-ant:agen-1}.

\begin{table}[H]
\caption{The specific action space of 2-ant: agent-0}
\label{pic:2-ant:agen-0}
\resizebox{\textwidth}{!}{
\begin{tabular}{ccccccc}
\hline
Num & Action                                                                                                          & Control Min & Control Max & Name (in XML file)           & Joint & Unit         \\ \hline
0   & \begin{tabular}[c]{@{}c@{}}Torque applied on the rotor\\  between the torso and front\\  left hip\end{tabular}  & -1          & 1           & hip\_1 (front\_left\_leg)    & hinge & torque (N m) \\ \hline
1   & \begin{tabular}[c]{@{}c@{}}Torque applied on the rotor \\ between the front left \\ two links\end{tabular}      & -1          & 1           & angle\_1 (front\_left\_leg)  & hinge & torque (N m) \\ \hline
2   & \begin{tabular}[c]{@{}c@{}}Torque applied on the rotor \\ between the torso and front \\ right hip\end{tabular} & -1          & 1           & hip\_2 (front\_right\_leg)   & hinge & torque (N m) \\ \hline
3   & \begin{tabular}[c]{@{}c@{}}Torque applied on the rotor \\ between the front right \\ two links\end{tabular}     & -1          & 1           & angle\_2 (front\_right\_leg) & hinge & torque (N m) \\ \hline
\end{tabular}
}
\end{table}

\begin{table}[htbp]
\caption{The specific action space of 2-ant: agent-1}
\label{pic:2-ant:agen-1}
\resizebox{\textwidth}{!}{
\begin{tabular}{ccccccc}
\hline
Num & Action                                                                                                          & Control Min & Control Max & Name (in XML file)           & Joint & Unit         \\ \hline
0   & \begin{tabular}[c]{@{}c@{}}Torque applied on the rotor\\  between the torso and front\\  left hip\end{tabular}  & -1          & 1           & hip\_1 (front\_left\_leg)    & hinge & torque (N m) \\ \hline
1   & \begin{tabular}[c]{@{}c@{}}Torque applied on the rotor \\ between the front left \\ two links\end{tabular}      & -1          & 1           & angle\_1 (front\_left\_leg)  & hinge & torque (N m) \\ \hline
2   & \begin{tabular}[c]{@{}c@{}}Torque applied on the rotor \\ between the torso and front \\ right hip\end{tabular} & -1          & 1           & hip\_2 (front\_right\_leg)   & hinge & torque (N m) \\ \hline
3   & \begin{tabular}[c]{@{}c@{}}Torque applied on the rotor \\ between the front right \\ two links\end{tabular}     & -1          & 1           & angle\_2 (front\_right\_leg) & hinge & torque (N m) \\ \hline
\end{tabular}
}
\end{table}

\textbf{2-ant-diag.} The \texttt{Ant} is partitioned into 2 parts, split diagonally, the front part (containing the front legs) and the back part (containing the back legs). The action space of agent-0 and agent-1 as shown in \autoref{pic:2-ant-diag:agen-0} and \autoref{pic:2-ant-diag:agen-1}.

\begin{table}[H]
\caption{The specific action space of 2-ant-diag: agent-0}
\label{pic:2-ant-diag:agen-0}
\resizebox{\textwidth}{!}{
\begin{tabular}{ccccccc}
\hline
Num & Action                                                                                                      & Control Min & Control Max & Name (in XML file)          & Joint & Unit         \\ \hline
0   & \begin{tabular}[c]{@{}c@{}}Torque applied on the rotor \\ between the torso and front left hip\end{tabular} & -1          & 1           & hip\_1 (front\_left\_leg)   & hinge & torque (N m) \\ \hline
1   & \begin{tabular}[c]{@{}c@{}}Torque applied on the rotor \\ between the front left two links\end{tabular}     & -1          & 1           & angle\_1 (front\_left\_leg) & hinge & torque (N m) \\ \hline
2   & \begin{tabular}[c]{@{}c@{}}Torque applied on the rotor \\ between the torso and back right hip\end{tabular} & -1          & 1           & hip\_4 (right\_back\_leg)   & hinge & torque (N m) \\ \hline
3   & \begin{tabular}[c]{@{}c@{}}Torque applied on the rotor \\ between the back right two links\end{tabular}     & -1          & 1           & angle\_4 (right\_back\_leg) & hinge & torque (N m) \\ \hline
\end{tabular}
}
\end{table}

\begin{table}[H]
\caption{The specific action space of 4-ant: agent-1}
\label{pic:2-ant-diag:agen-1}
\resizebox{\textwidth}{!}{
\begin{tabular}{ccccccc}
\hline
Num & Action                                                                                                       & Control Min & Control Max & Name (in XML file)           & Joint & Unit         \\ \hline
0   & \begin{tabular}[c]{@{}c@{}}Torque applied on the rotor \\ between the torso and front right hip\end{tabular} & -1          & 1           & hip\_2 (front\_right\_leg)   & hinge & torque (N m) \\ \hline
1   & \begin{tabular}[c]{@{}c@{}}Torque applied on the rotor \\ between the front right two links\end{tabular}     & -1          & 1           & angle\_2 (front\_right\_leg) & hinge & torque (N m) \\ \hline
2   & \begin{tabular}[c]{@{}c@{}}Torque applied on the rotor \\ between the torso and back left hip\end{tabular}   & -1          & 1           & hip\_3 (back\_leg)           & hinge & torque (N m) \\ \hline
3   & \begin{tabular}[c]{@{}c@{}}Torque applied on the rotor \\ between the back left two links\end{tabular}       & -1          & 1           & angle\_3 (back\_leg)         & hinge & torque (N m) \\ \hline
\end{tabular}
}
\end{table}

\textbf{4-ant.} The \texttt{Ant} is partitioned into 4 parts, with each part corresponding to a leg of the ant. The action space of agent-0, agent-1, agent-2, and agent-3 as shown in \autoref{pic:4-ant:agent-0}, \autoref{pic:4-ant:agent-1}, \autoref{pic:4-ant:agent-2} and \autoref{pic:4-ant:agent-3}.

\begin{table}[H]
\caption{The specific action space of 4-ant: agent-0}
\label{pic:4-ant:agent-0}
\resizebox{\textwidth}{!}{
\begin{tabular}{ccccccc}
\hline
Num & Action                                                                                                      & Control Min & Control Max & Name (in XML file)          & Joint & Unit         \\ \hline
0   & \begin{tabular}[c]{@{}c@{}}Torque applied on the rotor \\ between the torso and front left hip\end{tabular} & -1          & 1           & hip\_1 (front\_left\_leg)   & hinge & torque (N m) \\ \hline
1   & \begin{tabular}[c]{@{}c@{}}Torque applied on the rotor \\ between the front left two links\end{tabular}     & -1          & 1           & angle\_1 (front\_left\_leg) & hinge & torque (N m) \\ \hline
\end{tabular}
}
\end{table}

\begin{table}[H]
\caption{The specific action space of 2-ant-diag: agent-1}
\label{pic:4-ant:agent-1}
\resizebox{\textwidth}{!}{
\begin{tabular}{ccccccc}
\hline
Num & Action                                                                                                       & Control Min & Control Max & Name (in XML file)           & Joint & Unit         \\ \hline
0   & \begin{tabular}[c]{@{}c@{}}Torque applied on the rotor \\ between the torso and front right hip\end{tabular} & -1          & 1           & hip\_2 (front\_right\_leg)   & hinge & torque (N m) \\ \hline
1   & \begin{tabular}[c]{@{}c@{}}Torque applied on the rotor \\ between the front right two links\end{tabular}     & -1          & 1           & angle\_2 (front\_right\_leg) & hinge & torque (N m) \\ \hline
\end{tabular}
}
\end{table}

\begin{table}[H]
\caption{The specific action space of 4-ant: agent-2}
\label{pic:4-ant:agent-2}
\resizebox{\textwidth}{!}{
\begin{tabular}{ccccccc}
\hline
Num & Action                                                                                                     & Control Min & Control Max & Name (in XML file)   & Joint & Unit         \\ \hline
0   & \begin{tabular}[c]{@{}c@{}}Torque applied on the rotor \\ between the torso and back left hip\end{tabular} & -1          & 1           & hip\_3 (back\_leg)   & hinge & torque (N m) \\ \hline
1   & \begin{tabular}[c]{@{}c@{}}Torque applied on the rotor \\ between the back left two links\end{tabular}     & -1          & 1           & angle\_3 (back\_leg) & hinge & torque (N m) \\ \hline
\end{tabular}
}
\end{table}

\begin{table}[H]
\caption{The specific action space of 4-ant: agent-3}
\label{pic:4-ant:agent-3}
\resizebox{\textwidth}{!}{
\begin{tabular}{ccccccc}
\hline
Num & Action                                                                                                      & Control Min & Control Max & Name (in XML file)          & Joint & Unit         \\ \hline
0   & \begin{tabular}[c]{@{}c@{}}Torque applied on the rotor \\ between the torso and back right hip\end{tabular} & -1          & 1           & hip\_4 (right\_back\_leg)   & hinge & torque (N m) \\ \hline
1   & \begin{tabular}[c]{@{}c@{}}Torque applied on the rotor \\ between the back right two links\end{tabular}     & -1          & 1           & angle\_4 (right\_back\_leg) & hinge & torque (N m) \\ \hline
\end{tabular}
}
\end{table}

In addition to the robots mentioned in this paper, we also provide other multi-agent versions of robots. Due to space constraints, we did not elaborate on them extensively in the paper. However, you can refer to \url{https://www.safety-gymnasium.com/} for more detailed information.

\subsection{Task Representation}
\begin{figure}[H]
  \centering
  \includegraphics[width=\linewidth]{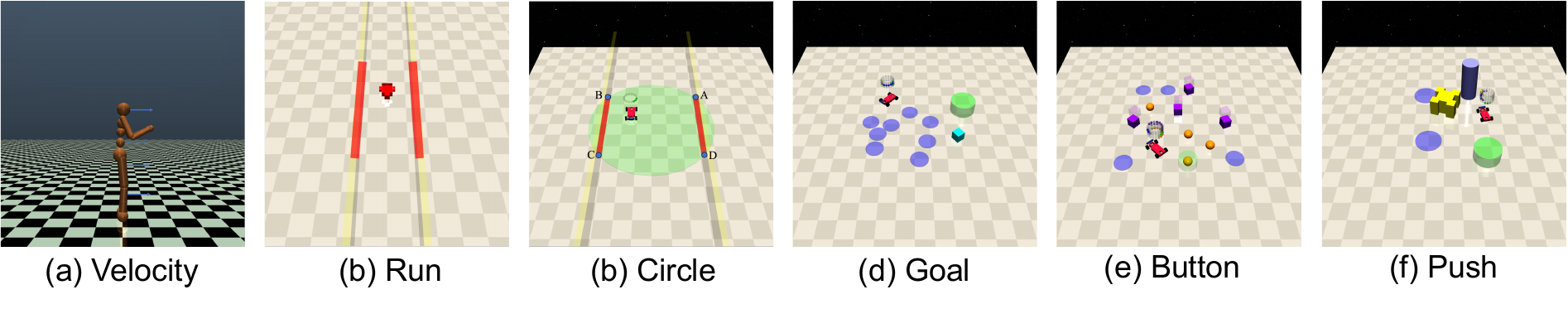}
  \caption{Tasks of Gymnasium-based Environments.}
  \label{pic: tasks}
\end{figure}

As shown in \autoref{pic: tasks}, the Gymnasium-based learning environments support the following tasks:

\textbf{Velocity:} the robot aims to facilitate coordinated leg movement of the robot in the forward (right) direction by exerting torques on the hinges.

\textbf{Run:} the robot starts with a random initial direction and a specific initial speed as it embarks on a journey to reach the opposite side of the map.

\textbf{Circle:} the reward is maximized by moving along the green circle, and the agent is not allowed to enter the outside of the red region, so its optimal constrained path follows the line segments $AD$ and $BC$. The reward function: $R(s) =\frac{v^T[-y, x]}{1+\left|\|[x, y]\|_2-d\right|}$, the cost function is $C(s) =\mathbf{1}\left[|x|>x_{\mathrm{lim}}\right]$, where $x,y$ are the coordinates in the plane, $v$ is the velocity, and $d, x_{\mathrm{lim}}$ are environmental parameters.

\textbf{Goal:} the robot navigates to multiple goal positions. After successfully reaching a goal, its location is randomly reset while maintaining the overall layout. Achieving a goal position, indicated by entering the goal circle, yields a sparse reward. Additionally, a dense reward encourages the robot's progress by rewarding proximity to the goal.

\textbf{Push:} the objective is to move a box to a series of goal positions. Like the goal task, a new random goal location is generated after each successful achievement. The sparse reward is earned when the yellow box enters the designated goal circle. The dense reward consists of two components: one for moving the agent closer to the box and another for bringing the box closer to the final goal.

\textbf{Button:} the objective is to activate a series of goal buttons distributed throughout the environment. The agent's goal is to navigate towards and make contact with the currently highlighted button, known as the goal button. Once the correct button is pressed, a new goal button is selected and highlighted while preserving the rest of the environment. The sparse reward is earned upon successfully pressing the current goal button, while the dense reward component provides a bonus for progressing toward the highlighted goal button.

\subsection{Constraint Specification}
\begin{figure}[H]
  \centering
  \includegraphics[width=\linewidth]{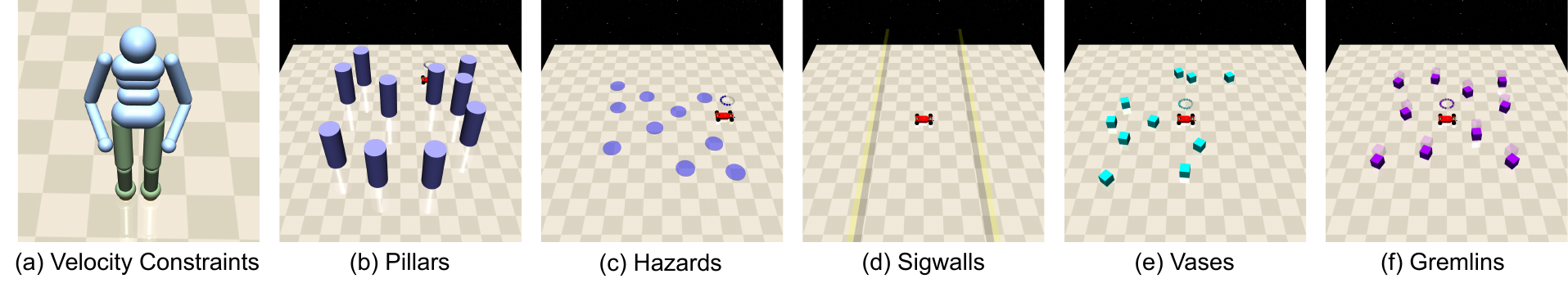}
  \caption{Constraints of Gymnasium-based Environments.}
\end{figure}

\textbf{Velocity-Constraint} consists of a series of safety tasks based on MuJoCo agents~\cite{todorov2012mujoco}. In these tasks, agents, such as \texttt{Ant}, \texttt{HalfCheetah}, and \texttt{Humanoid}, are trained to move faster for higher rewards, while also being imposed a velocity constraint for safety considerations. Formally, for an agent moving on a two-dimensional plane, the velocity is calculated as 
$
v(s,a)=\sqrt{v^2_x+v^2_y};
$
for an agent moving along a straight line, the velocity is calculated as $
v(s,a)=|v_x|,
$
where $v_x$, $v_y$ are the velocities of the agent in the $x$ and $y$ directions respectively.
Then, 
$cost(s,a) = [v(s,a) > v_{limit}],$
Here, $[P]$ denotes a notation where the value is $1$ if the proposition $P$ is true, and $0$ otherwise.

\textbf{Pillars} are employed to represent large cylindrical obstacles within the environment. In the general setting, contact with a pillar incurs costs.

\textbf{Hazards} are utilized to model areas within the environment that pose a risk, resulting in costs when an agent enters such areas.

\textbf{Sigwalls} are designed specifically for Circle tasks. They serve as visual representations of two or four solid walls, which limit the circular area to a smaller region. Crossing the wall from inside the safe area to the outside incurs costs.

\textbf{Vases} are specifically designed for Goal tasks. They represent static and fragile objects within the environment. Touching or displacing these objects incurs costs for the agent.

\textbf{Gremlins} are specifically employed in the Button tasks. They represent moving objects within the environment that can interact with the agent.

\subsection{Vision-only Tasks}
\label{app:vision-only-tasks}

In recent years, vision-only SafeRL has gained significant attention as a focal point of research, primarily due to its applicability in real-world contexts \cite{ma2022conservative, as2022constrained}. While the initial iteration of Safety Gym offered rudimentary visual input support, there is room for enhancing the realism and complexity of its environment. To effectively evaluate vision-based safe reinforcement learning algorithms, we have devised some more realistic visual tasks utilizing MuJoCo. This enhanced environment facilitates the incorporation of both RGB and RGB-d inputs. More details can be referred to our online documentation: \url{https://www.safety-gymnasium.com/en/latest/environments/safe_vision.html}.
\begin{figure}[H]
    \centering
      \begin{subfigure}{0.3\linewidth}
        \centering
        \includegraphics[width=\linewidth]{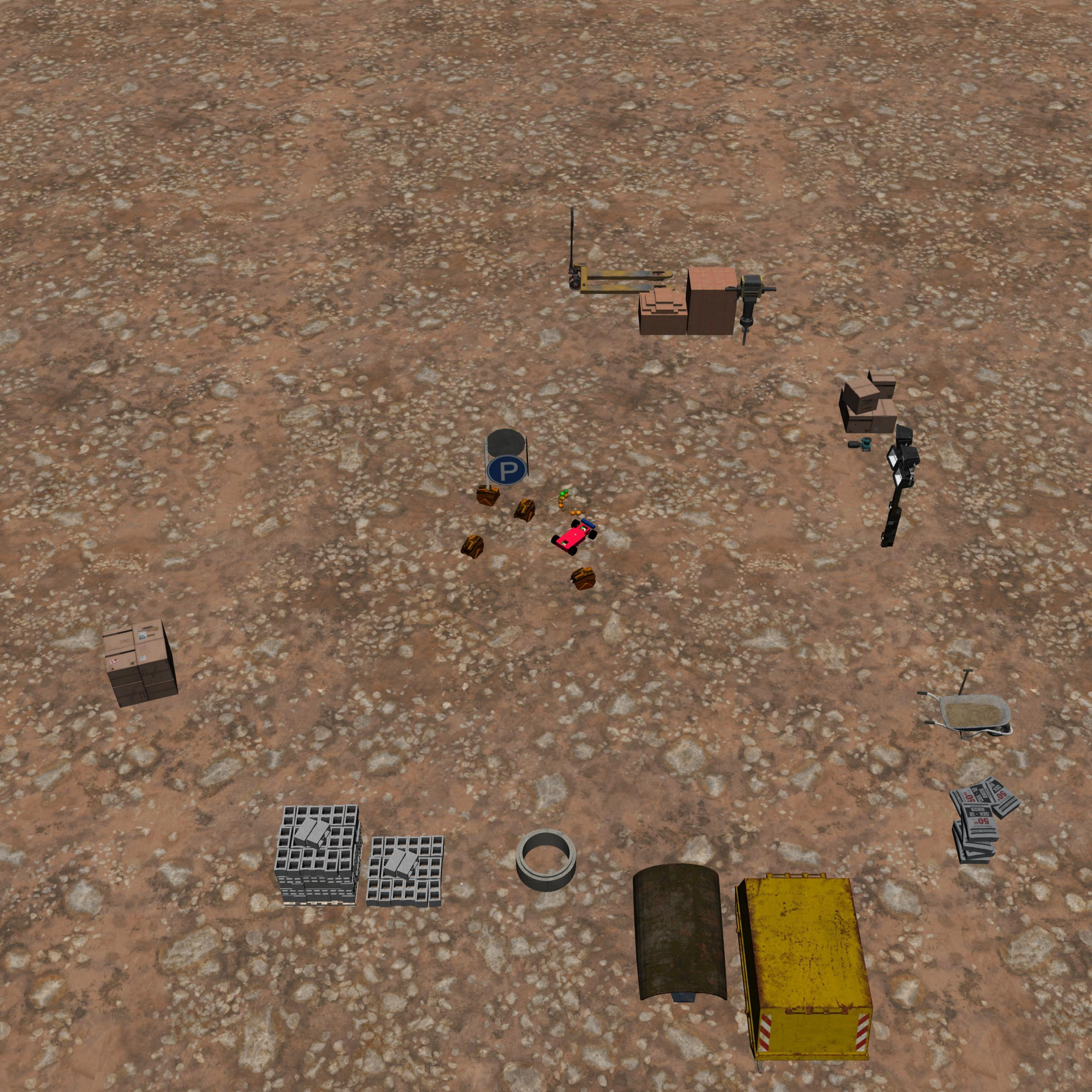}
        \caption{BuildingButton0}
      \end{subfigure}
      \begin{subfigure}{0.3\linewidth}
        \centering
        \includegraphics[width=\linewidth]{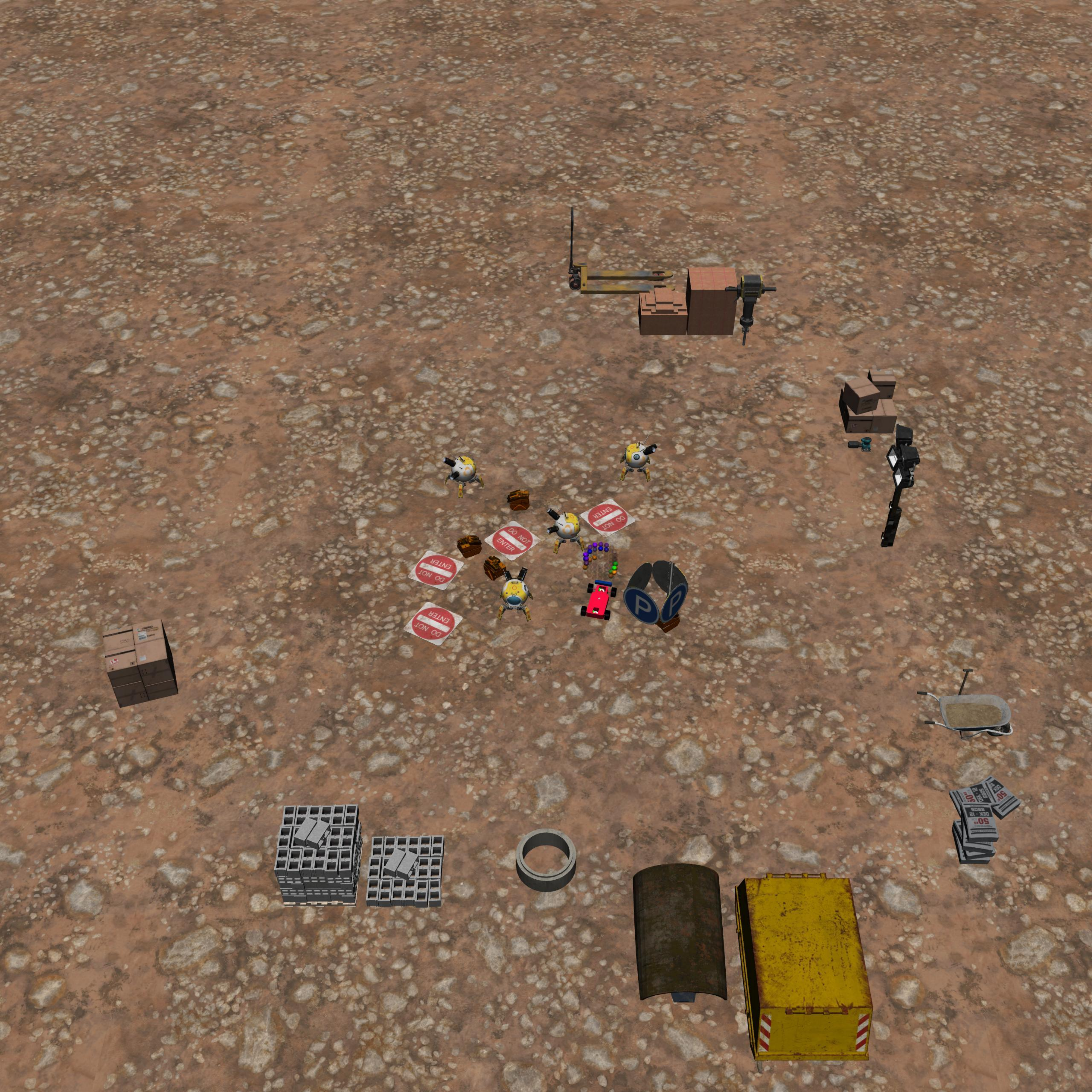}
        \caption{BuildingButton1}
      \end{subfigure}
      \begin{subfigure}{0.3\linewidth}
        \centering
        \includegraphics[width=\linewidth]{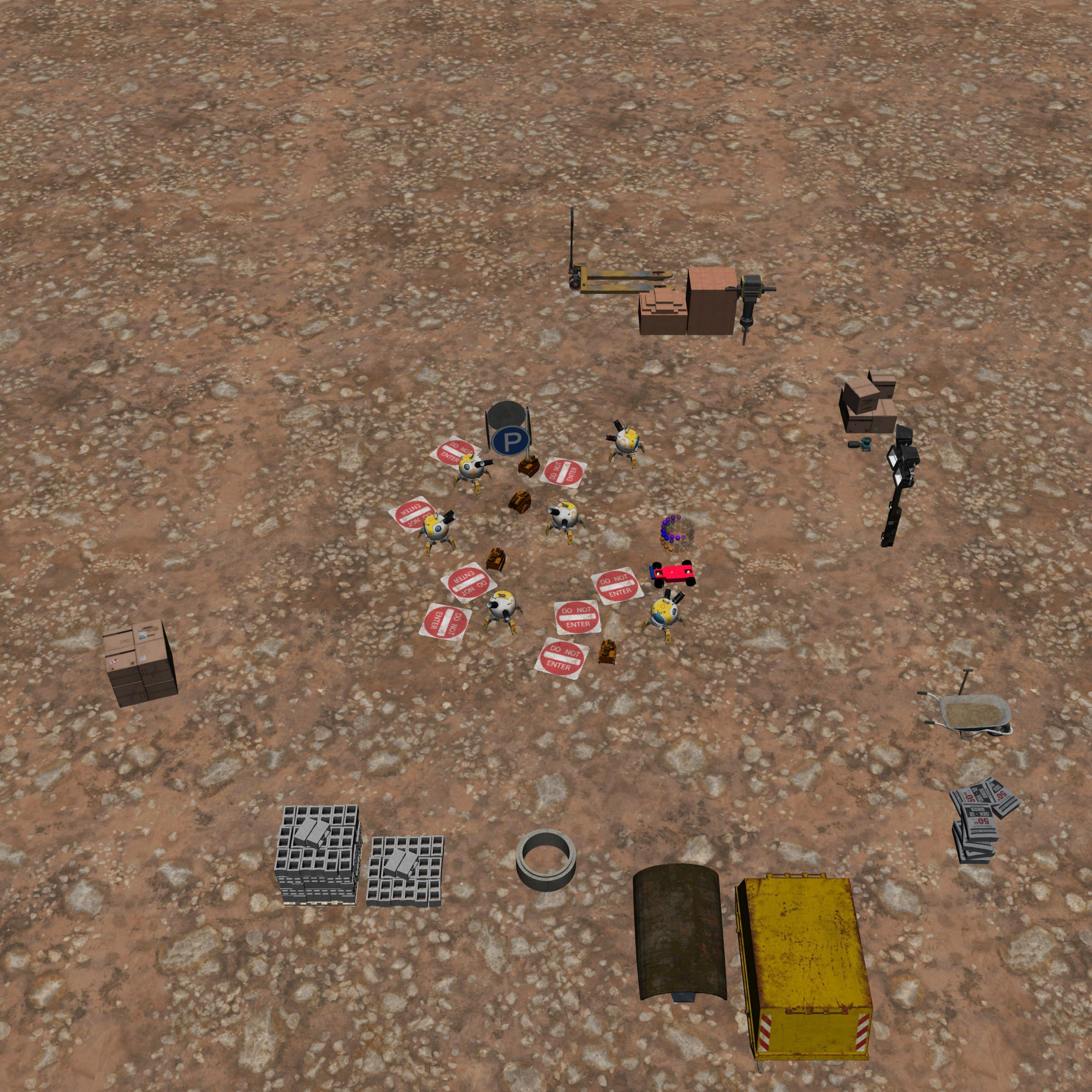}
        \caption{BuildingButton2}
      \end{subfigure}
      \caption{Overview of BuildingButton tasks.}
\end{figure}

\textbf{The Level 0 of BuildingButton} requires the agent to operate multiple machines within a construction site.

\textbf{The Level 1 of BuildingButton} requires the agent to proficiently and accurately operate multiple machines within a construction site, while concurrently evading other robots and obstacles present in the area.

\textbf{The Level 2 of BuildingButton} requires the agent to proficiently and accurately operate multiple machines within a construction site, while concurrently evading a heightened number of other robots and obstacles in the area.
\begin{figure}[H]
    \centering
      \begin{subfigure}{0.3\linewidth}
        \centering
        \includegraphics[width=\linewidth]{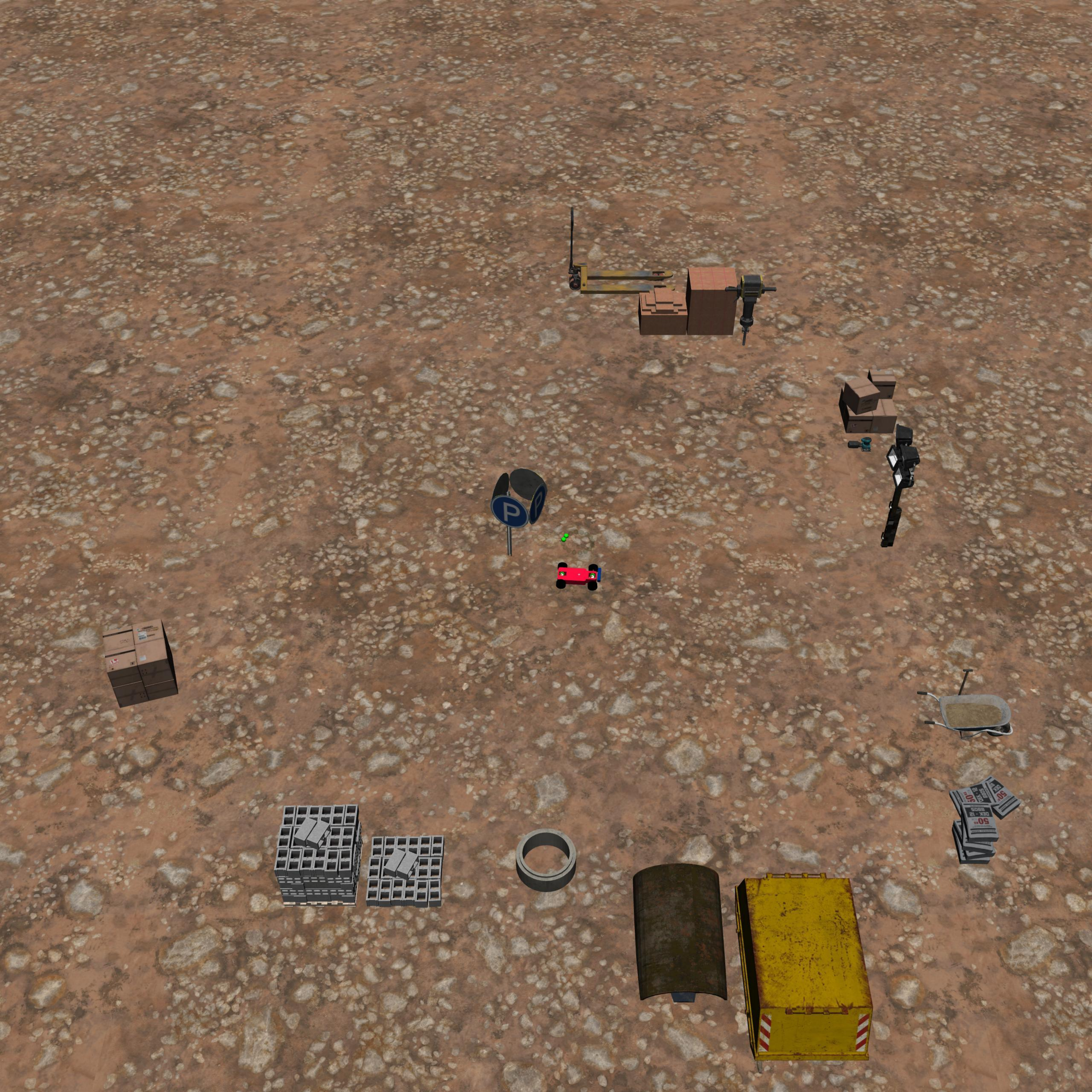}
        \caption{BuildingGoal0}
      \end{subfigure}
      \begin{subfigure}{0.3\linewidth}
        \centering
        \includegraphics[width=\linewidth]{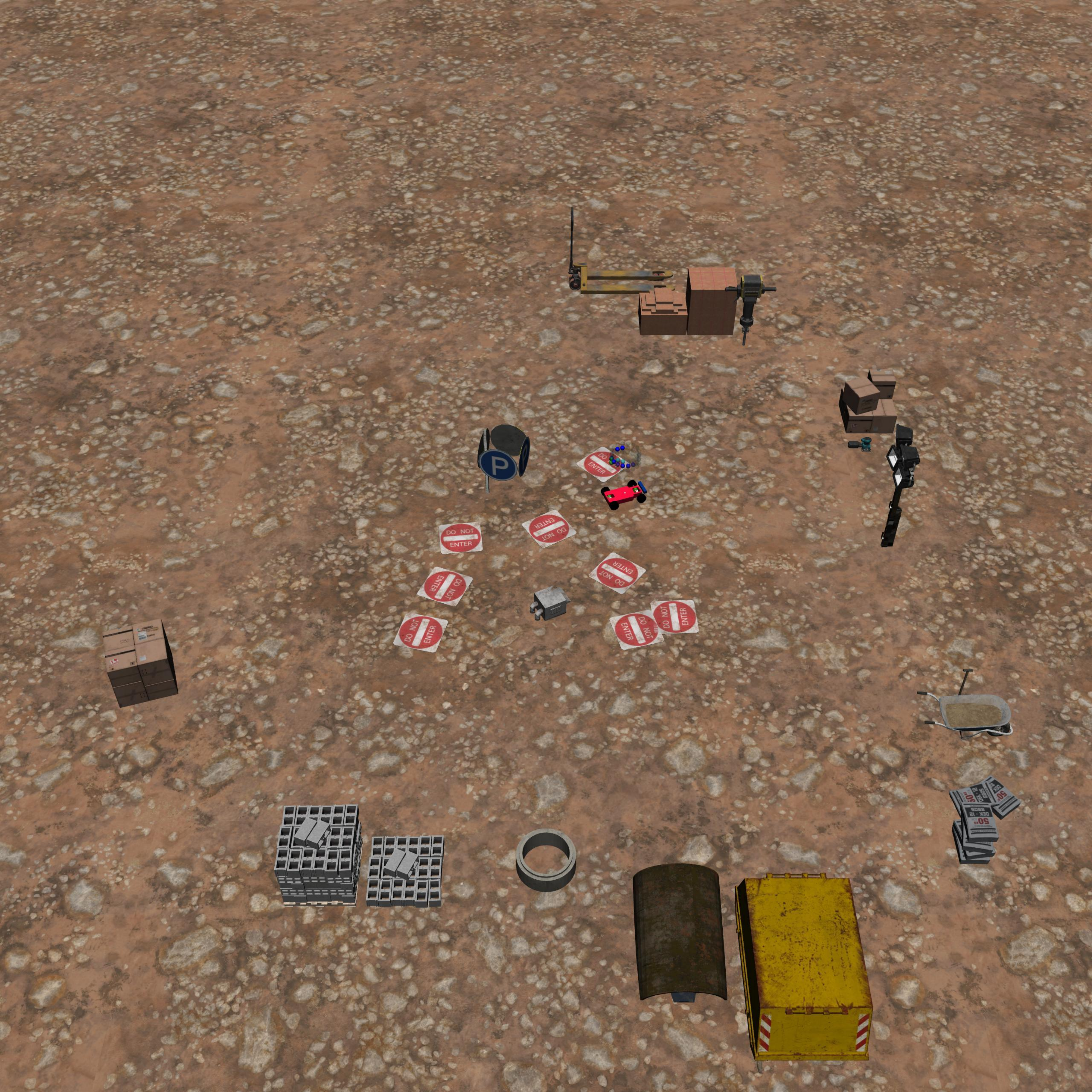}
        \caption{BuildingGoal1}
      \end{subfigure}
      \begin{subfigure}{0.3\linewidth}
        \centering
        \includegraphics[width=\linewidth]{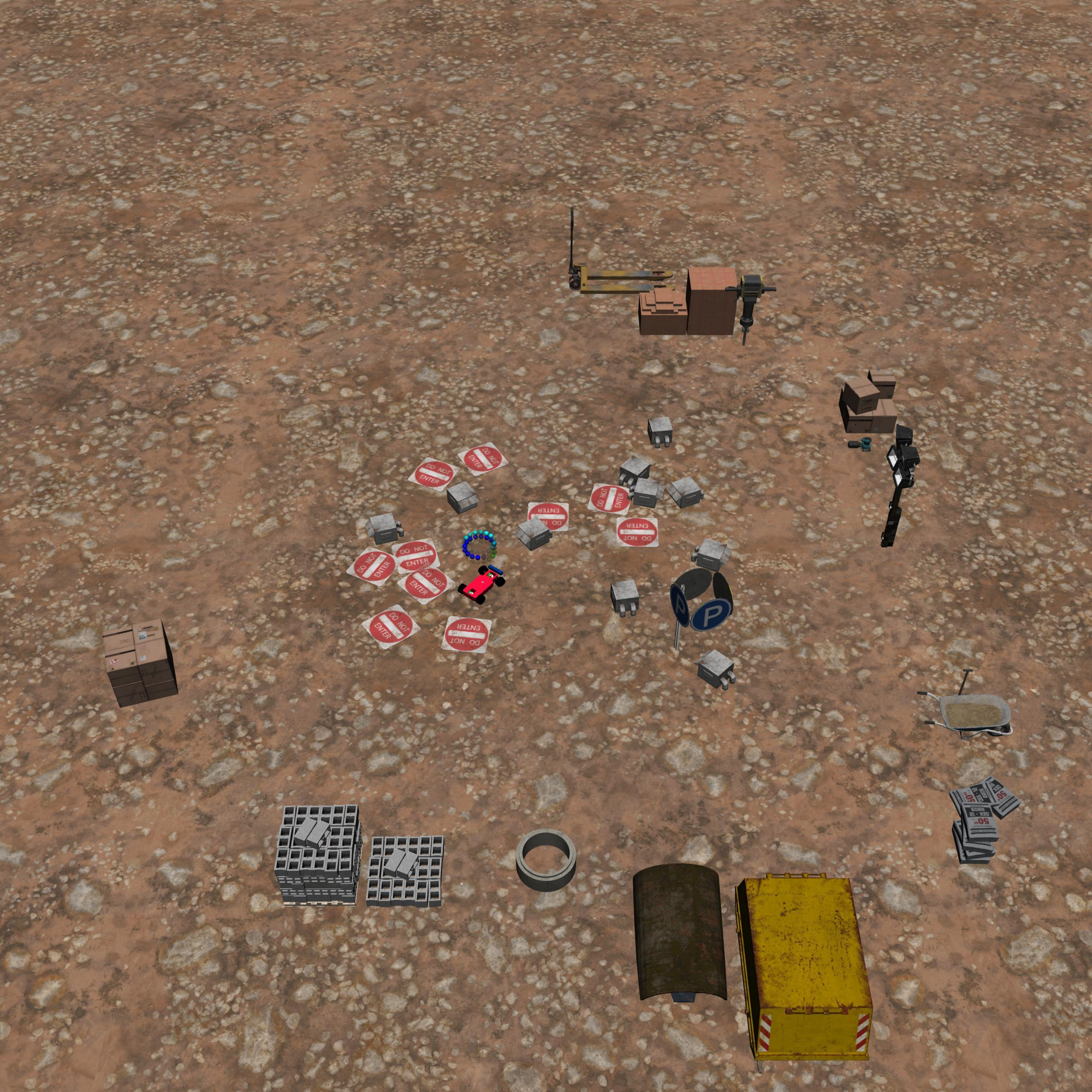}
        \caption{BuildingGoal2}
      \end{subfigure}
      \caption{Overview of BuildingGoal tasks.}
\end{figure}

\textbf{The Level 0 of BuildingGoal} requires the agent to dock at designated positions within a construction site.

\textbf{The Level 1 of BuildingGoal} requires the agent to dock at designated positions within a construction site while ensuring to avoid entry into hazardous areas.

\textbf{The Level 2 of BuildingGoal} requires the agent to dock at designated positions within a construction site, 
while ensuring to avoid entry into hazardous areas and circumventing the site’s exhaust fans.

\begin{figure}[H]
    \centering
      \begin{subfigure}{0.3\linewidth}
        \centering
        \includegraphics[width=\linewidth]{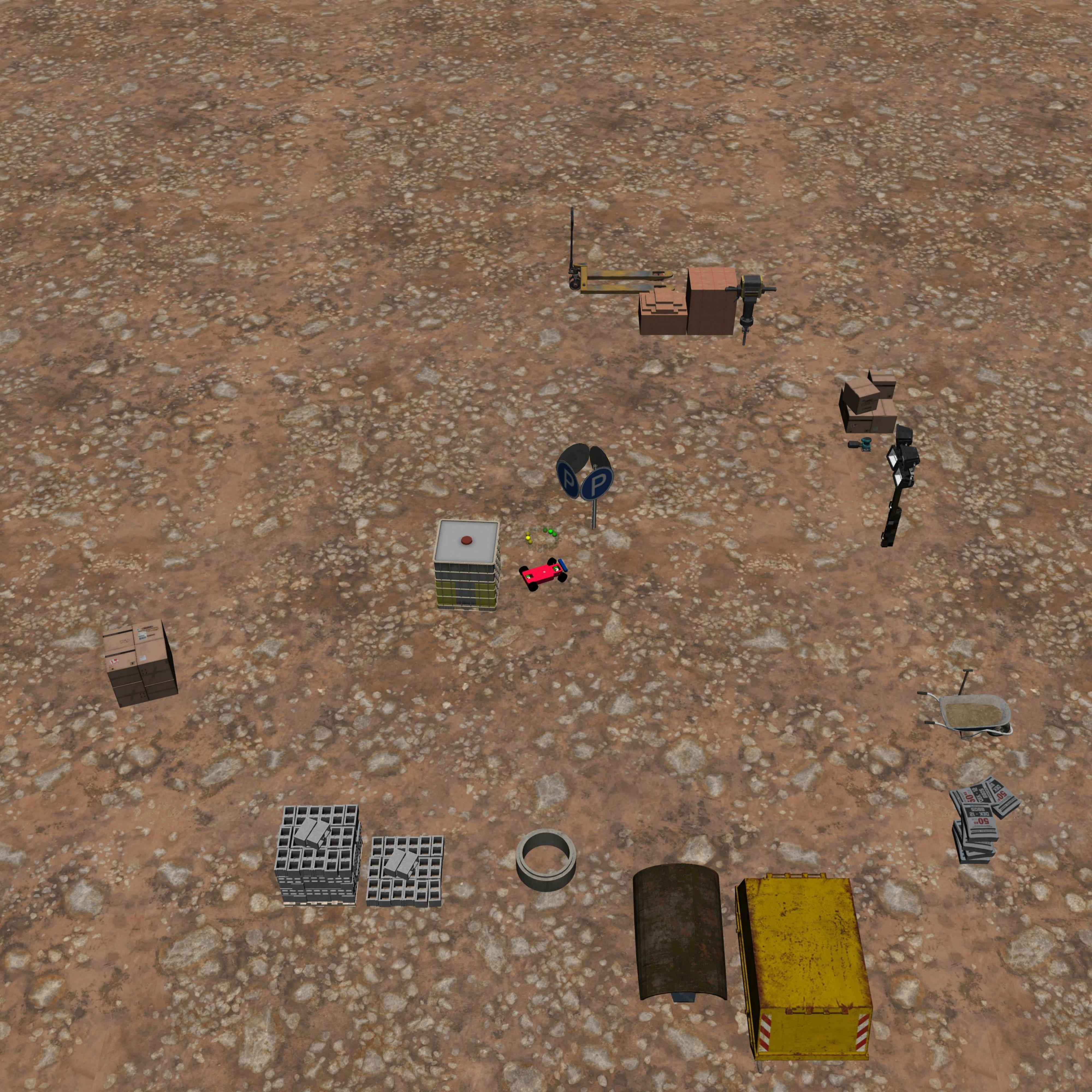}
        \caption{BuildingPush0}
      \end{subfigure}
      \begin{subfigure}{0.3\linewidth}
        \centering
        \includegraphics[width=\linewidth]{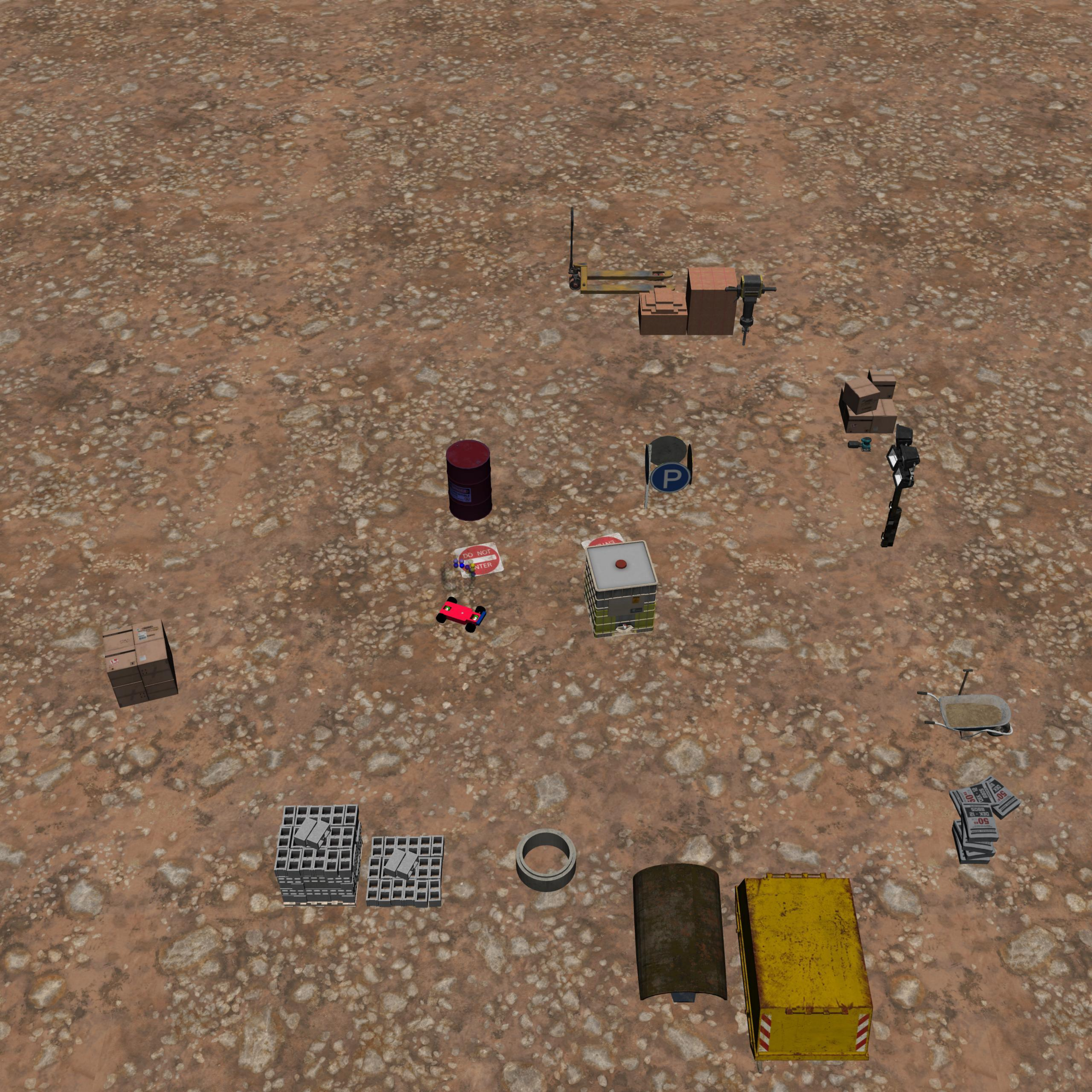}
        \caption{BuildingPush1}
      \end{subfigure}
      \begin{subfigure}{0.3\linewidth}
        \centering
        \includegraphics[width=\linewidth]{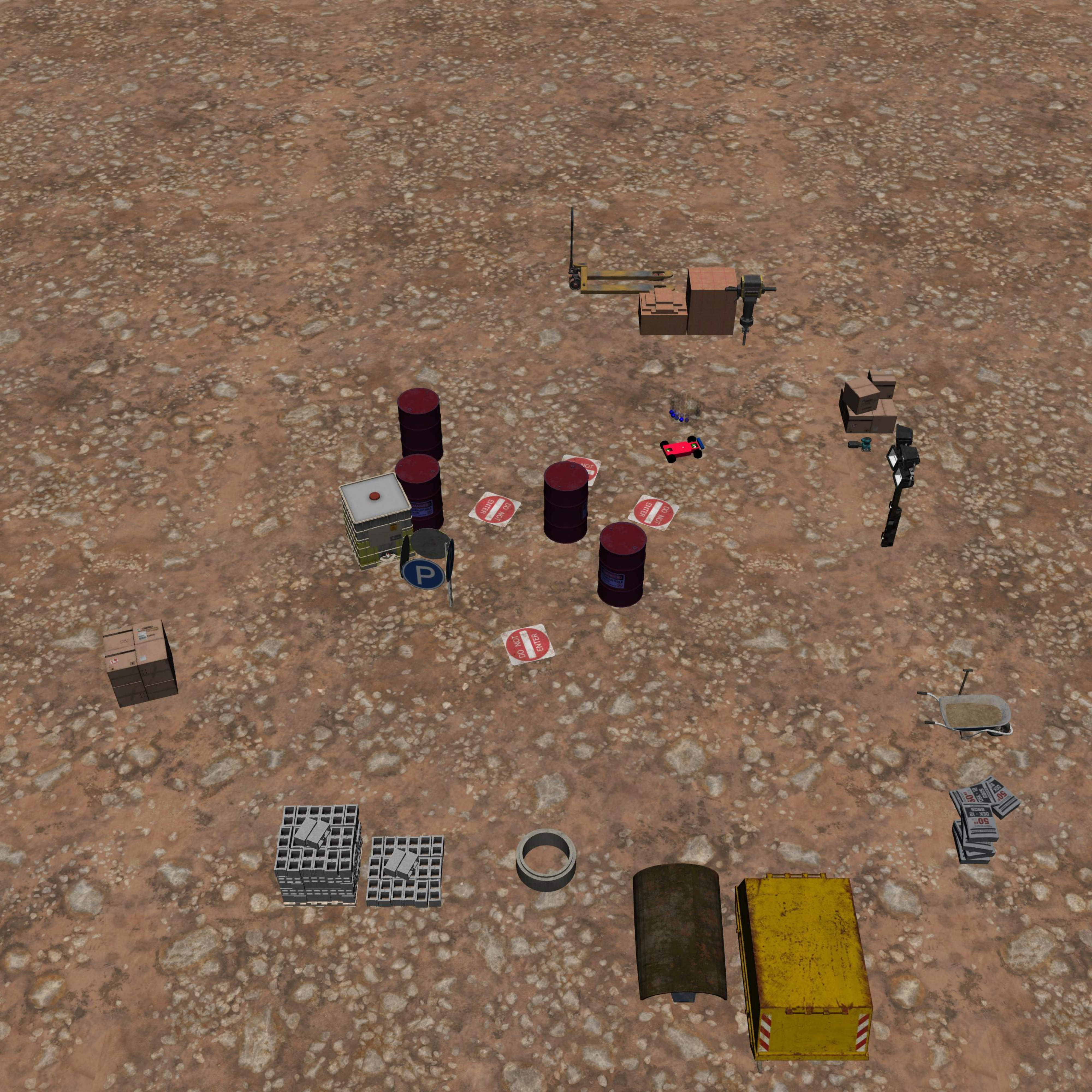}
        \caption{BuildingPush2}
      \end{subfigure}
      \caption{Overview of BuildingPush tasks.}
\end{figure}

\textbf{The Level 0 of BuildingPush} requires the agent to relocate the box to designated locations within a construction site.

\textbf{The Level 1 of BuildingPush} requires the agent to relocate the box to designated locations within a construction site while avoiding areas demarcated as restricted.

\textbf{The Level 2 of BuildingPush} requires the agent to relocate the box to designated locations within a construction while avoiding numerous hazardous fuel drums and areas demarcated as restricted.

\begin{figure}[H]
    \centering
      \begin{subfigure}{0.3\linewidth}
        \centering
        \includegraphics[width=\linewidth]{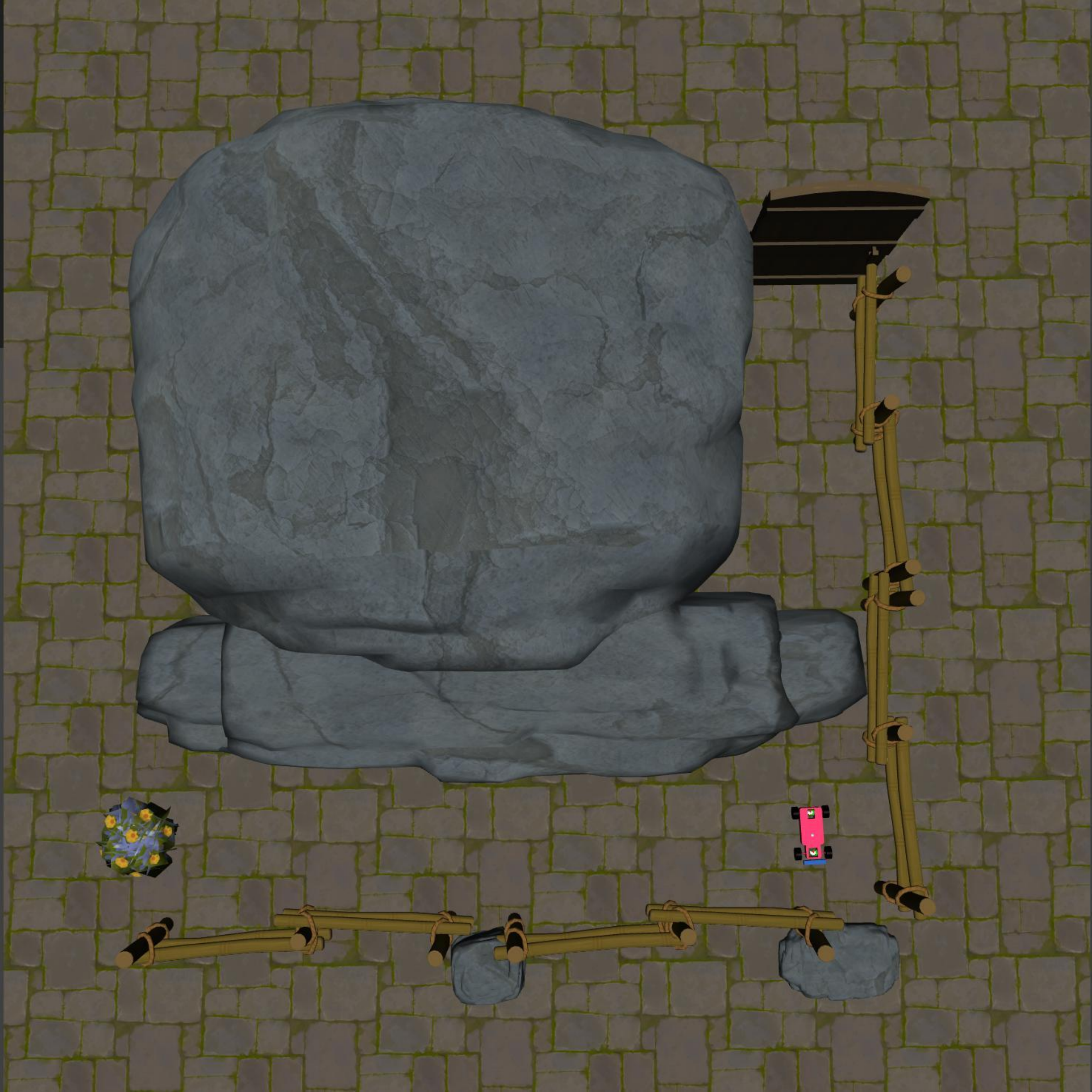}
        \caption{Race0}
      \end{subfigure}
      \begin{subfigure}{0.3\linewidth}
        \centering
        \includegraphics[width=\linewidth]{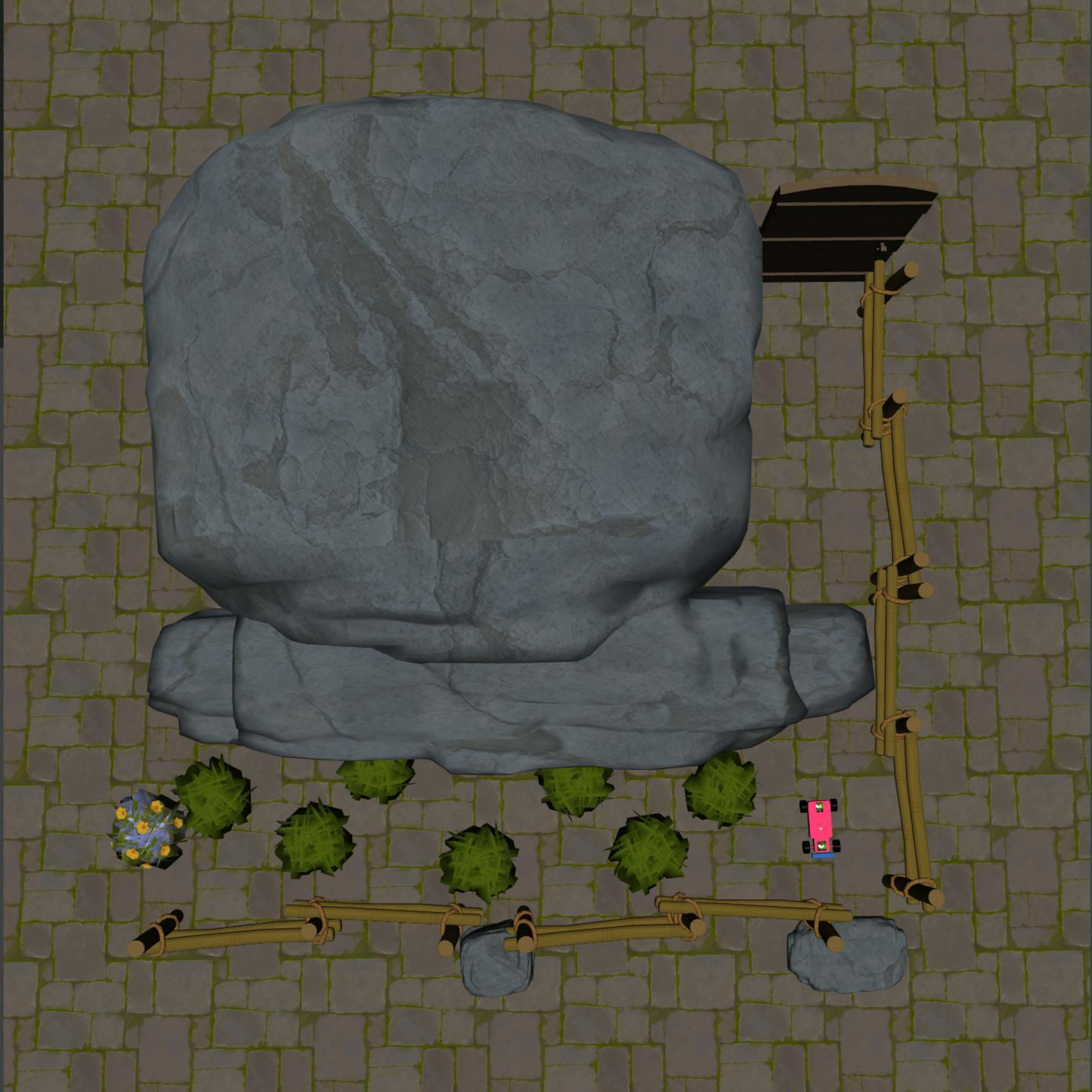}
        \caption{Race1}
      \end{subfigure}
      \begin{subfigure}{0.3\linewidth}
        \centering
        \includegraphics[width=\linewidth]{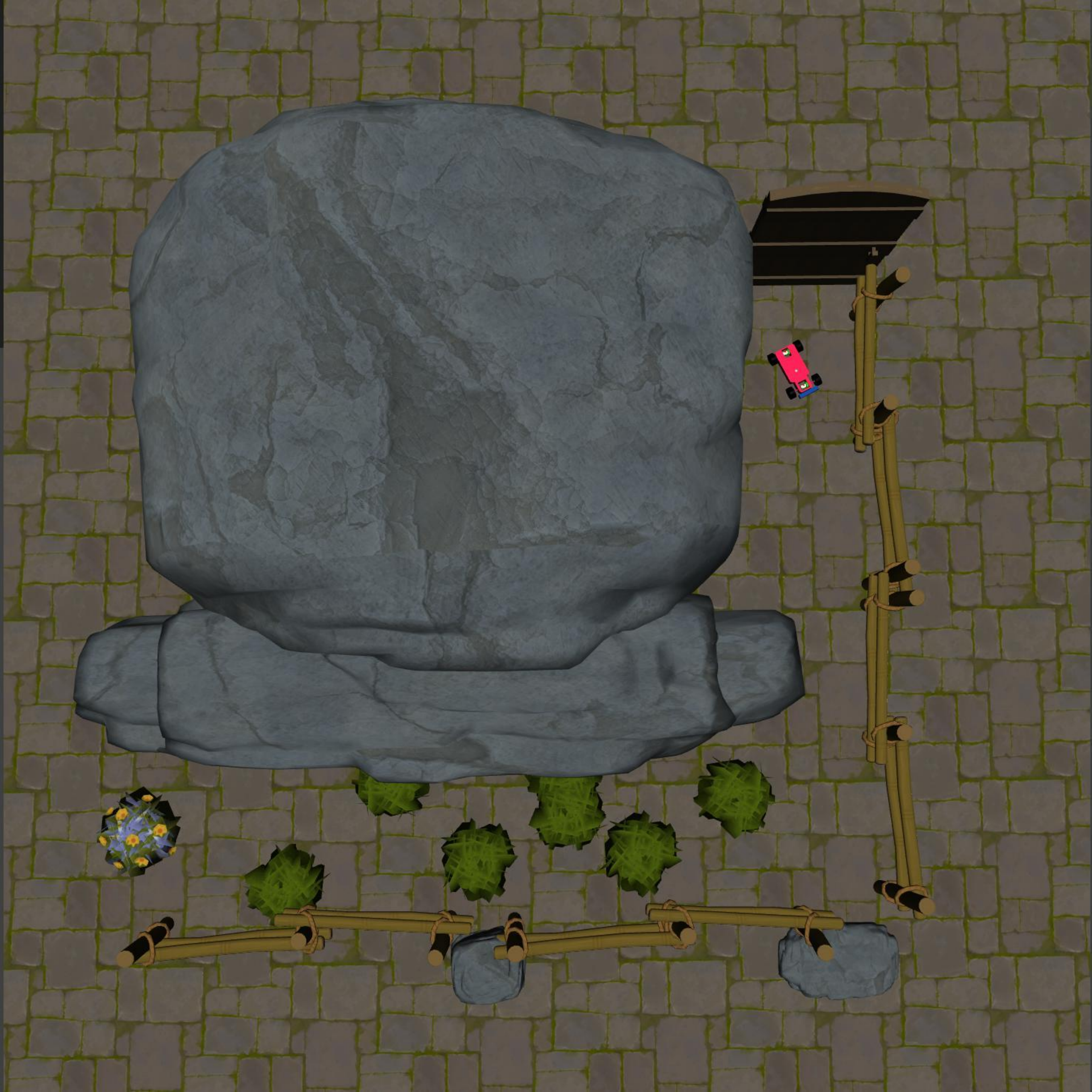}
        \caption{Race2}
      \end{subfigure}
      \caption{Overview of Race tasks.}
\end{figure}







\textbf{The Level 0 of Race} requires the agent to reach the goal position.

\textbf{The Level 1 of Race} requires the agent to reach the goal position while ensuring it avoids straying into the grass and prevents collisions with roadside objects.

\textbf{The Level 2 of Race} requires the agent to reach the goal position from a distant starting point while ensuring it avoids straying into the grass and prevents collisions with roadside objects.

\begin{figure}[H]
    \centering
      \begin{subfigure}{0.3\linewidth}
        \centering
        \includegraphics[width=\linewidth]{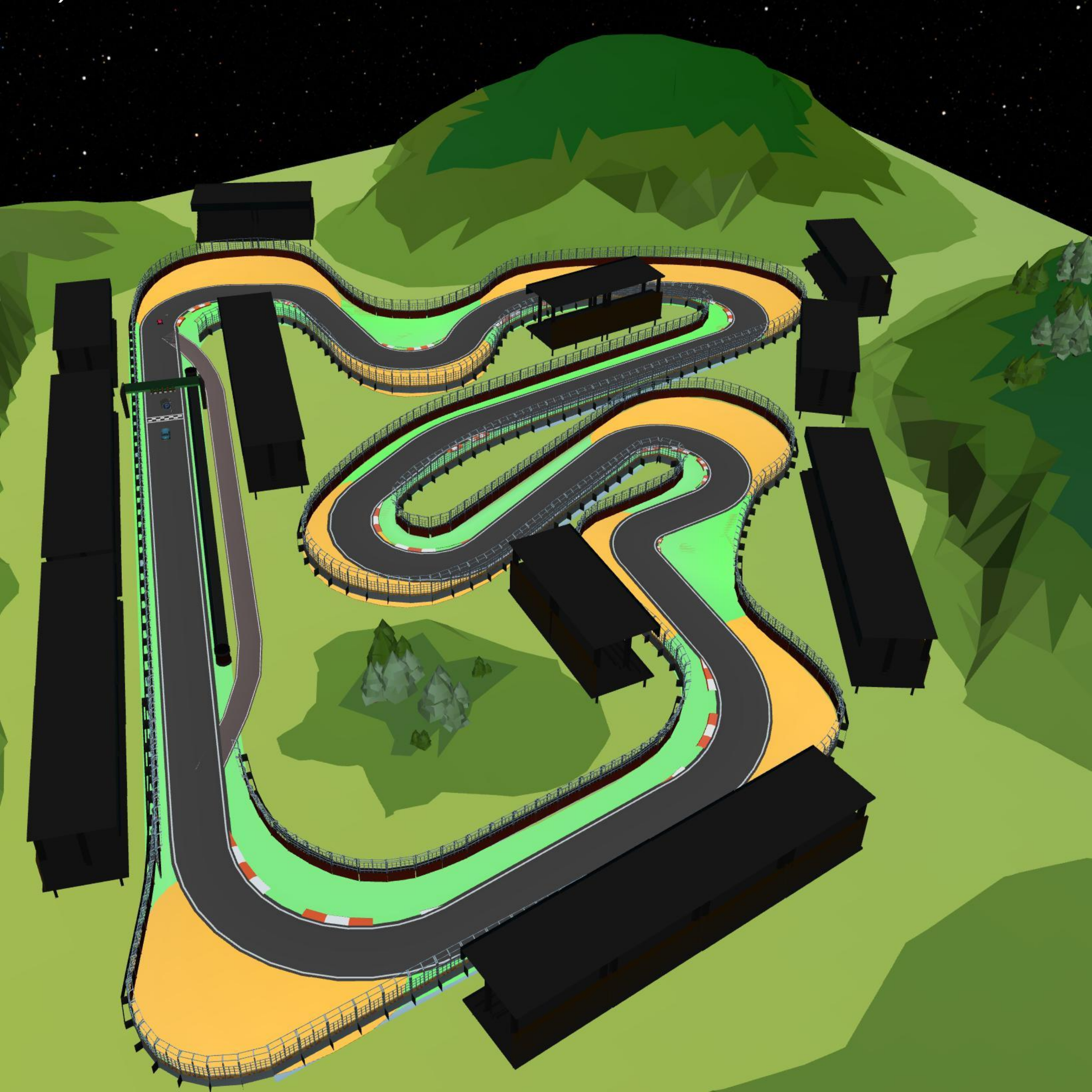}
        \caption{FormulaOne0}
      \end{subfigure}
      \begin{subfigure}{0.3\linewidth}
        \centering
        \includegraphics[width=\linewidth]{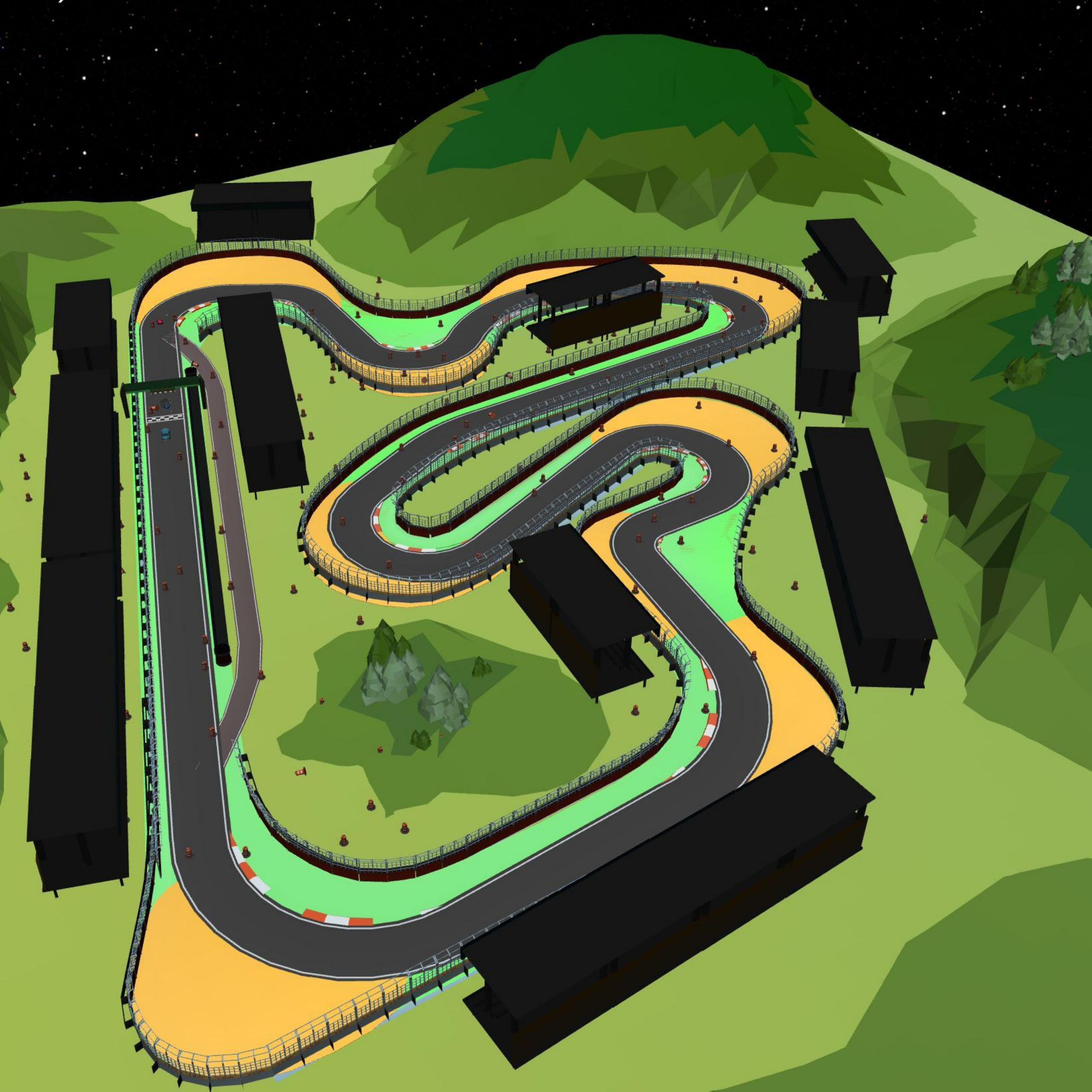}
        \caption{FormulaOne1}
      \end{subfigure}
      \begin{subfigure}{0.3\linewidth}
        \centering
        \includegraphics[width=\linewidth]{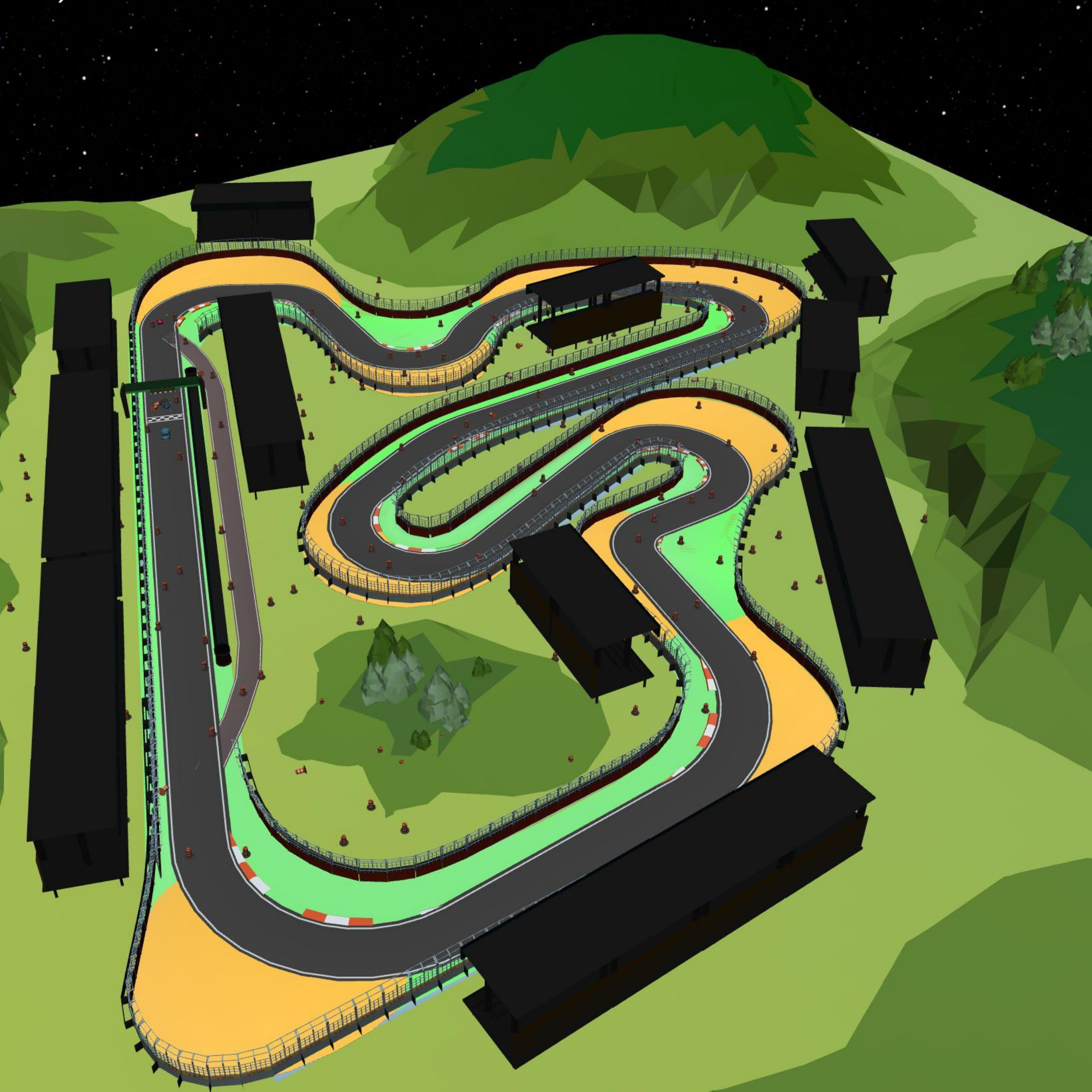}
        \caption{FormulaOne2}
      \end{subfigure}
      \caption{Overview of FormulaOne tasks.}
\end{figure}

\textbf{The Level 0 of FormulaOne} requires the agent to maximize its reach to the goal position. For each episode, the agent is randomly initialized at one of the seven checkpoints.

\textbf{The Level 1 of FormulaOne} requires the agent to maximize its reach to the goal position while circumventing barriers and racetrack fences. For each episode, the agent is randomly initialized at one of the seven checkpoints.

\textbf{The Level 2 of FormulaOne} requires the agent to maximize its reach to the goal position while circumventing barriers and racetrack fences. For each episode, the agent is randomly initialized at one of the seven checkpoints. Notably, the barriers surrounding the checkpoints are denser.

\subsection{Some Issues about Safety Gym}
\begin{figure}[H]
  \centering
  \includegraphics[width=\linewidth]{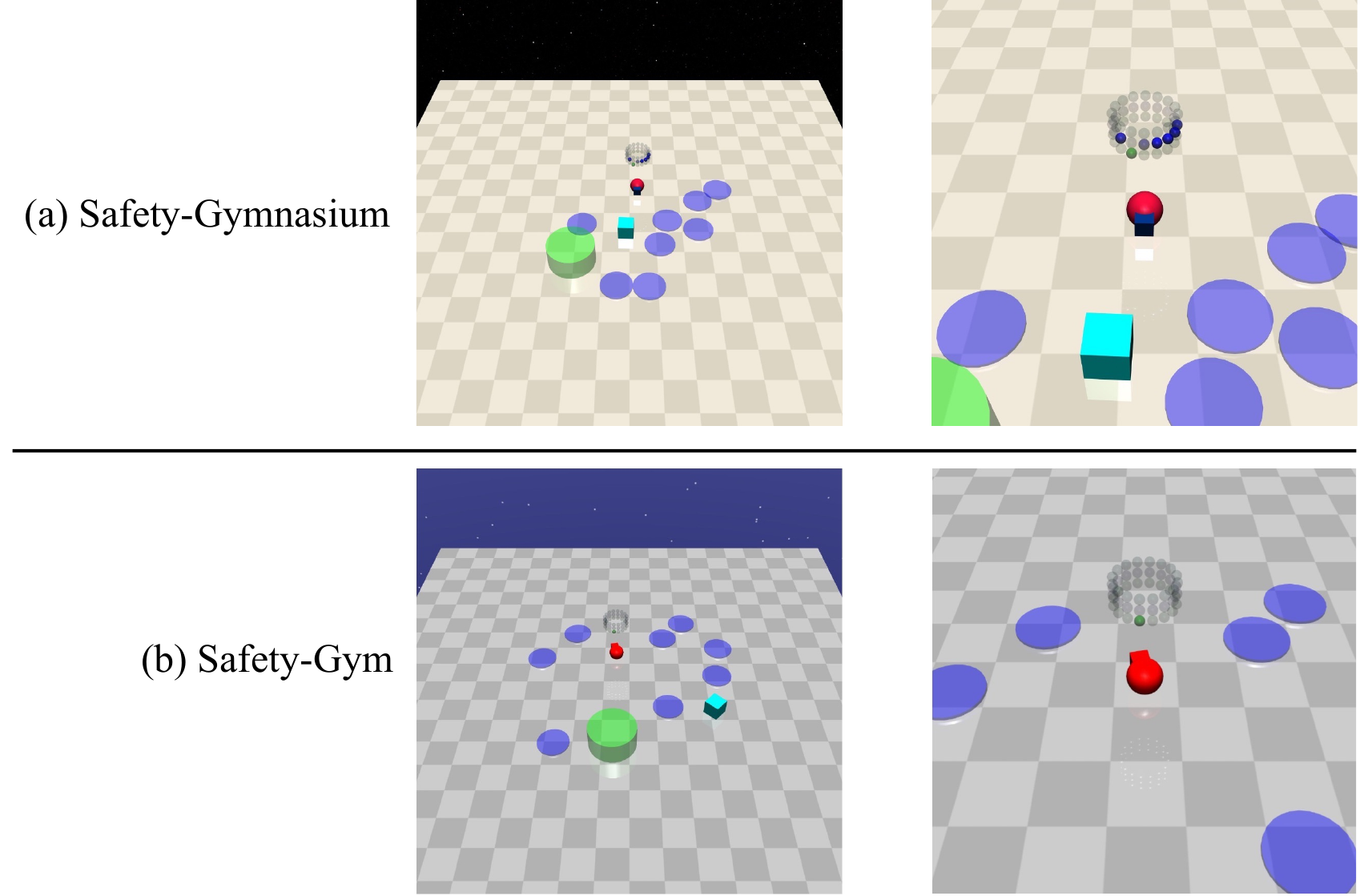}
  \caption{The difference between \texttt{Safety-Gymnasium} and Safety Gym.}
  \label{pic:lidar_bug}
\end{figure}
\textbf{The bug of Natural Lidar.} As shown in \autoref{pic:lidar_bug}, the original Natural Lidar in Safe-Gym\footnote{\url{https://github.com/openai/safety-gym}} has a problem of not being able to detect low-lying objects, which may affect comprehensive environmental observations.

\textbf{The problem of observation space.} In Safety Gym, by default, the observation space is presented as a one-dimensional array. The implementation leads to all ranges in observation space to be $[-\infty, +\infty]$, as shown in the following code:

\begin{lstlisting}
if self.observation_flatten:
    self.obs_flat_size = sum([np.prod(i.shape) for i in self.obs_space_dict.values()])
    self.observation_space = gym.spaces.Box(-np.inf, np.inf, (self.obs_flat_size,), dtype=np.float32)
\end{lstlisting}

While this representation does not lead to behavioral errors in the environment, it can be somewhat misleading for users. To address this issue, we have implemented the Gymnasium's flatten mechanism in the Safety Gym to handle the representation of the observation space. This mechanism reorganizes the observation space into a more intuitive and easily understandable format, enabling users to process and analyze the observation data more effectively.

\begin{lstlisting}
self.obs_info.obs_space_dict = gymnasium.spaces.Dict(obs_space_dict)

if self.observation_flatten:
    self.observation_space = gymnasium.spaces.utils.flatten_space(
        self.obs_info.obs_space_dict
    )
else:
    self.observation_space = self.obs_info.obs_space_dict
assert self.obs_info.obs_space_dict.contains(
    obs
), f'Bad obs {obs} {self.obs_info.obs_space_dict}'

if self.observation_flatten:
    obs = gymnasium.spaces.utils.flatten(self.obs_info.obs_space_dict, obs)
    return obs
\end{lstlisting}

\textbf{Missing cost information.} In Safety Gym, by default, there are only two possible outputs for the cost: \texttt{0} and \texttt{1}, representing whether a cost is incurred or not.

\begin{lstlisting}
# Optionally remove shaping from reward functions.
if self.constrain_indicator:
    for k in list(cost.keys()):
        cost[k] = float(cost[k] > 0.0)  # Indicator function
\end{lstlisting}

We believe that this representation method loses some information. For example, when the robot collides with a vase and causes the vase to move at different velocities, there should be different cost values associated with it to indicate subtle differences in violating constraint behaviors. Additionally, these costs incurred by the actions are accumulated into the total cost. In typical cases, algorithms use the total cost to update the policy if the total cost generated by different obstacles is limited to only two states \texttt{0} and \texttt{1}, the learning potential for multiple constraints is lost when multiple costs are triggered simultaneously.

\textbf{Neglected dependency maintenance leads to conflicts.}

The \textbf{numpy~=1.17.4} will cause the following problems:

\begin{lstlisting}
ValueError: numpy.ndarray size changed, may indicate binary incompatibility. Expected 96 from C header, got 80 from PyObject
\end{lstlisting}
\begin{lstlisting}
AttributeError: module 'numpy' has no attribute 'complex'.
\end{lstlisting}
\newpage
\section{Details of Isaac Gym-based Learning Environments}
\subsection{Supported Agents}
Safety-DexteroudsHand is based on Bi-DexHands (refer to \cite{chen2022towards} for more details). Bi-DexHands aims to establish a comprehensive learning framework for two Shadow Hands, enabling them to possess a wide range of skills similar to those of humans. The Shadow Hand's joint limitations are as follows (refer to \autoref{joint_limit}). The thumb exhibits 5 degrees of freedom with 5 joints, while the other fingers have 3 degrees of freedom and 4 joints each. The joints located at the fingertips are not controllable. Similar to human fingers, the distal joints of the fingers are interconnected, ensuring that the angle of the middle joint is always greater than or equal to that of the distal joint. This design allows the middle phalange to be curved while the distal phalange remains straight. Additionally, an extra joint (LF5) is located at the end of the little finger, enabling it to rotate in the same direction as the thumb. The wrist comprises two joints, facilitating a complete 360-degree rotation of the entire hand.

\begin{table}[htbp]
    \centering
    \caption{Finger range of motion.}
    \resizebox{\columnwidth}{!}{
    \begin{tabular}{cccc}
    \toprule  
    Joints &  Corresponds to the number of \autoref{pic:shadowhand-dyn}& Min & Max\\\hline
    Finger Distal (FF1,MF1,RF1,LF1) & 15, 11, 7, 3 &  0° & 90°\\\hline
    Finger Middle (FF2,MF2,RF2,LF2) & 16, 12, 8, 4 &  0° & 90°\\\hline
    Finger Base Abduction (FF3,MF3,RF3,LF3) & 17, 13, 9, 5 & -15° &90°\\\hline
    Finger Base Lateral (FF4,MF4,RF4,LF4) & 18, 14, 10, 6 & -20° &20°\\\hline
    Little Finger Rotation(LF5) & 19 & 0° &45°\\\hline
    Thumb Distal (TH1) & 20 & -15° &90°\\\hline
    Thumb Middle (TH2)&	21 & -30° &30°\\\hline
    Thumb Base Abduction (TH3)& 22 & -12° &12°\\\hline
    Thumb Base Lateral (TH4)& 23 & 0° &70°\\\hline
    Thumb Base Rotation (TH5)& 24 & -60° &60°\\\hline
    Hand Wrist Abduction (WR1) & 1 & -40° &28°\\\hline
    Hand Wrist Lateral (WR2)& 2 & -28° &8°\\
    \bottomrule 
    \end{tabular}
    }
    \label{joint_limit}
\end{table}

Stiffness, damping, friction, and armature are also important physical parameters in robotics. For each Shadow Hand joint, we show our DoF properties in \autoref{dof_properties}. This part can be adjusted in the Isaac Gym simulator. 

\begin{table}[htbp]
    \centering
    \caption{DoF properties of Shadow Hand.}
    \begin{tabular}{ccccc}
    \toprule  
    Joints& Stiffness & Damping & Friction & Armature\\\hline
    WR1 & 100 & 4.78 & 0 & 0\\\hline
    WR2 & 100 & 2.17 & 0 & 0\\\hline
    FF2 & 100 & 3.4e+38 & 0 & 0\\\hline
    FF3 & 100 & 0.9 & 0 & 0\\\hline
    FF4 & 100 & 0.725 & 0 & 0\\\hline
    MF2 & 100 & 3.4e+38 & 0 & 0\\\hline
    MF3 & 100 & 0.9 & 0 & 0\\\hline
    MF4 & 100 & 0.725 & 0 & 0\\\hline
    RF2 & 100 & 3.4e+38 & 0 & 0\\\hline
    RF3 & 100 & 0.9 & 0 & 0\\\hline
    RF4 & 100 & 0.725 & 0 & 0\\\hline
    LF2 & 100 & 3.4e+38 & 0 & 0\\\hline
    LF3 & 100 & 0.9 & 0 & 0\\\hline
    LF4 & 100 & 0.725 & 0 & 0\\\hline
    TH2 & 100 & 3.4e+38 & 0 & 0\\\hline
    TH3 & 100 & 0.99 & 0 & 0\\\hline
    TH4 & 100 & 0.99 & 0 & 0\\\hline
    TH5 & 100 & 0.81 & 0 & 0\\

    \bottomrule 
    \end{tabular}
    \label{dof_properties}
\end{table}

\begin{figure}[h]
  \centering
  \includegraphics[width=\linewidth]{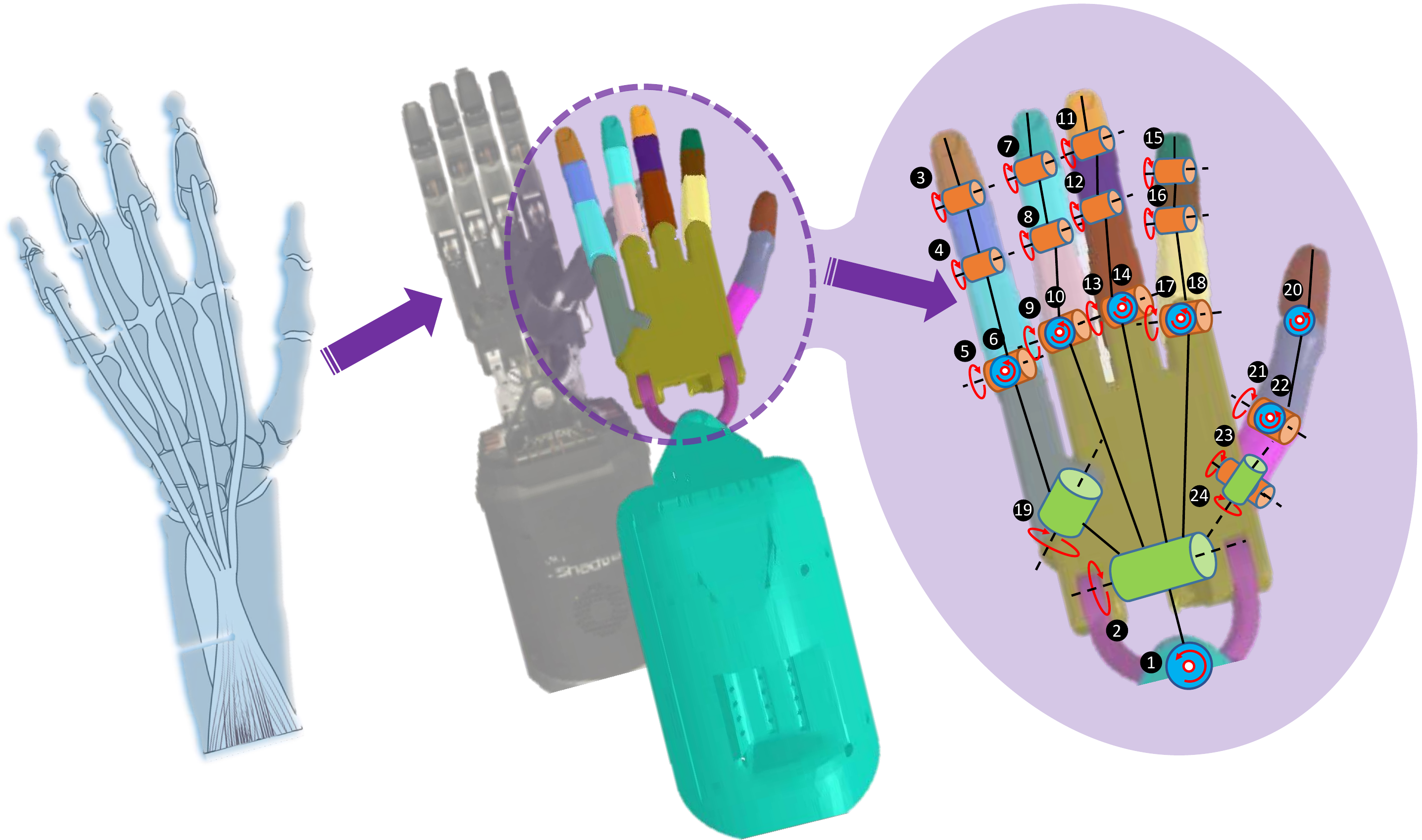}
  \caption{Degree-of-Freedom (DOF) configuration of the Shadow Hand similar to the skeleton of a human hand.}
  \label{pic:shadowhand-dyn}
\end{figure}

\subsection{Task Representation}

\begin{table}[htbp]
    \centering
    \caption{Observation space of dual Shadow Hands.}
    \begin{tabular}{c|c}
    \toprule  
    Index&Description\\\hline
    0 - 23&	right Shadow Hand dof position\\\hline
    24 - 47&	right Shadow Hand dof velocity\\\hline
    48 - 71	& right Shadow Hand dof force\\\hline
    72 - 136&	right Shadow Hand fingertip pose, linear velocity, angle velocity (5 x 13)\\\hline
    137 - 166&	right Shadow Hand fingertip force, torque (5 x 6)\\\hline
    167 - 169&	right Shadow Hand base position\\\hline
    170 - 172&	right Shadow Hand base rotation\\\hline
    173 - 198&	right Shadow Hand actions\\\hline
    
    199 - 222&	left Shadow Hand dof position\\\hline
    223 - 246&	left Shadow Hand dof velocity\\\hline
    247 - 270	& left Shadow Hand dof force\\\hline
    271 - 335&	left Shadow Hand fingertip pose, linear velocity, angle velocity (5 x 13)\\\hline
    336 - 365&	left Shadow Hand fingertip force, torque (5 x 6)\\\hline
    366 - 368&	left Shadow Hand base position\\\hline
    369 - 371&	left Shadow Hand base rotation\\\hline
    372 - 397&	left Shadow Hand actions\\
    \bottomrule 
    \end{tabular}
    \label{observation_dual hands}
\end{table}

\section*{Hand Over}

This scenario encompasses a specific environment comprising two Shadow Hands positioned opposite each other, with their palms facing upwards. The objective is to pass an object between these hands. Initially, the object will randomly descend within the area of the Shadow Hand on the right side. The hand on the right side then grasps the object and transfers it to the other hand. It is important to note that the base of each hand remains fixed throughout the process. Furthermore, the hand initially holding the object cannot directly make contact with the target hand or roll the object towards it. Hence, the object must be thrown into the air, maintaining its trajectory until it reaches the target hand.

In this task, there are 398-dimensional observations and 40-dimensional actions. The reward function is closely tied to the positional discrepancy between the object and the target. As the pose error diminishes, the reward increases significantly. The detailed observation space for each agent can be found in \autoref{observation_handover}, while the corresponding action space is outlined in \autoref{action_handover}.

\textbf{Observations} The observational space for the Hand Over task consists of 398 dimensions, as indicated in \autoref{observation_handover}. However, it is important to highlight that in this particular task, the base of the dual hands remains fixed. Therefore, the observation for the dual hands is compared to a reduced 24-dimensional space, as described in \autoref{observation_dual hands}.

\begin{table}[htbp]
    \centering
    \caption{Observation space of Hand Over.}
    \begin{tabular}{c|c}
    \toprule
    Index&Description\\\hline
    0 - 373&	dual hands observation shown in \autoref{observation_dual hands}\\\hline
    374 - 380&	object pose\\\hline
    381 - 383&	object linear velocity\\\hline
    384 - 386&	object angle velocity\\\hline
    387 - 393&	goal pose\\\hline
    394 - 397&	goal rot - object rot\\
    \bottomrule
    \end{tabular}
    \label{observation_handover}
\end{table}

\textbf{Actions} The action space for a single hand in the Hand Over task comprises 40 dimensions, as illustrated in \autoref{action_handover}.

\begin{table}[htbp]
    \centering
    \caption{Action space of Hand Over.}
    \begin{tabular}{c|c}
    \toprule
    Index&Description\\\hline
    0 - 19&	right Shadow Hand actuated joint \\\hline
    20 - 39&	left Shadow Hand actuated joint\\
    \bottomrule
    \end{tabular}
    \label{action_handover}
\end{table}

\textbf{Rewards} Let the positions of the object and the goal be denoted as $x_o$ and $x_g$ respectively. The translational position difference between the object and the goal, represented as $d_t$, can be computed as $d_t = \lVert x_o - x_g \rVert_2$. Similarly, we define the angular position difference between the object and the goal as $d_a$. The rotational difference, denoted as $d_r$, is then calculated as $d_r = 2 \arcsin(\mathrm{clamp}(\lVert d_a \rVert_2, \text{max} = 1.0))$.

The rewards for the Hand Over task are determined using the following formula:
\begin{equation}
r = \exp(-0.2(\alpha d_t + d_r))
\end{equation}
Here, $\alpha$ represents a constant that balances the rewards between translational and rotational aspects.

\section*{Hand Over Catch}
This environment is made up of a half Hand Over, and Catch Underarm \cite{chen2022towards}, the object needs to be thrown from the vertical hand to the palm-up hand. 

\textbf{Observations} The observational space for this combined task encompasses 422 dimensions, as illustrated in \autoref{catch_over2underarm_obs}.

\begin{table}[htbp]
    \centering
    \caption{Observation space of Hand Over Catch.}
    \begin{tabular}{c|c}
    \toprule  
    Index&Description\\\hline
    0 - 397&	dual hands observation shown in \autoref{observation_dual hands}\\\hline
    398 - 404&	object pose\\\hline
    405 - 407&	object linear velocity\\\hline
    408 - 410&	object angle velocity\\\hline
    411 - 417&	goal pose\\\hline
    418 - 421&	goal rot - object rot\\
    \bottomrule 
    \end{tabular}
    \label{catch_over2underarm_obs}
\end{table}

\textbf{Actions} The action space, consisting of 52 dimensions, is illustrated in \autoref{catch_over2underarm_action}, providing a comprehensive representation of the available actions.

\begin{table}[htbp]
    \centering
    \caption{Action space of Hand Over Catch.}
    \begin{tabular}{c|c}
    \toprule
    Index&Description\\\hline
    0 - 19&	right Shadow Hand actuated joint\\\hline
    20 - 22&	right Shadow Hand base translation\\\hline
    23 - 25&	right Shadow Hand base rotation\\\hline
    26 - 45&	left Shadow Hand actuated joint\\\hline
    46 - 48&	left Shadow Hand base translation\\\hline
    49 - 51&	left Shadow Hand base rotation\\
    \bottomrule
    \end{tabular}
    \label{catch_over2underarm_action}
\end{table}

\textbf{Rewards} Let's denote the positions of the object and the goal as $x_o$ and $x_g$, respectively. The translational position difference between the object and the goal denoted as $d_t$, can be calculated as $d_t = \Vert x_o - x_g \Vert_2$. Additionally, we define the angular position difference between the object and the goal as $d_a$. The rotational difference, denoted as $d_r$, is given by the formula $d_r = 2\arcsin(\text{{clamp}}(\Vert d_a \Vert_2, \text{{max}} = 1.0))$. Finally, the rewards are determined using the specific formula:
\begin{equation}
r = \exp[-0.2(\alpha d_t + d_r)]
\end{equation}
Here, $\alpha$ represents a constant that balances the translational and rotational rewards.

\subsection{Constraint Specification}
\begin{figure}[ht]
  \centering
  \includegraphics[width=\linewidth]{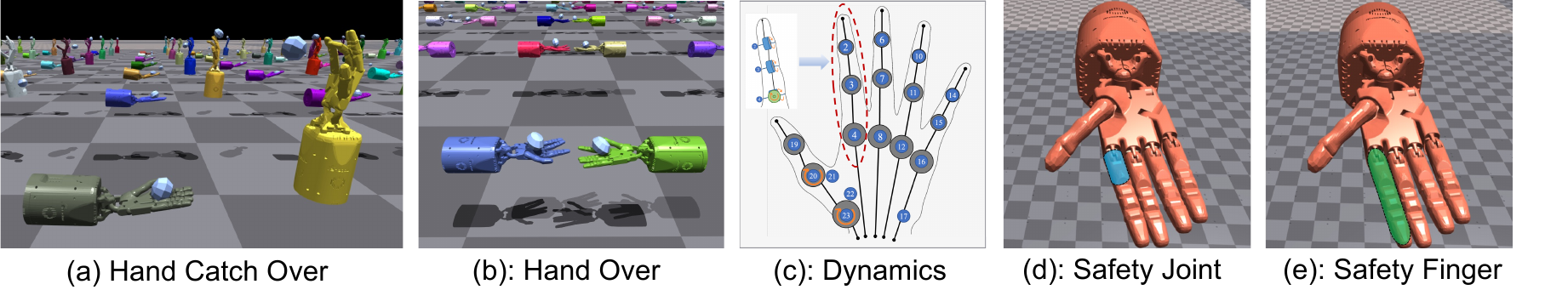}
  \caption{Tasks of Safety-DexterousHands.}
  \label{apppic: dexterous-hand}
\end{figure}

\textbf{Safety Joint} constrains the freedom of joint \ding{175} of the forefinger (please refer to \autoref{apppic: dexterous-hand} (c) and (d)). Without the constraint, joint \ding{175} has freedom of $[-20\degree,20\degree]$. The safety tasks restrict joint \ding{175} within $[-10\degree, 10\degree]$. Let $\mathtt{ang\_4}$ be the angle of joint \ding{175},
and the cost is defined as:
\begin{flalign}
c_t=\mathbb{I}(\mathtt{ang\_4}\not\in [-10\degree, 10\degree]).
\end{flalign}

\textbf{Safety Finger} constrains the freedom of joints  \ding{173},  \ding{174} and \ding{175} of forefinger (please refer to \autoref{apppic: dexterous-hand} (c) and (e)). Without the constraint, joints \ding{173} and \ding{174} have freedom of $[0\degree,90\degree]$ and joint \ding{175} of $[-20\degree,20\degree]$. The safety tasks restrict joints \ding{173}, \ding{174}, and \ding{175} within $[22.5\degree, 67.5\degree]$, $[22.5\degree, 67.5\degree]$, and $[-10\degree, 10\degree]$ respectively.
Let $\mathtt{ang\_2},\mathtt{ang\_3}, \mathtt{ang\_4}$ be the angles of joints \ding{173},  \ding{174}, \ding{175}, and the cost is defined as:
\begin{flalign}
c_t=\mathbb{I}(
\mathtt{ang\_2}\not\in [22.5\degree,67.5\degree],
~\text{or}~
\mathtt{ang\_3}\not\in [22.5\degree,67.5\degree],
~\text{or}~
\mathtt{ang\_4}\not\in [-10\degree,10\degree]
).
\end{flalign}
\end{document}